\pdfoutput=1

\documentclass[11pt]{article}

\PassOptionsToPackage{dvipsnames}{xcolor}
\usepackage[preprint]{acl}

\usepackage{times}
\usepackage{latexsym}
\usepackage[T1]{fontenc}
\usepackage[utf8]{inputenc}
\usepackage{microtype}
\usepackage{inconsolata}
\usepackage{hyperref}
\usepackage{url}

\usepackage{times}
\usepackage{latexsym}
\usepackage[T1]{fontenc}
\usepackage[utf8]{inputenc}

\usepackage{microtype}

\usepackage{inconsolata}

\usepackage{graphicx}

\usepackage{xspace}
\usepackage{amsfonts}
\usepackage{amsthm}
\usepackage{amsmath}
\usepackage{amssymb}
\usepackage{cleveref}
\usepackage{bbm}
\usepackage{caption}
\usepackage{subcaption}
\usepackage{xcolor}
\usepackage{tikz}
\usetikzlibrary{bayesnet}
\usepackage{thm-restate}
\usepackage{booktabs}
\usepackage[most]{tcolorbox}
\usepackage{enumitem}
\usepackage{wasysym}
\usepackage{multirow}
\usepackage{scalerel}
\usepackage{float}
\usetikzlibrary{decorations.pathreplacing,positioning}

\definecolor{darkblue}{rgb}{0, 0, 0.5}
\hypersetup{colorlinks=true, citecolor=darkblue, linkcolor=darkblue, urlcolor=darkblue}
\usepackage{refcount}
\usepackage{amssymb}\usepackage{pifont}

\allowdisplaybreaks

\DeclareFontFamily{OT1}{pzc}{}
\DeclareFontShape{OT1}{pzc}{m}{it}{<-> s * [1.10] pzcmi7t}{}
\DeclareMathAlphabet{\mathpzc}{OT1}{pzc}{m}{it}

\usepackage{array,multirow} 
\usepackage[disable]{todonotes}

\newtheorem{defin}{Definition}
\newtheorem{sublemma}{SubLemma}
\newtheorem{lemma}{Lemma}
\newtheorem{theorem}{Theorem}
\newtheorem{myfact}{Fact}

\newtheorem{proposition}{Proposition}

\newtheorem{corollary}{Corollary}
\newtheorem{reduction}{Reduction}

\newtheorem{mysublemmastep}{Step}
\newtheorem{mylemmastep}{Step}
\counterwithin*{mysublemmastep}{sublemma}
\counterwithin*{mylemmastep}{lemma}

\usepackage{cleveref}
\crefname{section}{\S}{\S\S}
\Crefname{section}{\S}{\S\S}
\crefname{table}{Tab.}{}
\crefname{figure}{Fig.}{Figs.}
\crefname{algorithm}{Alg.}{}
\crefname{equation}{Eq.}{Eqs.}
\crefname{appendix}{App.}{}
\crefname{theorem}{Theorem}{}
\crefname{myfact}{Fact}{Facts}
\Crefname{myfact}{Fact}{Facts}
\crefname{prop}{Proposition}{}
\crefname{defin}{Definition}{}
\crefname{reduction}{Reduction}{}
\crefname{cor}{Corollary}{}
\crefname{observation}{Observation}{}
\crefname{assumption}{Assumption}{}
\crefname{sublemma}{SubLemma}{SubLemmas}
\crefname{lemma}{Lemma}{Lemmas}
\crefformat{section}{\S#2#1#3}
\crefformat{footnote}{#2\footnotemark[#1]#3}

\DeclareMathOperator*{\BigCirc}{\bigcirc}
\DeclareMathOperator*{\argmin}{\mathrm{arg\,min}}

\definecolor{blind_blue}{HTML}{547FEF}
\definecolor{blind_magenta}{HTML}{DC267F}
\definecolor{blind_yellow}{HTML}{FFB000}
\definecolor{blind_orange}{HTML}{FE6100}
\newcommand{\colourbase}{blind_orange}
\newcommand{\coloursubword}{blind_magenta}
\newcommand{\colourcharacter}{blind_blue}
\newcommand{\colourmerge}{blind_magenta}

\newcommand{\colourtemp}{black}
\newcommand{\coloursat}{black}
\newcommand{\colourtok}{black}

\newcommand{\mymacro}[2]{\newcommand{#1}{{\color{\colourbase}#2}}}
\newcommand{\mytemp}[2]{\newcommand{#1}{{\color{\colourtemp}#2}}}
\newcommand{\mymerge}[2]{\newcommand{#1}{{\color{\colourmerge}#2}}}
\newcommand{\mysat}[2]{\newcommand{#1}{{\color{\coloursat}#2}}}
\newcommand{\mytok}[2]{\newcommand{#1}{{\color{\colourtok}#2}}}
\newcommand{\mysubword}[2]{\newcommand{#1}{{\color{\coloursubword}#2}}}
\newcommand{\mycharacter}[2]{\newcommand{#1}{{\color{\colourcharacter}#2}}}

\mytemp{\toktomaxsat}{g}

\newcommand{\defn}[1]{\textbf{#1}}
\mycharacter{\character}{c}
\mycharacter{\characters}{\mathbf{\character}}
\mycharacter{\alphabet}{\Sigma}
\mycharacter{\alphabetsize}{n}
\mysubword{\subword}{s}
\mysubword{\subwords}{\mathbf{\subword}}
\mysubword{\vocab}{\mathcal{S}}
\mysubword{\vocabopt}{\vocab_{\mathtt{opt}}}
\mymacro{\tokenise}{\mathtt{tok}}
\mymacro{\concat}{\mathtt{concat}}
\mymacro{\sumlengths}{\mathtt{sum}}

\mysubword{\vocabboundary}{\vocab_{\mathtt{base}}}
\mysubword{\vocabboundaryup}{\vocab'}

\mysubword{\vocabchoice}{\vocab_{\mathtt{choice}}}
\mysubword{\vocabmissing}{\vocab_{\mathtt{missing}}}
\mysubword{\vocabnc}{\vocab_{\mathtt{nc}}}

\mysubword{\vocabchoiceup}{\vocab''}
\mysubword{\vocabmissingup}{\vocab_{\mathtt{missing}}}
\mysubword{\vocabncup}{\vocab_{\mathtt{nc}}}

\mytok{\dataset}{\mathcal{D}}
\mytok{\vocabsize}{K}
\mytok{\maxsymbols}{\delta}
\mytok{\maxsymbolsopt}{\delta_{\mathtt{opt}}}

\mytemp{\nrepeatvar}{f}
\mytemp{\satnvariables}{J}
\mytemp{\satnclauses}{I}
\mytemp{\vocabvariables}{\vocab_{0}}
\mytemp{\vcnvertices}{J}
\mytemp{\vcmedges}{I}
\mytemp{\addseqntargets}{J}

\mysat{\minclauses}{\gamma}
\mysat{\minclausessol}{\gamma^{\star}}
\mysat{\minclausesopt}{\minclauses_{\texttt{opt}}}
\mysat{\satvar}{X}
\mysat{\satval}{x}
\mytemp{\satyesvar}{x^{\mathtt{T}}}
\mysubword{\satyesvarjtok}{\mathbf{x}^{\mathtt{T}}_j}
\mysubword{\satyesvarjprimetok}{\mathbf{x}^{\mathtt{T}}_{j'}}
\mysubword{\satnotvarjtok}{\mathbf{x}^{\mathtt{F}}_j}
\mysubword{\satnotvarjprimetok}{\mathbf{x}^{\mathtt{F}}_{j'}}
\mysubword{\satyesvarktok}{\mathbf{x}^{\mathtt{T}}_k}

\mysubword{\satnotvarktok}{\mathbf{x}^{\mathtt{F}}_k}

\mycharacter{\satyesvarj}{\mathbf{x}^{\mathtt{T}}_j}
\mycharacter{\satyesvarjprime}{\mathbf{x}^{\mathtt{T}}_{j'}}
\mycharacter{\satnotvarj}{\mathbf{x}^{\mathtt{F}}_j}
\mycharacter{\satnotvarjprime}{\mathbf{x}^{\mathtt{F}}_{j'}}
\mysat{\satvars}{\mathcal{X}}
\mysat{\satvals}{\mathpzc{x}}
\mytemp{\satvalsol}{\satval^{\star}}
\mytemp{\satvalsolj}{\satval^{\star}_j}
\mytemp{\satvalssol}{\satvals^{\star}}
\mytemp{\satvalsoljnostar}{\satval_j}
\mysat{\satclauses}{\mathcal{C}}
\mysat{\satliteral}{L}
\mysat{\satliteralval}{\ell}
\mysat{\sateq}{\varphi}
\mytemp{\spacesymbol}{\subwordstring{1}}
\mytemp{\spacesymboltwo}{\symbolone\symbolone}
\mytemp{\oldspacesymbol}{\circledcirc}

\mytemp{\one}{\mathbbm{1}}
\mytemp{\valfalse}{\mathtt{F}}
\mytemp{\valtrue}{\mathtt{T}}

\mytemp{\approxfactor}{\rho}

\mymerge{\merge}{m}
\mymerge{\merges}{\mathbf{m}}
\mymerge{\mergesopt}{\merges_{\mathtt{opt}}}
\mymerge{\mergespart}{\merges^{\mathtt{p}}}
\mymerge{\mergepart}{m^{\mathtt{p}}}

\mymerge{\mergeset}{\mathcal{M}}
\mymacro{\compression}{\Delta}
\mymacro{\mincompression}{\Delta_{\mathrm{min}}}
\mymacro{\tokeniselengthfun}{\mathtt{toklen}}

\mymacro{\vocabbasemerge}{\vocab_{0}}
\mymacro{\mergebase}{\merges_{0}}
\mymacro{\mergefunc}{\mathtt{merge}}
\mymacro{\emptystring}{\emptyset}
\newcommand{\bigfunccomp}{\bigodot}

\mymacro{\directtoken}{\tokenise_{\text{\pointer}}}

\mymacro{\bottomuptoken}{\tokenise_{\uparrow}}

\mymacro{\reductionfuncthree}{\mathrm{R3}}

\mymacro{\reductionfunc}{\mathrm{R1}}

\mymacro{\reductiontwofunc}{\mathrm{R2}}

\mymacro{\maxsat}{\mathrm{M2S}}

\mymacro{\tomaxsat}{\mathrm{3OM2S}}

\mymacro{\mintok}{\mathrm{Tok}_{\text{\pointer}}}

\mymacro{\mintokmerge}{\mathrm{Tok}_{\uparrow}}

\newcommand{\charstring}[1]{{\color{\colourcharacter}#1}}
\newcommand{\subwordstring}[1]{{\color{\coloursubword}#1}}

\newcommand{\mathcomment}[1]{\text{\textcolor{gray}{#1}}}

\mymacro{\bijectionvocabsat}{\mathrm{Conv}_{\vocab\to\satvals}}
\mymacro{\bijectionmergesat}{\mathrm{Conv}_{\merges\to\satvals}}

\mymerge{\mergesymbol}{\circledcirc}
\newcommand{\mergestring}[2]{\subwordstring{#1} \mathbin{\mergesymbol} \subwordstring{#2}}

\mymacro{\stringequiv}{\stackrel{{\circ}}{=}}
\newcommand{\defeq}{\stackrel{\texttt{\tiny def}}{=}}
\mymacro{\detokenise}{\mathtt{detok}}
\mymacro{\objectivefunc}{\mathfrak{G}}
\mytemp{\objectivefunclength}{\objectivefunc_{\texttt{comp}}}
\mytemp{\objectivefuncreduce}{\objectivefunc_{\texttt{r}}}
\mytemp{\objectivefunctwosat}{\objectivefunc_{M2S}}
\mytemp{\objectivefuncstring}{\objectivefunc_{\mathtt{s}}}
\mytemp{\btheta}{\boldsymbol{\theta}}
\mytemp{\ptheta}{p_{\btheta}}
\mytemp{\R}{\mathbb{R}}
\mytemp{\N}{\mathbb{N}}

\DeclareRobustCommand*{\circled}[1]{\tikz[baseline=(char.base)]{\node[shape=circle,draw,inner sep=1pt] (char) {\normalfont\small #1};}}

\mymacro{\ntokopt}{\mathrm{Tok}^{\star,{\alphabetsize}}}
\mymacro{\alltok}{\mathrm{Tok}}
\mymacro{\allntok}{\mathrm{Tok}^{{\alphabetsize}}}
\mymacro{\dirntokopt}{\mathrm{Tok}_{\text{\pointer}}^{{\alphabetsize,\star}}}
\mymacro{\bupntokopt}{\mathrm{Tok}_{\uparrow}^{{\alphabetsize,\star}}}
\mymacro{\dirtwotokopt}{\mathrm{Tok}_{\text{\pointer}}^{{2,\star}}}
\mymacro{\buptwotokopt}{\mathrm{Tok}_{\uparrow}^{{2,\star}}}
\mymacro{\dirntok}{\mathrm{Tok}_{\text{\pointer}}^{{\alphabetsize}}}
\mymacro{\bupntok}{\mathrm{Tok}_{\uparrow}^{{\alphabetsize}}}

\mymacro{\dirbtok}{\mathrm{Tok}_{\text{\pointer}}^{2}}
\mymacro{\bupbtok}{\mathrm{Tok}_{\uparrow}^{2}}
\mymacro{\minutok}{\mathrm{Tok}_{\text{\pointer}}^{1}}

\newcommand{\dutok}{\texttt{D-1-TOK}\xspace}

\newcommand{\symbolzero}{\charstring{0}}
\newcommand{\symbolone}{\charstring{1}}

\newcommand{\symbolonetok}{\subwordstring{1}}
\mysubword{\spacesymboltwotok}{\symbolonetok\symbolonetok}
\mysubword{\utoken}{a}

\mysubword{\vocablength}{\mathcal{L}}
\mysubword{\vocabsublentok}{\ell}

\mycharacter{\vocabsublen}{\ell}

\mymacro{\splitop}{\mathtt{split}}

\mycharacter{\bigvalue}{B}
\mysubword{\bigvaluetok}{B}
\newcommand{\instance}{\mathcal{I}}

\mymacro{\OPTop}{\mathrm{OPT}}

\mymacro{\costop}{\mathop{\mathop{cost}}}

\mymacro{\encop}{\mathtt{enc}}
\newcommand{\enc}[1]{\ensuremath{\encop\big(\vertex_{#1}\big)}}

\mymacro{\godop}{\mathrm{tok}^1\text{ }}

\mymacro{\minuope}{\mathrm{Tok}_{\mathtt{OPE}}^{1}} 

\mymacro{\addseq}{\mathrm{AddChain}}

\mytemp{\vertices}{\mathcal{V}}
\mytemp{\edges}{\mathcal{E}}
\mytemp{\vertex}{v}
\mytemp{\cover}{\mathcal{C}}
\mytemp{\kbudget}{\psi}
\mymacro{\vc}{\mathrm{VC}}

\newcommand{\p}{$\mathsf{P}$\xspace}
\newcommand{\np}{\ensuremath{\mathsf{NP}}\xspace}
\newcommand{\apx}{$\mathsf{APX}$\xspace}
\newcommand{\ffunctionlreddef}{f}
\newcommand{\gfunctionlreddef}{g}

\newcommand{\solutionlred}{s}

\mysubword{\coinassignments}{\mathcal{Z}}
\mysubword{\coinassignment}{\mathbf{z}}

\mytemp{\powerset}{\mathcal{P}}

\newcommand{\pequalnp}{\ensuremath{\mathsf{P}=\mathsf{NP}}\xspace}
\newcommand{\pnequalnp}{\ensuremath{\mathsf{P}\neq\mathsf{NP}}\xspace}
\newcommand{\pcp}{\ensuremath{\mathsf{PCP}}\xspace}
\newcommand{\yes}{\ensuremath{\mathsf{YES}}\xspace}
\newcommand{\no}{\ensuremath{\mathsf{NO}}\xspace}

\newcommand{\fwr}{\texttt{FINITE-WORD-RANK}\xspace}
\newcommand{\wordset}{\mathcal{X}}
\newcommand{\generatorset}{\mathcal{Y}}
\newcommand{\word}{\mathbf{x}}
\newcommand{\rankwordlength}{\ell}
\newcommand{\generatorword}{y}
\mymacro{\wordrank}{\mathrm{rank}}

\newcommand{\alphabetfwr}{\Sigma_{\texttt{wr}}}
\newcommand{\fwrrep}{R}
\newcommand{\fwrcodelength}{L}
\newcommand{\fwrtotallength}{T}
\mycharacter{\fwrbit}{y}
\mymacro{\binaryencodingfunc}{\texttt{enc}}
\newcommand{\wordsymb}{x}
\newcommand{\fwrpair}[2]{\binaryencodingfunc_{#2}(#1)}
\newcommand{\fwrencoded}[1]{\binaryencodingfunc(#1)}
\newcommand{\fwrdataset}{\dataset}

\newcommand{\tokenisespace}{\mathcal{T}}

\mytemp{\objectivefuncnll}{\objectivefunc_{\mathtt{ll}}}
\newcommand{\totalfreq}{F}
\newcommand{\maxfreq}{f_{\max}}
\newcommand{\fwrpairlength}{q}
\newcommand{\fwrrunbound}{b}
\newcommand{\cubicmaxcut}{\texttt{3-MAX-CUT}\xspace}
\mymacro{\mcfunc}{\mathrm{\mu}}

\mytemp{\srcgraph}{G}
\mytemp{\cuttarget}{\kappa}
\mytemp{\colouring}{\sigma}
\mytemp{\srccolouring}{\sigma_{G}}
\mymacro{\cutfunc}{\mathrm{cut}}
\newcommand{\cutsize}[2]{\cutfunc_{#1}(#2)}
\mytemp{\maxcutval}{\mcfunc}
\newcommand{\maxcutof}[1]{\maxcutval(#1)}
\mytemp{\cutvalue}{C}
 
\mymacro{\blowupfunc}{\mathrm{blowup}}
\mytemp{\blowgraph}{\srcgraph_{\vcnvertices}}
\mytemp{\blowvertices}{\vertices_{\vcnvertices}}
\mytemp{\blowedges}{\edges_{\vcnvertices}}
 
\mytemp{\labelbase}{\beta}
\mytemp{\markB}{B}
\mytemp{\markW}{W}
\newcommand{\opp}[1]{\overline{#1}}      \mytemp{\basecost}{\Delta}
\newcommand{\vtxlen}[1]{\ell_{#1}}        \newcommand{\collen}[2]{\ell^{#2}_{#1}}   \newcommand{\bridgelen}[1]{\ell'_{#1}}    \newcommand{\edgelen}[1]{\ell''_{#1}}     \newcommand{\edgelenij}[2]{\ell''_{#1,#2}}
          \mytemp{\tokenfreq}{f}
\mytemp{\unigramprob}{p}
\mysubword{\vocabcompl}{\vocab_{\mathtt{compl}}}
\mysubword{\vocabcanon}{\vocab_{\colouring}}
\mytemp{\nbichrom}{g}
\newcommand{\palette}[1]{\Pi_{#1}}
 
\mymacro{\reductionfuncfive}{\mathrm{R}_{\text{\pointer}}}
\newcommand{\reductionfuncfivefull}{\reductionfuncfive(\srcgraph)}
\newcommand{\ffunctionap}{f_{\dutok}}
\newcommand{\gfunctionap}{g_{\dutok}}
\mymacro{\decodefunc}{\mathrm{dec}}
\mytemp{\apconst}{\alpha}
\mytemp{\targetratio}{r}
\mymacro{\perfratio}{\mathrm{R}}
\newcommand{\apred}{\leq_{\mathrm{AP}}}
\mytemp{\nclones}{\lambda}
\mytemp{\gapconst}{\varepsilon_{0}}
\mytemp{\blowdeg}{d}
\newcommand{\blowgraphl}{\srcgraph_{\nclones}}
\newcommand{\blowverticesl}{\vertices_{\nclones}}
\newcommand{\blowedgesl}{\edges_{\nclones}}
\newcommand{\arcs}{\mathcal{A}_{\nclones}}
\mytemp{\walk}{\omega}
\mytemp{\nwalks}{M}
\newcommand{\sepsym}[1]{\charstring{\diamond_{#1}}}
\mycharacter{\srcalphabet}{\Gamma}
\mycharacter{\srcword}{\mathbf{w}}
\mytemp{\srclen}{S}
\mytemp{\conflictgraph}{\mathcal{H}}
\mytemp{\indnum}{\alpha}
\mytemp{\covernum}{\tau}
\mymacro{\covfunc}{\mathrm{cov}}
\mytemp{\bigramset}{\mathcal{U}}
\mytemp{\collapsecost}{\nu}
\newcommand{\threesat}{\texttt{3SAT}\xspace}

\mymacro{\idfunc}{\mathrm{id}}
\mytemp{\bucodelength}{L}
\mytemp{\buhalflen}{r}
\mytemp{\buenclen}{q}
\newcommand{\buhalf}[1]{\charstring{\langle #1\rangle}}
\newcommand{\buenc}[1]{\charstring{\widetilde{#1}}}
\mytemp{\nprep}{H}
\newcommand{\mergesprep}{\merges_{\mathtt{prep}}}
\newcommand{\mergessrc}{\merges_{\mathtt{src}}}
\mymacro{\srcdecode}{\mathrm{src}}
\mytemp{\ncentres}{Z}
\mymacro{\reductionfuncbu}{\mathrm{R}_{\uparrow}}
\newcommand{\reductionfuncbufull}{\reductionfuncbu(\srcgraph,\cuttarget)}

\newcommand{\buendsym}{\charstring{\#}}

\mytemp{\walks}{\Omega}

\title{Tokenisation is not in APX}

\newcommand{\makesf}[1]{\textsf{{{#1}}}}
\newcommand{\ethemailadress}[1]{\href{mailto:#1@inf.ethz.ch}{\makesf{#1}}}
\newcommand{\sofiaemailaddress}[1]{\href{mailto:#1@uni-sofia.bg}{\makesf{#1}}}

\author{Violeta Kastreva,~ ~Philip Whittington,~ ~Dennis Komm,~ ~Tiago Pimentel \\
  Department of Computer Science, ETH Z\"urich\\
   \sofiaemailaddress{vkastreva}\makesf{@ethz.ch},
  \makesf{\{}\ethemailadress{philip.whittington},
  \ethemailadress{dennis.komm},
  \ethemailadress{tiago.pimentel}\makesf{\}@inf.ethz.ch}
  }

\begin{document}
\maketitle
\begin{abstract}
In this paper, we prove that, unless \pequalnp, no polynomial-time algorithm can find a tokeniser whose objective is within any constant factor of the optimum, meaning that tokenisation is not in the \apx complexity class.
We prove this for direct and bottom-up tokenisation, under both a compression or unigram log-likelihood objectives, and for binary or larger alphabets.
We further show that, even over unary alphabets, direct tokenisation is \apx-hard, and thus admits no polynomial-time approximation scheme.\looseness=-1
\end{abstract}

\section{Introduction}

\newcommand{\saveforcr}[1]{}
\newcommand{\bpe}{\texttt{BPE}\xspace}
\newcommand{\unigramlm}{\texttt{UnigramLM}\xspace}

Tokenisers are the ``glasses'' through which language models (LMs) see the world, converting raw text---made of bytes or characters---into sequences of tokens.
They define an LM's API costs, 
impact the models' throughput, 
affect their performance, and 
directly shape output distributions \citep{zouhar-etal-2023-tokenization,lesci-etal-2025-causal, ahia-etal-2023-languages}.
It is thus natural to pose the question:
\begin{quote}
    \emph{how should one choose a tokeniser?}
\end{quote}
A recent line of research frames this question as an optimisation problem: 
define an objective function, and find a tokeniser that performs well on it.
In fact, famous tokenisation algorithms can be seen as instantiations of this principle.
\bpe is a greedy algorithm for compression, iteratively merging the most frequent pair of adjacent tokens \citep{gage1994new, sennrich-etal-2016-neural}, while
\unigramlm is an expectation-maximisation algorithm for log-likelihood \citep{kudo-2018-subword, meister2026unigramlm}.
Notably, these methods are purely heuristic, having no guarantees regarding their solutions' quality.\footnote{\bpe, in particular, can have arbitrarily bad compression \citep{kozma2024theoretical}. See \citet{yavuz-etal-2026-objective} for an analysis of how these methods' objective functions and optimisation algorithms interact when selecting a tokeniser.\looseness=-1}

\begin{figure}[t]
\centering
\begin{tikzpicture}[x=0.95mm,y=1mm,
  lbl/.style={font=\scriptsize},
  rob/.style={draw=black!60,fill=black!25,line width=0.3pt},
  roh/.style={draw=blind_magenta,fill=blind_magenta!45,line width=0.3pt},
  opn/.style={draw=black!60,dashed,line width=0.3pt}]
\foreach \y/\la/\lb/\g/\p in {
      0/unary direct/compression/24.2/35,
      -6.5/binary direct/compression/35/60,
      -13/binary direct/log-likelihood/23.1/60,
      -19.5/binary bottom-up/compression/35/67,
      -26/binary bottom-up/log-likelihood/23/46.5}{
    \node[lbl,anchor=west] at (0,\y+1.3) {\la};
    \node[lbl,anchor=west,gray] at (0,\y-1.7) {\lb};
    \draw[opn] (\p,\y-1.7) rectangle (74,\y+1.7);
    \draw[roh] (\g,\y-1.7) rectangle (\p,\y+1.7);
    \ifdim\g mm>23mm \draw[rob] (23,\y-1.7) rectangle (\g,\y+1.7);\fi
  }
\draw[->,>=latex,black!70] (22,-30.5) -- (76,-30.5);
  \foreach \x/\t in {23/{$1$}, 35/{$c$}, 46.5/{any $c$}, 60/{$N^{1-\varepsilon}$}, 74/{$N$}}{
    \draw[black!70] (\x,-29.7) -- (\x,-31.3);
    \node[lbl,anchor=base] at (\x,-34.4) {\t};
  }
  \node[lbl,anchor=north,gray] at (48.2,-34.9) {approximation ratio};
\draw[rob] (23,-41.2) rectangle (26.2,-38.8);
  \node[lbl,anchor=west] at (26.7,-40.1) {known before};
  \draw[roh] (46.5,-41.2) rectangle (49.8,-38.8);
  \node[lbl,anchor=west] at (50.3,-40.2) {this paper};
  \draw[opn] (66,-41.2) rectangle (69.3,-38.8);
  \node[lbl,anchor=west] at (69.8,-40.3) {open};
\end{tikzpicture}
\vspace{-7pt}
\caption{
Approximation ratios ruled out for tokenisation, assuming \pnequalnp.
A ratio of $1$ means exact solutions (\np-hardness), $c$ some constant (\apx-hardness), and ``any $c$'' every constant (not in \apx). 
Under compression, $N$ is achieved by character-level tokenisation.\looseness=-1
}
\label{fig:approximability}
\vspace{-7pt}
\end{figure}
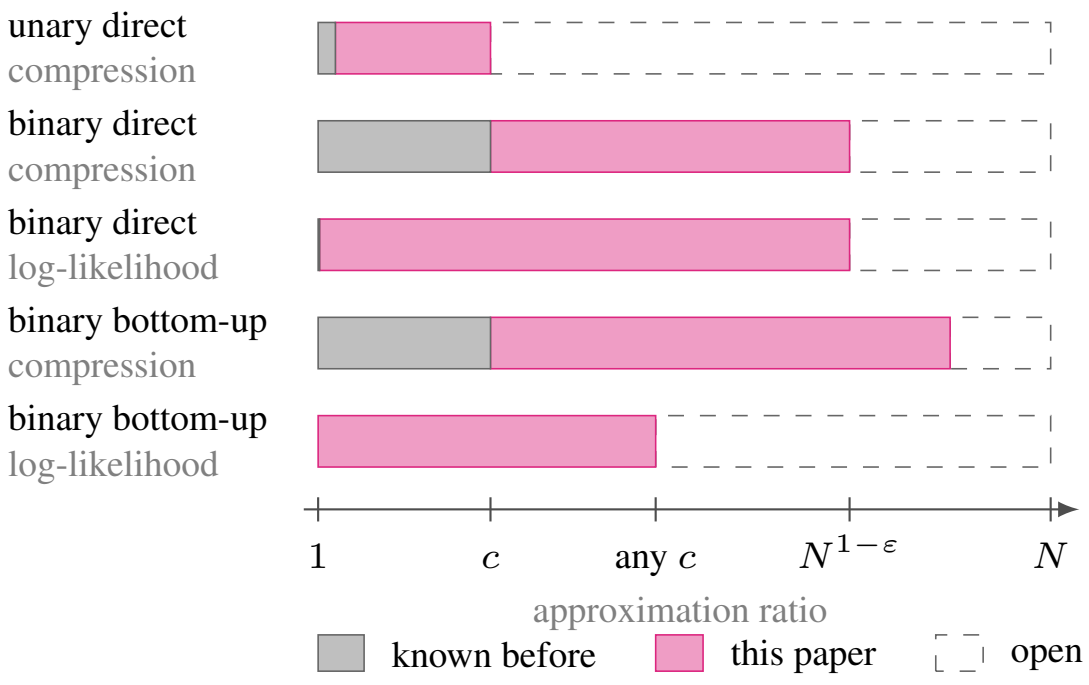

A few recent works have explored whether it is even possible to propose tokenisers with optimality guarantees---all focusing on compression as their analysed objective.
Studying unboundedly large alphabets, a set of initial papers proved that tokenisation's optimisation problem is \np-hard \citep{whittington-etal-2025-tokenisationnpc,kozma2024theoretical,lim-choo-lauw-2025-concurrent}.\footnote{More precisely, 
\citet{whittington-etal-2025-tokenisationnpc} showed that both direct and bottom-up tokenisation are \np-complete,
\citet{kozma2024theoretical} showed that bottom-up tokenisation, with or without partial merges, is \apx-hard, and \citet{lim-choo-lauw-2025-concurrent} showed that direct tokenisation is \np-hard.}
Follow up work then showed it is still \np-hard over unary alphabets, and \apx-hard over binary ones \citep{kastreva2026bounded}.\footnote{Explicitly, \apx-hardness means that, for some constant $c>1$, no polynomial-time algorithm can guarantee a solution within a factor $c$ of the optimum, unless \pequalnp.}
These results, however, leave open three main questions:
\begin{quote}
\emph{(i) is tokenisation hard for other objectives beyond compression?
(ii) does any constant-factor approximation algorithm exist for tokenisation?
(iii) is unary tokenisation also hard to approximate?}
\end{quote}

In this work, we tackle these three questions.
First, we answer (i) and (ii) showing that binary tokenisation is not in \apx; we prove this for direct and bottom-up tokenisation, covering both the compression and unigram log-likelihood objectives.
This means that, unless \pequalnp, no polynomial-time algorithm can approximate binary tokenisation within any constant factor, under either objective.\footnote{Concurrently, \citet{baliga2026pairmerge} showed this for bottom-up tokenisation, and \citet{ion2026hardtokenisation} and \citet{bastan2026bicriteria} for direct tokenisation, all under a compression objective.}
Second, we answer (iii) by proving that unary tokenisation is \apx-hard---so even in such simple alphabets, tokenisation cannot be approximated arbitrarily well (shown in \cref{sec:unary_apx} due to space).
\Cref{fig:approximability} summarises prior results and ours. 

Overall, our results rule out the existence of (unconditional) approximation algorithms for tokenisation.\footnote{In fact, for compression, our bounds nearly match those achieved by trivial character-level tokenisers.}
Guarantees for practical tokenisers must therefore be dataset-specific, rather than hold for all possible inputs.
In fact, in practice, tokenisers perform fairly well when applied to natural languages, coming within about $1\%$ of the optimal compression \citep{tempus2026convex}.
Future work should explore what makes natural language so much easier to tokenise than our worst-case constructions, and explore data-specific algorithmic guarantees.\looseness=-1

\section{Tokenisation}
\label{sec:tokenisation}

Let $\characters\in\alphabet^*$ be a \defn{character-string} over an alphabet $\alphabet$, and let $\dataset=\{\characters_n\}_{m=1}^{M}$ be a \defn{dataset} (a multiset of such strings) with total length \smash{$N\defeq\sum_{\characters\in\dataset}|\characters|$}.\footnote{
We follow the notation of \citet{whittington-etal-2025-tokenisationnpc} here.}
We say the alphabet is \defn{binary} if $|\alphabet|=2$ and \defn{unary} if $|\alphabet|=1$.
A tokeniser's job is then to \emph{segment} $\characters$ as a sequence of \defn{tokens} $\subwords = \langle\subword_1,\ldots,\subword_m\rangle$ from its \defn{vocabulary} $\vocab\!\subset\!\alphabet^+$, which we write as $\tokenise(\characters) \!=\! \subwords$ s.t.\ $\concat(\subwords)\defeq\subword_1\circ\cdots\circ\subword_m = \characters$.
To guarantee any string can be encoded, we enforce $\alphabet\subseteq\vocab$ and define the \defn{vocabulary's budget} as $\vocabsize \defeq |\vocab| - |\alphabet|$.

\newcommand{\tokeniseopt}{\tokenise^{\star}}
\newcommand{\objectivefuncopt}{\objectivefunc^{\star}}
\mytemp{\objectivefunclengthopt}{\objectivefunc^{\star}_{\texttt{comp}}}
\mytemp{\objectivefuncnllopt}{\objectivefunc^{\star}_{\mathtt{ll}}}

Given these definitions, how do we now choose a tokeniser?
As described above, we first choose an objective function $\objectivefunc$; 
given a class of realisable tokenisers $\tokenisespace$, we then select the one optimising it.
The \defn{optimal objective value} achievable is then:\looseness=-1
\begin{align}
     \objectivefuncopt(\dataset) \defeq \min_{\tokenise \in \tokenisespace} \objectivefunc(\tokenise, \dataset).
\end{align}

We say a tokenisation algorithm is a \defn{$\boldsymbol{\approxfactor}$-approximation} if, given any dataset $\dataset$, it returns a tokeniser $\tokenise$ with $\objectivefunc(\tokenise,\dataset) \leq \approxfactor\,\objectivefuncopt$.
A problem with no $\approxfactor\!=\!1$ approximation is then \np-hard,
with no $\approxfactor\!=\!1\!+\!\varepsilon$ approximation (for some $\varepsilon\!>\!0$) is \apx-hard,
and with no $\approxfactor\!=\!c$ (for all constants $c\!>\!1$) is not in the \apx class (all unless \pequalnp).\saveforcr{\footnote{\apx is the class of problems admitting an approximation with some constant $\approxfactor=c$. Formally, \apx requires integer objective values. The surrogate $\lceil\objectivefuncnll/\log2\rceil$ changes approximation ratios by at most a factor of $\frac{3}{2}$, as every positive value of $\objectivefuncnll$ is at least $2\log2$.\looseness=-1}}

\subsection{Tokenisation Objectives}
\vspace{-3pt}

Although no perfect objective seems to exist for tokenisation, several functions are said to correlate with a tokeniser's performance: e.g., unigram log-likelihood \citep{kudo-2018-subword}, 95\%-ile token frequency \citep{gowda-may-2020-finding}, R\'enyi efficiency \citep{zouhar-etal-2023-tokenization}, and compression \citep{galle-2019-investigating}.
It is thus natural to use them as an objective for tokenisation.
We analyse two of them here.

A tokeniser's \defn{compression} $\objectivefunclength$, defined as the number of tokens it uses to represent a dataset, and
its \defn{unigram negative log-likelihood} $\objectivefuncnll$, defined as the log-likelihood of the tokenised dataset under a unigram distribution.
Notably, compression is the objective behind \bpe, while log-likelihood is used by \unigramlm.\footnote{Notably, \unigramlm optimises for a marginal likelihood over all possible segmentations under latent $f_\subword$; instead, we optimise over the log-likelihood of actually produced tokens. Whether our results extend to \unigramlm remains open.}
Writing $\tokenfreq_{\subword}$ for the frequency of token $\subword$ in dataset $\dataset$ and 
$\totalfreq=\sum_{\subword} \tokenfreq_{\subword}$ as the total token count, we formalise these objectives as
\begin{subequations}
\begin{align}
    &\objectivefunclength(\tokenise,\dataset)
    \,\smash{\defeq}
    \sum_{\characters\in\dataset}
    |\tokenise(\characters)|,
    \label{eq:length_objective} \\
    &\objectivefuncnll(\tokenise,\dataset)
    \,\smash{\defeq}
    \sum_{\subword \in \vocab\colon\tokenfreq_{\subword}>0}
    \tokenfreq_{\subword}
    \log\frac{\totalfreq}{\tokenfreq_{\subword}}.
    \label{eq:nll_objective}
\end{align}
\end{subequations}

\vspace{-7pt}
\subsection{Classes of Tokenisers}
\vspace{-2pt}

Several classes of tokenisers exist.
We consider two of them here.
First, given a vocabulary $\vocab$ and a training dataset $\dataset$, \defn{direct tokenisers} segment each character string $\characters \in \dataset$ optimally under the given objective, being defined as:\begin{align}
       \directtoken[\vocab,\dataset](\characters)
    &\,\smash{\defeq}\!\!\!\!\!\!
    \argmin_{
    \substack{
        \subwords\in\vocab^* \\
        \texttt{s.t.}\, \concat(\subwords)=\characters
    }
    }\!\!\!\!\!\!
    \objectivefuncstring[\vocab, \dataset](\subwords, \{ \characters \})
    \label{eq:direct_tokenisation}
\end{align}
where $\objectivefuncstring[\vocab,\dataset]$ returns a function which outputs the objective value (compression or negative log-likelihood) of a specific segmentation $\subwords$ considering the entire training dataset.\footnote{For log-likelihood, $\subwords$ is thus evaluated under the unigram distribution over $\vocab$ that is optimal for the whole dataset $\dataset$.}
Second, \defn{bottom-up tokenisers} are specified by a sequence of \defn{merges} $\merges=\langle\merge_1,\ldots,\merge_{|\merges|}\rangle\in\mergeset^*$, where $\mergeset\defeq\alphabet^+\times\alphabet^+$ and each $\merge_k=\mergestring{\subword_k}{\subword_k'}$, which they apply in order to iteratively merge pairs of tokens in a string.
Writing $\mergefunc_{\merge_k}$ as a function that replaces, from left to right, every non-overlapping occurrence of adjacent tokens $\langle\subword_k,\subword_k'\rangle$ by the single token $\subword_k^{\texttt{new}} = \subword_k\circ\subword_k'$, we formalise these tokenisers as:\begin{align}
    \bottomuptoken[\merges](\characters)
    &\defeq
    \smash{\bigg(\bigfunccomp_{k=1}^{|\merges|}\mergefunc_{\merge_k}\bigg)}
    \big(\characters\big),
    \label{eq:bottomup_tokenisation}
\end{align}
where $\bigfunccomp$ denotes function composition.
\unigramlm is an example of a direct tokeniser, while \bpe is a bottom-up tokeniser. 
We write $\tokenisespace_{\vocabsize}$ for the set of tokenisers with budget $\vocabsize$, noting bottom-up tokenisers have $\vocabsize = |\vocab| - |\alphabet| = |\merges|$.

\vspace{-5pt}
\section{Direct Tokenisation is not in \boldmath\apx\unboldmath}
\vspace{-2pt}

We now prove that direct tokenisation is not in the \apx class.
We do so using a \defn{gap reduction} from a known \np-hard problem, finite-word rank, to \emph{binary} direct tokenisation.
From an instance of this \np-hard \emph{decision problem}---whose desired output is a \yes or \no answer---we build an instance of the tokenisation \emph{optimisation problem} whose optimal objective will be small if the answer to the finite-word rank instance is \yes, and large if it is \no.
By making this gap arbitrarily large, we then show that any constant-factor approximation to tokenisation would allow us to distinguish \yes from \no answers in finite-word rank.\footnote{Gap reductions are a classic technique, used, e.g., to show that the general travelling salesman problem admits no constant-factor approximation \citep{sahni1976pcomplete}.}
Unless \pequalnp, this is thus infeasible in polynomial time.
Notably, proving the hardness of tokenisation with $|\alphabet|=2$ also proves it for all $|\alphabet|>2$.
\saveforcr{This thus proves the result for most (if not all) relevant tokenisation use cases, leaving open only the unary case.}

\vspace{-3pt}
\subsection{Finite-Word Rank} 
\label{sec:finite_word_rank}

\newcommand{\vocabsizefwr}{\vocabsize_{\texttt{wr}}}

As the starting point of our reductions, we will use the finite-word rank decision problem, which was proven to be \np-complete by \citet{neraud1990elementariness}.
Let $\wordset = \{\word_m\}_{m=1}^{M}$ be a finite set of words, composed of strings 
in an alphabet $\alphabetfwr$, i.e., $\word_m \in \alphabetfwr^+$.
This problem then asks if there is any set of strings $\generatorset \subseteq \alphabetfwr^+$ with size at most $\vocabsizefwr$ which covers $\wordset$.\looseness=-1

\begin{defin}
Let $\wordset \subset \alphabetfwr^+$ be a finite set of words and $\vocabsizefwr \in \N$ a target size.
The \defn{finite-word rank} decision problem requires deciding whether there exists a set
$\generatorset \subseteq \alphabetfwr^+$ that covers $\wordset$, i.e., $\wordset \subseteq \generatorset^*$,
with size at most $|\generatorset| \leq \vocabsizefwr$.
Defining $\wordrank(\wordset)$ as $\wordset$'s smallest possible cover size, we write:
\begin{align}
    \vocabsizefwr 
    \stackrel{?}{\geq}
    \wordrank(\wordset) \defeq 
    \min_{
        \generatorset \subseteq \alphabetfwr^+
        \,\mathtt{s.t.}\, \wordset \subseteq \generatorset^*
    }
    |\generatorset|
\end{align}
\end{defin}

The finite-word rank problem is closely related to tokenisation, but differs in two main ways.
First, $\generatorset$ need not contain $\alphabetfwr$, allowing for solutions that cannot represent all strings in \(\alphabetfwr^*\).
Second, it asks about the size of $\generatorset$, instead of how compactly $\wordset$ can be represented using tokens in $\generatorset$.
Leveraging these differences will be the key to our results.

\subsection{Reducing from Finite-Word Rank}
\label{sec:fwr_binary_encoding}

We now provide the reduction from finite-word rank instances to binary direct tokenisation.
The main idea is to define a function $\binaryencodingfunc: \alphabetfwr \to \{0,1\}^*$ which will replace each symbol in finite-word rank instances with long unique binary blocks.
\vspace{-10pt}
\begin{reduction} \label{red:fwr_to_tok}
    Let $(\wordset,\vocabsizefwr)$ define a finite-word rank instance.
    Now, fix an integer $\fwrrep \geq 2$.
    We construct a tokenisation instance as:
    \begin{subequations}
    \begin{align}
        &\vocabsize = \vocabsizefwr, \alphabet = \{0, 1\} \\
        &\dataset = \{\binaryencodingfunc(\wordsymb_1) \circ \cdots \circ \binaryencodingfunc(\wordsymb_{|\word|}) \mid \word \in \wordset\} \\
        &\binaryencodingfunc(\wordsymb) \defeq \binaryencodingfunc_1(\wordsymb) \!\circ\! \binaryencodingfunc_2(\wordsymb) \!\circ\!\cdots\! \circ\! \binaryencodingfunc_\fwrrep(\wordsymb) \\
        &\binaryencodingfunc_r(\wordsymb) \defeq \charstring{11} \circ \charstring{0\fwrbit_1} \circ \charstring{0\fwrbit_2} \circ \cdots \circ \charstring{0\fwrbit_\fwrcodelength} \circ \charstring{0}
    \end{align}
    \end{subequations}
    where $\fwrbit_1\fwrbit_2\cdots\fwrbit_{\fwrcodelength}\in\{0,1\}^{\fwrcodelength}$ is a binary fixed-length unique representation of each pair in $\alphabetfwr\!\times\!\{1, \dots, R\}$;
    we thus have $\fwrcodelength \defeq 
    \left\lceil
        \log_2\!\left(|\alphabetfwr|\fwrrep\right)
    \right\rceil$.
\end{reduction}

We can now use this reduction to prove our main results regarding direct tokenisation.

\begin{restatable}{theorem}{compressionnotinapx}
\label{thm:binary_compression_not_apx}
Unless $\pequalnp$, binary direct tokenisation under the compression objective $\objectivefunclength$ is not in \apx. 
In fact, for every constant $\varepsilon>0$, it admits no polynomial-time $N^{1-\varepsilon}$-approximation.
\end{restatable}
\vspace{-5pt}
\begin{proof}[Proof sketch]
    See a formal proof in \cref{appendix:compressionnotinapx}. 
    For a target approximation factor $\approxfactor$, apply \cref{red:fwr_to_tok} to a finite-word rank instance $(\wordset, \vocabsizefwr)$ using $\fwrrep\geq\lceil\approxfactor\fwrtotallength\rceil$.
We can prove\looseness=-1
\begin{subequations}
\begin{align}
    \wordrank(\wordset)\leq\vocabsizefwr
    &\!\implies\!
    \objectivefunclengthopt(\fwrdataset)\!\leq\!\fwrtotallength, \label{eq:dirtok_comp_lowerbound}\\
    \wordrank(\wordset)>\vocabsizefwr
    &\!\implies\!
    \objectivefunclengthopt(\fwrdataset)\!>\!\fwrrep\!\geq\!\approxfactor\fwrtotallength.
\end{align}
\end{subequations}
where $\fwrtotallength \defeq \sum_{\word \in \wordset} |\word|$.
If $\wordrank(\wordset) \leq \vocabsizefwr$, simply encoding the optimal $\generatorset$ for finite-word rank with $\binaryencodingfunc$ would lead to at most $\fwrtotallength$ tokens being used by the tokeniser, leading to \cref{eq:dirtok_comp_lowerbound}.
If $\wordrank(\wordset)>\vocabsizefwr$, the tokeniser must cut some block $\binaryencodingfunc_r(\wordsymb)$ for every $r\leq\fwrrep$, requiring more than $\fwrrep$ tokens. Setting $\fwrrep=\fwrtotallength^{c}$ with $c\geq3/\varepsilon$ gives $N^{1-\varepsilon}\fwrtotallength<\fwrrep$, ruling out $N^{1-\varepsilon}$-approximations.\looseness=-1
\end{proof}

\vspace{-4pt}
\begin{restatable}{theorem}{logliknotinapx}
\label{thm:binary_loglik_not_apx}
Unless $\pequalnp$, binary direct tokenisation under the unigram
log-likelihood objective is not in \apx.
In fact, for every constant $\varepsilon>0$, it admits no polynomial-time $N^{1-\varepsilon}$-approximation.
\end{restatable}
\vspace{-5pt}
\begin{proof}[Proof sketch]
See a formal proof in \cref{appendix:logliknotinapx}.
Using \cref{red:fwr_to_tok} with $\fwrrep=\fwrtotallength^4$, we can use a similar logic to above---but while also bounding tokens frequencies---to prove that:
\begin{subequations}
\begin{align}
    \wordrank(\wordset)\leq\vocabsizefwr
    &\implies
    \objectivefuncnllopt(\fwrdataset)\leq\fwrtotallength\log\fwrtotallength,
     \\
    \wordrank(\wordset)>\vocabsizefwr 
    & \implies
    \objectivefuncnllopt(\fwrdataset)>\frac{\fwrtotallength^3}{4}. 
\end{align}
\end{subequations}
Setting $\fwrrep=\fwrtotallength^{c}$ with $c\geq4/\varepsilon$ instead raises the \no bound to $\fwrtotallength^{c-1}/4$, which exceeds $N^{1-\varepsilon}\fwrtotallength\log\fwrtotallength$ and so rules out $N^{1-\varepsilon}$-approximations.
\end{proof}

\section{Bottom-up Tokenisation is not in \boldmath\apx\unboldmath}
\label{sec:bottomup}

We now show that binary \emph{bottom-up} tokenisation is also not in the \apx class, under either objective.
To this end, we will reduce from \defn{maximum cut on cubic graphs}.
Here, we rely on a \defn{gap amplification reduction}.
Our reduction outputs a single binary string which collapses into one token within the budget on \yes instances, but keeps a growing number of tokens on \no instances.\footnote{
Notably, this reduction amplifies the gap of the \pcp theorem \citep{arora1998probabilistic,arora1998proof,dinur2007pcp}.
Known reductions carry this gap over to maximum cut on cubic graphs \citep{hastad2001optimal,berman2001efficient}.
}

\subsection{Gap Maximum Cut on Cubic Graphs}
\label{sec:bu_source}

First, let a graph $\srcgraph$ be \emph{cubic} if each of its vertices has degree 3, and \emph{simple} if it has no loops or multi-edges.
Further, let $\vcnvertices$ be a graph's number of vertices, and $\maxcutof{\srcgraph}$ be its maximum cut, i.e., the largest number of edges crossing a partition of its vertices into two sets (see \cref{def:cubic_maxcut} in the appendix).
Computing $\maxcutof{\srcgraph}$ on simple cubic graphs is \apx-complete \citep{alimonti2000some}.
   Moreover, there is a constant $\gapconst>0$ for which it is \np-hard to distinguish cubic graphs with $\maxcutof{\srcgraph}\geq\cuttarget$ from those with $\maxcutof{\srcgraph}\leq\cuttarget-\gapconst\vcnvertices$, for a given target $\cuttarget$ (\cref{lemma:gap_maxcut}).

\subsection{Reducing from Maximum Cut}

Given an instance $(\srcgraph,\cuttarget)$, we (i) replace every node of $\srcgraph$ by $\nclones=\lceil40/\gapconst\rceil$ copies, (ii) write all three-vertex walks of this graph into one string, and (iii) encode it as a binary string.

\newcommand{\srcgraphtilde}{\widetilde{\srcgraph}}
\newcommand{\verticestilde}{\widetilde{\vertices}}
\newcommand{\edgestilde}{\widetilde{\edges}}
\mymacro{\walkfunc}{\texttt{walk}}
\mymacro{\binaryfunc}{\texttt{bin}}
\mymacro{\graphtostringfunc}{\texttt{graph2str}}

\begin{reduction}
\label{red:bu}
    Let $(\srcgraph, \cuttarget)$ be an instance of maximum cut on a simple cubic graph.
    Now, let $\blowgraphl$ be a copy of $\srcgraph$ where every node $\vertex_j$ is replaced with $\nclones=\lceil40/\gapconst\rceil$ copies $\vertex_{j,n}$ ($n \leq \nclones$), and two nodes $\vertex_{j,n}, \vertex_{j',n'}$ are connected if $\vertex_j, \vertex_{j'}$ were connected in $\srcgraph$.
    We construct a tokenisation instance as:
    \begin{subequations}
    \begin{align}
        &\vocabsize = \nprep+3\nwalks+|\arcs|-\nclones^2\cuttarget,\,\, \alphabet = \{0,1\} \\
        &\dataset = \{\bigcirc_{x \in \graphtostringfunc(\blowgraphl)} \binaryfunc(x)\} \\
        &\graphtostringfunc(\blowgraphl) = \Big(\BigCirc_{\walk \in \walkfunc(\blowgraphl)} \walk\,\sepsym{\chi}\Big)\buendsym
    \end{align}
    \end{subequations}
    where $\BigCirc$ denotes string concatenation, $\walkfunc$ returns all three-vertex walks $\walk=\vertex\vertex'\vertex''$ in a graph, $\sepsym{\chi}$ are separator symbols, unique per pair of walk endpoints, $\buendsym$ is an end symbol, and $\binaryfunc$ returns a binary codeword for each node or symbol.
    Further, $|\arcs|$ is twice the number of edges of the blown-up graph, $\nwalks$ is the number of such walks, and $\nprep$ is the number of merges used to build all codewords.
\end{reduction}

Note that $\dataset$ has a single string, which we denote by $\buenc{\srcword}$. We now present our main theorems for bottom-up tokenisation.

\begin{restatable}{theorem}{bucompressionnotinapx}
\label{thm:bu_compression_not_apx}
Unless \pequalnp, binary bottom-up tokenisation under the compression objective is not in \apx.
In fact, it admits no polynomial-time $o(N/\log N)$-approximation on length-$N$ datasets.\looseness=-1
\end{restatable}
\vspace{-3pt}
\begin{proof}[Proof sketch]
See formal proof in \cref{appendix:bu_compression_proof}.
For graphs with large enough $\vcnvertices$, we can show that:
\begin{subequations}
    \begin{align}
    &\maxcutof{\srcgraph}\geq\cuttarget
    \implies
    \objectivefunclengthopt(\dataset)=1,
    \\
    &\maxcutof{\srcgraph}\leq\cuttarget\!-\!\gapconst\vcnvertices
    \implies
    \objectivefunclengthopt(\dataset)>\tfrac{\gapconst}{2}\nclones^2\vcnvertices
    \!
\end{align}
\end{subequations}
which presents a gap amplification.
\end{proof}

For log-likelihood, a single token would have negative log-likelihood $0$, so we add the string $\symbolzero$, which no merge can change, and use $\dataset=\{\buenc{\srcword},\symbolzero\}$.\looseness=-1

\begin{restatable}{theorem}{bulogliknotinapx}
\label{thm:bu_loglik_not_apx}
Unless \pequalnp, binary bottom-up tokenisation under the unigram negative log-likelihood objective $\objectivefuncnll$ is not in \apx. In fact, it admits no polynomial-time $o(\log N)$-approximation on datasets of total length $N$.
\end{restatable}
\vspace{-3pt}
\begin{proof}[Proof sketch]
See formal proof in \cref{appendix:bu_loglik_proof}.
Similarly to before, for large enough $\vcnvertices$, we show that:
\begin{align}
    &\maxcutof{\srcgraph}\geq\cuttarget
    \implies
    \objectivefuncnllopt(\dataset)\leq2\log2,
    \\
    &\maxcutof{\srcgraph}\leq\cuttarget-\gapconst\vcnvertices
    \implies 
    \objectivefuncnllopt(\dataset)>\log\big(\tfrac{\gapconst}{2}\nclones^2\vcnvertices\big) \nonumber
\end{align}
which again presents a gap amplification.
\end{proof}

\section{Conclusion}

In this work, we proved that binary tokenisation is not in \apx, and that unary direct tokenisation is \apx-hard.
In fact, our reductions leave open only relatively trivial approximation guarantees, close in approximation ratio to, e.g., character-level tokenisation.
Any practical approximation guarantee for tokenisers must thus be dataset-specific.
Future work should thus seek such conditional guarantees, e.g., by certifying how close a tokeniser is to optimal on a given target dataset.\footnote{
A first step towards this goal was done by \citet{tempus2026convex} using linear programming.}

\section*{Limitations}
Our hardness results rely on carefully constructed datasets, which differ substantially from natural text; our results thus say little about the natural-language datasets tokenisers are trained on, where current algorithms may be near-optimal.
 Further, our analysis covers two objectives, compression and unigram log-likelihood, and two kinds of tokenisers, direct and bottom-up; 
other objectives, such as R\'enyi efficiency, and other variants, such as those allowing partial merges \citep{kozma2024theoretical}, remain open.
We suspect, however, that most interesting classes of tokenisers and objective functions will also be hard to approximate.

\section*{Acknowledgements}

We would like to thank Gregor Bachmann and Jan Tempus for conversations about this topic.

\bibliography{references}

\onecolumn
\appendix 
\section{Binary Direct Tokenisation is not in  \boldmath\apx\unboldmath}
\subsection{Proof of \cref{thm:binary_compression_not_apx} (Compression)}
\label{appendix:compressionnotinapx}
\compressionnotinapx*

\begin{proof}
Consider a \fwr instance $(\wordset,\vocabsizefwr)$ and its binary encoding
$\fwrdataset$ from \cref{sec:fwr_binary_encoding} (\cref{fig:fwr_enc}).
Recall that
$\fwrtotallength = \sum_{\word\in\wordset}|\word|$.
We first establish the following two implications:
\begin{align}
    \wordrank(\wordset)\leq\vocabsizefwr
    &\implies
    \objectivefunclengthopt(\fwrdataset)\leq\fwrtotallength,
    \label{eq:fwr_compression_yes}
    \\
    \wordrank(\wordset)>\vocabsizefwr
    &\implies
    \objectivefunclengthopt(\fwrdataset)>\fwrrep.
    \label{eq:fwr_compression_no}
\end{align}
This is done in steps \circled{1} and \circled{2} below.
We then show how to choose $\fwrrep$ such that no constant approximation exists (step \circled{3}).
Finally, we strengthen these results to show no $N^{1-\varepsilon}$-approximation exists (step \circled{4}).

\paragraph{Step \circled{1}. First implication (\cref{eq:fwr_compression_yes}).}
Suppose $\wordrank(\wordset)\leq\vocabsizefwr$.
Then there exists a generating set
$\generatorset\subseteq\alphabetfwr^+$ with
$|\generatorset|\leq\vocabsizefwr$ and
$\wordset\subseteq\generatorset^*$.
We construct a direct-tokenisation vocabulary containing, for every
$\generatorword\in\generatorset$, the token
$\fwrencoded{\generatorword}$.
If $|\generatorset|<\vocabsizefwr$, the remaining vocabulary entries may be
chosen arbitrarily.

Take any word $\word\in\wordset$.
Since $\generatorset$ generates $\wordset$, we can write
\begin{align}
    \word
    =
    \generatorword_1
    \generatorword_2
    \cdots
    \generatorword_{\rankwordlength},
    \qquad
    \generatorword_i\in\generatorset.
\qquad
\implies
\qquad
    \fwrencoded{\word}
    =
    \fwrencoded{\generatorword_1}
    \fwrencoded{\generatorword_2}
    \cdots
    \fwrencoded{\generatorword_{\rankwordlength}}
\end{align}
where the implication follows from the construction of the binary encoding.
Hence $\fwrencoded{\word}$ can be represented using at most $\rankwordlength$ tokens.
Moreover, every $\generatorword_i$ is non-empty, so $\rankwordlength\leq|\word|$.
Since direct tokenisation chooses a minimum-length representation,
\begin{align}
    \left|
        \directtoken[\vocab](\fwrencoded{\word})
    \right|
    \leq \rankwordlength
    \leq |\word|.
\end{align}
Summing over all $\word\in\wordset$ gives
\begin{align}
    \objectivefunclength(\directtoken[\vocab],\fwrdataset)
    &\leq
    \sum_{\word\in\wordset}|\word|
    =
    \fwrtotallength.
\end{align}
Therefore,
\begin{align}
    \wordrank(\wordset)\leq\vocabsizefwr
    \implies
    \objectivefunclengthopt(\fwrdataset)
    \leq
    \fwrtotallength.
\end{align}

\paragraph{Step \circled{2}. Second implication (\cref{eq:fwr_compression_no}).}
We prove the contrapositive of \cref{eq:fwr_compression_no}.
Suppose there exists a direct-tokenisation vocabulary $\vocab$ with at most
$\vocabsizefwr$ tokens such that
\begin{align}
    \objectivefunclength(\directtoken[\vocab],\fwrdataset)
    \leq \fwrrep.
    \label{eq:fwr_backward_assumption}
\end{align}
Across the entire tokenised dataset, there are thus at most $\fwrrep$ tokens, and at most
$\fwrrep - 1$ token boundaries.
Now, letting $\rankwordlength_m$ be the number of tokens used to encode $\fwrencoded{\word_m}$, we have that:
\begin{align}
    \sum_{m=1}^{M}(\rankwordlength_m-1)
    <
    \sum_{m=1}^{M}\rankwordlength_m
    \leq \fwrrep,
\end{align}
where $\rankwordlength_m-1$ then represents the number of boundaries between tokens for each word.
As the number of token boundaries is smaller than $\fwrrep$---and since each such boundary can cut through at most one of the $\fwrrep$ blocks $\fwrpair{\wordsymb}{r}$---
there must exist at least one index $j^\star$ such that
\emph{no} block $\fwrpair{\character}{j^\star}$ is cut by a token boundary
anywhere in $\fwrdataset$ (\cref{fig:fwr_cut}).

We now construct a generating set for $\wordset$.
For each non-character token $s\in\vocab$, read from left to right all complete
blocks of the form $\fwrpair{\character}{j^\star}$ contained in $s$, and record
their corresponding symbols $\character$.
If no such block occurs in $s$, we discard it.
Let $\generatorset$ be the set of non-empty words obtained in this way.
Since $\vocab$ contains at most $\vocabsizefwr$ non-character tokens,
\begin{align}
    |\generatorset|\leq\vocabsizefwr.
\end{align}

\begin{figure*}[t]
\centering
\begin{subfigure}{\textwidth}
\centering
\begin{tikzpicture}[x=1mm,y=1mm,
  cell/.style={draw,minimum height=5.2mm,minimum width=14mm,inner xsep=1pt,inner ysep=0pt,outer sep=0pt,font=\footnotesize,anchor=west},
  dots/.style={cell,minimum width=7mm},
  bit/.style={cell,minimum width=6.2mm},
  lbl/.style={font=\footnotesize},
  brc/.style={decorate,decoration={brace,amplitude=3pt,raise=1.5pt}}]
\node[lbl,anchor=east] at (17,0) {$\fwrpair{\charstring{a}}{2}={}$};
  \node[bit] (z1) at (17,0) {$\symbolone$};
  \node[bit,right=0pt of z1] (z2) {$\symbolone$};
  \node[bit,right=0pt of z2] (z3) {$\symbolzero\fwrbit_1$};
  \node[bit,right=0pt of z3] (z4) {$\symbolzero\fwrbit_2$};
  \node[bit,right=0pt of z4] (z5) {$\cdots$};
  \node[bit,right=0pt of z5] (z6) {$\symbolzero\fwrbit_{\fwrcodelength}$};
  \node[bit,right=0pt of z6] (z7) {$\symbolzero$};
  \draw[brc] (z1.north west) -- (z2.north east) node[midway,above=4.5pt,lbl] {start};
  \draw[brc] (z3.north west) -- (z7.north east) node[midway,above=4.5pt,lbl] {identifier of $(\charstring{a},2)$};
\node[lbl,anchor=east] at (17,-14) {$\fwrencoded{\charstring{ab}}={}$};
  \node[cell] (a1) at (17,-14) {$\fwrpair{\charstring{a}}{1}$};
  \node[cell,right=0pt of a1] (a2) {$\fwrpair{\charstring{a}}{2}$};
  \node[dots,right=0pt of a2] (a3) {$\cdots$};
  \node[cell,right=0pt of a3] (a4) {$\fwrpair{\charstring{a}}{\fwrrep}$};
  \node[cell,right=0pt of a4] (b1) {$\fwrpair{\charstring{b}}{1}$};
  \node[cell,right=0pt of b1] (b2) {$\fwrpair{\charstring{b}}{2}$};
  \node[dots,right=0pt of b2] (b3) {$\cdots$};
  \node[cell,right=0pt of b3] (b4) {$\fwrpair{\charstring{b}}{\fwrrep}$};
  \draw[densely dashed,gray] (z1.south west) -- (a2.north west);
  \draw[densely dashed,gray] (z7.south east) -- (a2.north east);
  \draw[brc,decoration={mirror}] (a1.south west) -- (a4.south east) node[midway,below=3.5pt,lbl,text height=2.2ex,text depth=0.4ex] {$\fwrencoded{\charstring{a}}$: $\fwrrep$ blocks};
  \draw[brc,decoration={mirror}] (b1.south west) -- (b4.south east) node[midway,below=3.5pt,lbl,text height=2.2ex,text depth=0.4ex] {$\fwrencoded{\charstring{b}}$: $\fwrrep$ blocks};
\draw[blind_magenta,line width=1.2pt] ([yshift=-26pt]a1.south west) -- ([yshift=-26pt]b4.south east)
    node[midway,below=1pt,lbl] {token $\subwordstring{\fwrencoded{ab}}$ if $\charstring{ab}\in\generatorset$};
\end{tikzpicture}
\caption{Binary encoding of the word $\word=\charstring{ab}$.}
\label{fig:fwr_enc}
\end{subfigure}

\bigskip
\begin{subfigure}{\textwidth}
\centering
\begin{tikzpicture}[x=1mm,y=1mm,
  cell/.style={draw,minimum height=5.2mm,minimum width=14mm,inner sep=0pt,outer sep=0pt,font=\footnotesize,anchor=west},
  keep/.style={cell,fill=blind_yellow!30},
  cut/.style={cell,fill=blind_magenta!15},
  lbl/.style={font=\footnotesize},
  tok/.style={blind_magenta,line width=1.2pt},
  bnd/.style={blind_magenta,line width=1.5pt}]
\newcommand{\fwrBlockRow}[3]{\foreach \st/\sy/\ix [count=\k] in {#3} {\node[\st] (#2-\k) at ({17+14*(\k-1)},#1) {$\fwrpair{\charstring{\sy}}{\ix}$};}}
\newcommand{\fwrBoundary}[1]{\draw[bnd] ([yshift=2.6pt]#1.north) -- (#1.north);
    \draw[bnd] (#1.south) -- ([yshift=-4.4pt]#1.south);}
\newcommand{\fwrTokenSpan}[5]{\draw[tok] ([xshift=#2,yshift=-2.5pt]#1) -- ([xshift=#4,yshift=-2.5pt]#3)
      node[midway,below=0.5pt,lbl] {$\charstring{#5}$};}
\node[lbl,anchor=east] at (15,0) {$\fwrencoded{\charstring{ab}}$};
  \fwrBlockRow{0}{r}{cell/a/1,cell/a/2,keep/a/3,cell/b/1,cell/b/2,keep/b/3}
  \fwrTokenSpan{r-1.south west}{1pt}{r-6.south east}{-1pt}{ab}
\node[lbl,anchor=east] at (15,-12.5) {$\fwrencoded{\charstring{bab}}$};
  \fwrBlockRow{-12.5}{s}{cell/b/1,cell/b/2,keep/b/3,cut/a/1,cell/a/2,keep/a/3,cell/b/1,cell/b/2,keep/b/3}
  \fwrBoundary{s-4}
  \fwrTokenSpan{s-1.south west}{1pt}{s-4.south}{-2pt}{b}
  \fwrTokenSpan{s-4.south}{2pt}{s-9.south east}{-1pt}{ab}
\node[lbl,anchor=east] at (15,-25) {$\fwrencoded{\charstring{bb}}$};
  \fwrBlockRow{-25}{t}{cell/b/1,cell/b/2,keep/b/3,cell/b/1,cut/b/2,keep/b/3}
  \fwrBoundary{t-5}
  \fwrTokenSpan{t-1.south west}{1pt}{t-5.south}{-2pt}{b}
  \fwrTokenSpan{t-5.south}{2pt}{t-6.south east}{-1pt}{b}
\node[keep,minimum width=4.5mm,minimum height=3.4mm] (k) at (17,-37) {};
  \node[lbl,anchor=west] at (k.east) {never cut ($j^\star{=}3$)};
  \node[cut,minimum width=4.5mm,minimum height=3.4mm] (c) at (51,-37) {};
  \draw[bnd] ([yshift=1.8pt]c.north) -- (c.north) (c.south) -- ([yshift=-1.8pt]c.south);
  \node[lbl,anchor=west] at (c.east) {cut};
  \draw[tok] (70,-37) -- (75,-37);
  \node[lbl,anchor=west] at (75,-37) {token};
\end{tikzpicture}
\caption{Tokenising $\wordset=\{\charstring{ab},\charstring{bab},\charstring{bb}\}$, with $\fwrrep=3$.}
\label{fig:fwr_cut}
\end{subfigure}
\caption{
(a)~Each symbol becomes $\fwrrep$ blocks $\fwrpair{\character}{j}$, each a start marker $\symbolone\symbolone$ followed by an identifier of $(\character,j)$.
(b)~Each boundary cuts at most one block, so fewer than $\fwrrep$ boundaries leave some index $j^\star$ uncut; the $j^\star$-blocks of each token spell a word, and these words generate $\wordset$.
}
\label{fig:fwr}
\end{figure*}
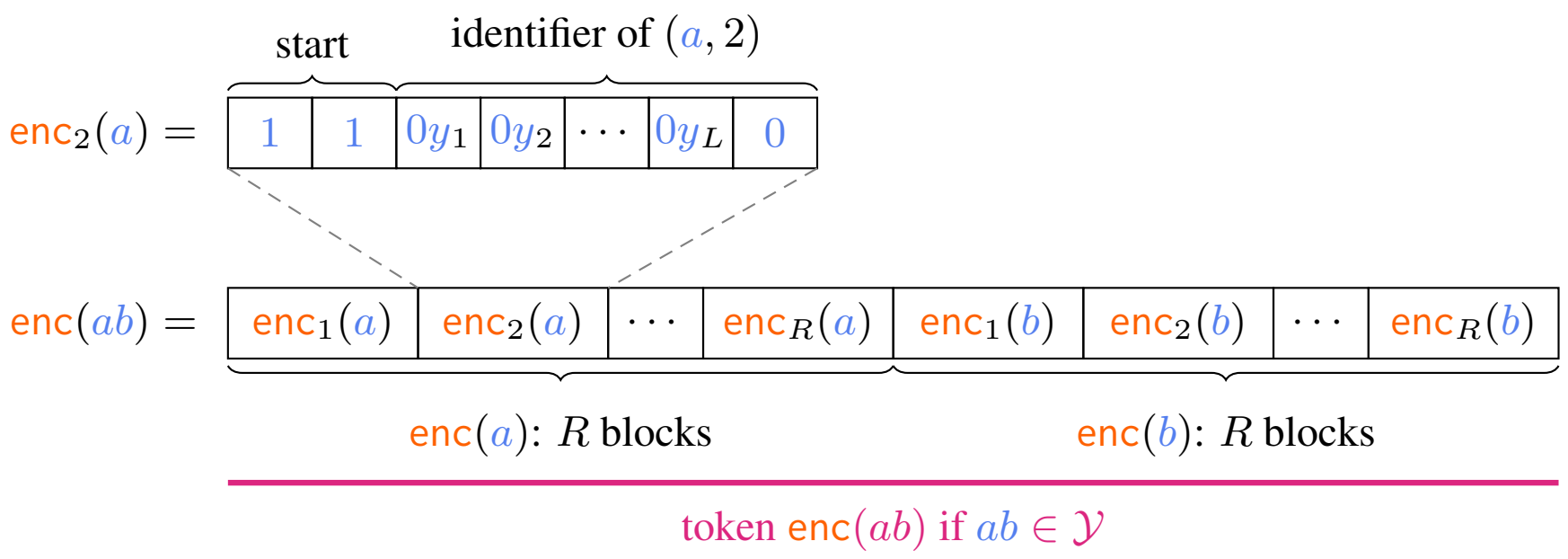
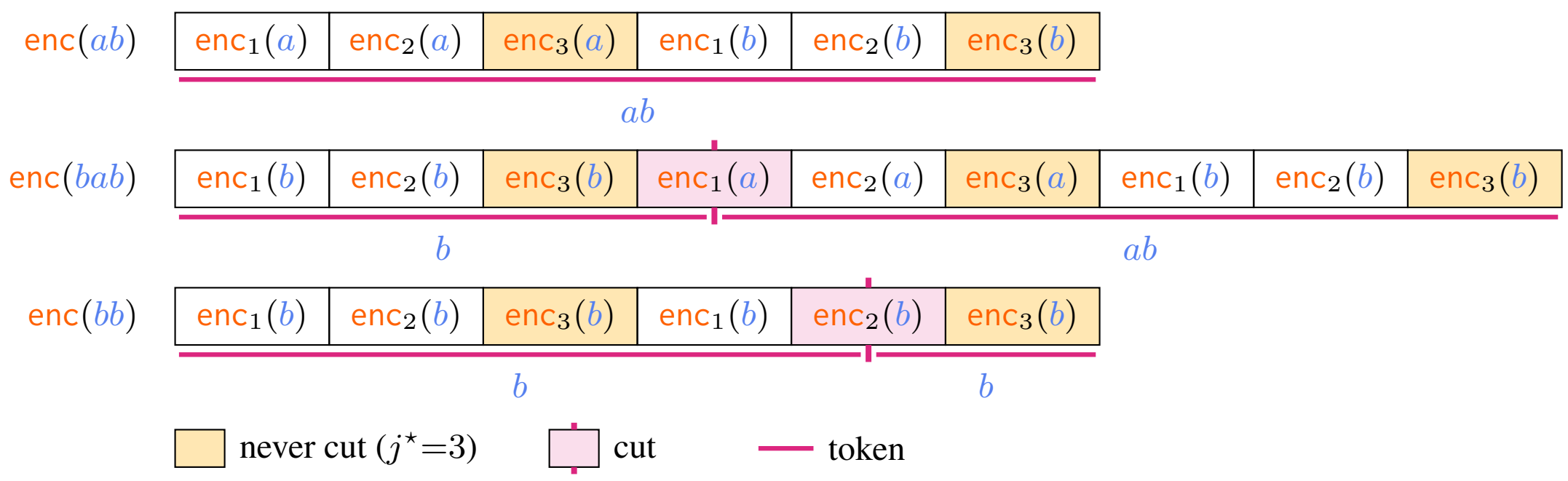 It remains to show that $\generatorset$ generates $\wordset$.
Take any
$\word=\character_1\cdots\character_m\in\wordset$.
Its encoding $\fwrencoded{\word}$ contains, in order, the blocks
\begin{align}
    \fwrpair{\character_1}{j^\star},
    \fwrpair{\character_2}{j^\star},
    \dots,
    \fwrpair{\character_m}{j^\star}.
\end{align}
By the choice of $j^\star$, every one of these blocks lies entirely inside
a single token.
Thus, reading the complete $j^\star$-blocks from the tokens of
$\fwrencoded{\word}$ recovers
$\character_1,\dots,\character_m$ in order.
The corresponding words from $\generatorset$ therefore concatenate to
$\word$, and hence $\word\in\generatorset^*$.

Since this holds for every $\word\in\wordset$, $\wordset\subseteq\generatorset^*$, and therefore
$\wordrank(\wordset)\leq|\generatorset|
    \leq\vocabsizefwr.$
We have shown that
\begin{align}
    \objectivefunclengthopt(\fwrdataset)\leq\fwrrep
    \implies
    \wordrank(\wordset)\leq\vocabsizefwr.
\end{align}
Taking the contrapositive gives
\begin{align}
    \wordrank(\wordset)>\vocabsizefwr
    \implies
    \objectivefunclengthopt(\fwrdataset)>\fwrrep,
\end{align}
proving \cref{eq:fwr_compression_no}.
\paragraph{Step \circled{3}. No constant approximation.}
Suppose that there exists a polynomial-time
$\approxfactor$-approximation algorithm for some constant
$\approxfactor\geq1$. We are going to show that this algorithm could be used to decide \fwr in polynomial time, which is impossible unless \pequalnp. 

To this end, set
$\fwrrep
    =
    \left\lceil
        \approxfactor\fwrtotallength
    \right\rceil$.
Since $\approxfactor$ is constant, $\fwrrep=O(\fwrtotallength)$ and the
construction remains polynomial in size. If $\wordrank(\wordset)\leq\vocabsizefwr$, then by
\cref{eq:fwr_compression_yes} the approximation algorithm returns a solution
of value at most
\begin{align}
    \approxfactor
    \objectivefunclengthopt(\fwrdataset)
    \leq
    \approxfactor\fwrtotallength
    \leq
    \fwrrep.
\end{align}
Conversely, if $\wordrank(\wordset)>\vocabsizefwr$, then by
\cref{eq:fwr_compression_no} every feasible solution has value strictly
greater than $\fwrrep$.

We could thus decide \fwr as follows: build $\fwrdataset$, run the approximation algorithm, count the tokens its vocabulary uses, and answer ``$\wordrank(\wordset)\leq\vocabsizefwr$'' if and only if there are at most $\fwrrep$.
Each of these steps takes polynomial time, which is impossible unless \pequalnp, as \fwr is \np-hard.
Since $\approxfactor$ was arbitrary, no polynomial-time constant-factor approximation exists.\looseness=-1

\paragraph{Step \circled{4}. No $N^{1-\varepsilon}$-approximation.}
The same argument rules out much larger ratios.
Fix $\varepsilon>0$ and set $\fwrrep=\fwrtotallength^{c}$ for a constant $c\geq3/\varepsilon$, and suppose that a polynomial-time $N^{1-\varepsilon}$-approximation algorithm exists.
Let $\vocab$ be the vocabulary it returns on $\fwrdataset$.
We show that the number of tokens it uses would reveal whether $\wordrank(\wordset)\leq\vocabsizefwr$; for sufficiently large $\fwrtotallength$,
\begin{subequations}
\begin{align}
    \wordrank(\wordset)\leq\vocabsizefwr
    &\implies
    \objectivefunclength(\directtoken[\vocab],\fwrdataset)
    \leq
    \fwrrep,
    \label{eq:fwr_compression_alg_yes}
    \\
    \wordrank(\wordset)>\vocabsizefwr
    &\implies
    \objectivefunclength(\directtoken[\vocab],\fwrdataset)
    >
    \fwrrep.
    \label{eq:fwr_compression_alg_no}
\end{align}
\end{subequations}

We now prove both implications.

\textit{First implication (\cref{eq:fwr_compression_alg_yes}).}
The returned vocabulary is within a factor $N^{1-\varepsilon}$ of the optimum, which step \circled{1} (\cref{eq:fwr_compression_yes}) bounds by $\fwrtotallength$ on \yes instances.
It thus uses at most $N^{1-\varepsilon}\fwrtotallength$ tokens, and it suffices to show that this is below $\fwrrep$.
Each encoded pair has $2\fwrcodelength+3=O(\log\fwrtotallength)$ characters, so the dataset has total length $N=\fwrtotallength\,\fwrrep\,(2\fwrcodelength+3)=\fwrtotallength^{c+1}(2\fwrcodelength+3)$, and for sufficiently large $\fwrtotallength$:
\begin{subequations}
\begin{align}
    N^{1-\varepsilon}\,\fwrtotallength
    &=
    \big(\fwrtotallength^{c+1}(2\fwrcodelength+3)\big)^{1-\varepsilon}\,\fwrtotallength
    & \mathcomment{$N=\fwrtotallength^{c+1}(2\fwrcodelength+3)$} \\
    &\leq
    \fwrtotallength^{(c+1)(1-\varepsilon)}\,(2\fwrcodelength+3)\,\fwrtotallength
    & \mathcomment{$(2\fwrcodelength+3)^{1-\varepsilon}\leq2\fwrcodelength+3$} \\
    &\leq
    \fwrtotallength^{c-1}\,(2\fwrcodelength+3)
    & \mathcomment{$\varepsilon(c+1)\geq3$} \\
    &<
    \fwrtotallength^{c}
    =
    \fwrrep,
    & \mathcomment{$2\fwrcodelength+3=O(\log\fwrtotallength)<\fwrtotallength$}
\end{align}
\end{subequations}
which proves the first implication.

\textit{Second implication (\cref{eq:fwr_compression_alg_no}).}
The implication proved in step \circled{2} (\cref{eq:fwr_compression_no}) holds for every choice of $\fwrrep$: on \no instances, every vocabulary within the budget uses more than $\fwrrep$ tokens.
This includes the returned vocabulary $\vocab$, which proves the second implication (\cref{eq:fwr_compression_alg_no}).

Using this, we can decide \fwr in polynomial time.
Hence, unless \pequalnp, no polynomial-time $N^{1-\varepsilon}$-approximation algorithm exists, which concludes the proof.

\end{proof}

\subsection{Proof of \cref{thm:binary_loglik_not_apx} (Log-Likelihood)}
\label{appendix:logliknotinapx}
\logliknotinapx*

\begin{proof}
Consider a \fwr instance $(\wordset,\vocabsizefwr)$ and its binary encoding
$\fwrdataset$ from \cref{sec:fwr_binary_encoding} (\cref{fig:fwr_enc}).
Recall that
\begin{align}
    \fwrtotallength
    =
    \sum_{\word\in\wordset}|\word|,
\end{align}
and, for this reduction, set
$\fwrrep=\fwrtotallength^4.$
Let $\fwrpairlength
    \defeq
    |\fwrpair{\wordsymb}{r}|
    =
    2\fwrcodelength+3$ 
denote the common length of an encoded pair, and let     $\fwrrunbound
    \defeq
    \fwrtotallength+2\fwrpairlength-1.$
We establish the following gap:
\begin{align}
    \wordrank(\wordset)\leq\vocabsizefwr
    &\implies
    \objectivefuncnllopt(\fwrdataset)
    \leq
    \fwrtotallength\log\fwrtotallength,
    \label{eq:fwr_loglik_yes}
    \\
    \wordrank(\wordset)>\vocabsizefwr
    &\implies
    \objectivefuncnllopt(\fwrdataset)
    >
    \frac{\fwrtotallength^3}{4},
    \label{eq:fwr_loglik_no}
\end{align}
for sufficiently large $\fwrtotallength$.
This is done in steps \circled{1} and \circled{2}. We then show that no constant approximation exists (step \circled{3}), and finally that no $N^{1-\varepsilon}$-approximation exists (step \circled{4}).
\paragraph{Step \circled{1}. First implication (\cref{eq:fwr_loglik_yes}).}
Suppose $\wordrank(\wordset)\leq\vocabsizefwr$.
As in the first implication of
\cref{thm:binary_compression_not_apx}, there exists a vocabulary $\vocab$ and a segmentation of $\fwrdataset$ over $\vocab$ using at most $\fwrtotallength$ tokens.
Let $\totalfreq\leq\fwrtotallength$ denote this total token count and let
$\tokenfreq_\subword$ denote the frequency of token $\subword$. Since $\directtoken[\vocab]$ minimises the log-likelihood over all segmentations of $\fwrdataset$, we have
\begin{align}
    \objectivefuncnll(\directtoken[\vocab],\fwrdataset)
    &\leq
    \sum_{\subword\in\vocab}
    \tokenfreq_\subword\log\frac{\totalfreq}{\tokenfreq_\subword}
    \leq
    \sum_{\subword\in\vocab}
    \tokenfreq_\subword\log\totalfreq
    =
    \totalfreq\log\totalfreq
    \leq \fwrtotallength\log\fwrtotallength.
\end{align}
Therefore,
\begin{align}
    \wordrank(\wordset)\leq\vocabsizefwr
    \implies
    \objectivefuncnllopt(\fwrdataset)
    \leq
    \fwrtotallength\log\fwrtotallength.
\end{align}

\paragraph{Step \circled{2}. Second implication (\cref{eq:fwr_loglik_no}).}
Suppose $\wordrank(\wordset)>\vocabsizefwr$, and let $\vocab^\star$ be a
vocabulary minimising the negative log-likelihood objective.
Let $\totalfreq^\star$ be the total number of tokens produced by
$\directtoken[\vocab^\star]$ on $\fwrdataset$, let $\tokenfreq_\subword$ be the frequency of
token $\subword$, and let
\begin{align}
    \maxfreq
    \defeq
    \max_{\subword\in\vocab^\star} \tokenfreq_\subword.
\end{align}

We first bound the longest consecutive occurrence of any token $\subword$ to have length at most $\fwrrunbound= \fwrtotallength+2\fwrpairlength-1$.
The proof for this step is given in \cref{lemma:boundedfreq}. 
We then use this bound to control token frequencies.
Let $M=|\wordset|$ be the number of strings in the original \fwr instance.
Fix a token $\subword$ with frequency $\tokenfreq_\subword$.
This token occurs in sequences of at most $\fwrrunbound$ consecutive appearances at a time, and each of these sequences must end before either another token (of which there are at most $\totalfreq^\star-\tokenfreq_\subword$) or the strings end (of which there are $M$).
Its frequency can thus be bounded as:
\begin{align}
    \tokenfreq_\subword
    \leq
    \fwrrunbound
    \left(
        \totalfreq^\star-\tokenfreq_\subword+M
    \right).
    \label{eq:fwr_frequency_bound}
\end{align}
with the same holding for the most frequent token.
Now, with a slight rearrangement, we can show that:
\begin{subequations}
\begin{align}
    \fwrrunbound
    \left(
        \totalfreq^\star-\maxfreq+M
    \right)
    &\geq
    \maxfreq \\
    \fwrrunbound
    \left(
        \totalfreq^\star-\maxfreq
    \right)
    &\geq
    \maxfreq - \fwrrunbound\,M \\
    (\fwrrunbound + 1)
    \left(
        \totalfreq^\star-\maxfreq
    \right)
    &\geq
    \totalfreq^\star - \fwrrunbound\,M \\
\totalfreq^\star-\maxfreq
    &\geq
    \frac{
        \totalfreq^\star-\fwrrunbound M
    }{
        \fwrrunbound+1
    }.
    \label{eq:fwr_fmax_bound}
\end{align}
\end{subequations}
We can now lower-bound the optimal negative log-likelihood: 
\begin{subequations}
\begin{align}
    \objectivefuncnllopt(\fwrdataset)
    &=
    \sum_{\subword\in\vocab^\star}
    \tokenfreq_\subword\log\frac{\totalfreq^\star}{\tokenfreq_\subword}
   & \mathcomment{definition of ll objective} \\
    &\geq
    \sum_{\subword\in\vocab^\star}
    \tokenfreq_\subword\log\frac{\totalfreq^\star}{\maxfreq}
    & \mathcomment{$\tokenfreq_\subword \leq \maxfreq$}\\
    &=
    \totalfreq^\star
    \log\frac{\totalfreq^\star}{\maxfreq}
    \\
     &\geq
    \totalfreq^\star
    \left(
        1-\frac{\maxfreq}{\totalfreq^\star}
    \right)
    & \mathcomment{$-\log x \geq 1-x$}\label{eq:fwr_log_inequality}
    \\
    &=
    \totalfreq^\star-\maxfreq 
    \\
    &\geq
    \frac{
        \totalfreq^\star-\fwrrunbound M
    }{
        \fwrrunbound+1
    }. &\mathcomment{\cref{eq:fwr_fmax_bound}}
    \label{eq:fwr_nll_frequency_lower_bound}
\end{align}
\end{subequations}

By the second implication of
\cref{thm:binary_compression_not_apx}, the assumption
$\wordrank(\wordset)>\vocabsizefwr$ implies that every feasible vocabulary produces more than $\fwrrep$ tokens on $\fwrdataset$.
As such, we have that $\totalfreq^\star>\fwrrep$ and
thus,
\begin{align}
    \objectivefuncnllopt(\fwrdataset)
    &>
    \frac{
        \fwrrep-\fwrrunbound M
    }{
        \fwrrunbound+1
    }.
    \label{eq:fwr_nll_R_lower_bound}
\end{align}

Recall that $\fwrrep=\fwrtotallength^4$.
Moreover, since every word in $\wordset$ is non-empty,
$M\leq\fwrtotallength$.
Each encoding pair has
$\fwrpairlength=2\fwrcodelength + 3=O(\log\fwrtotallength)$ characters, and therefore, for sufficiently
large $\fwrtotallength$,
$\fwrrunbound = \fwrtotallength + 2\fwrpairlength - 1 <2\fwrtotallength.$
Consequently,
\begin{subequations}
    \begin{align}
    \objectivefuncnllopt(\fwrdataset)
    &>
    \frac{
        \fwrtotallength^4-\fwrrunbound M
    }{
        \fwrrunbound+1
    }
    & \mathcomment{\cref{eq:fwr_nll_R_lower_bound}, $\fwrrep=\fwrtotallength^4$} \\
    &\geq
    \frac{
        \fwrtotallength^4-\fwrrunbound\,\fwrtotallength
    }{
        \fwrrunbound+1
    }
    & \mathcomment{$M\leq\fwrtotallength$} \\
    &>
    \frac{
        \fwrtotallength^4-2\fwrtotallength^2
    }{
        2\fwrtotallength
    }
    & \mathcomment{$\fwrrunbound<2\fwrtotallength$} \\
    &=
    \frac{\fwrtotallength^3}{2}-\fwrtotallength \\
    &\geq
    \frac{\fwrtotallength^3}{4}
    & \mathcomment{$\fwrtotallength\geq2$}
\end{align}
\end{subequations}
proving the implication (\cref{eq:fwr_loglik_no}).

\paragraph{Step \circled{3}. No constant approximation.}
Suppose for contradiction that there exists a polynomial-time $\approxfactor$-approximation algorithm for some constant
$\approxfactor\geq1$. We show that it could be used to decide \fwr in polynomial time, which is impossible unless \pequalnp.

For sufficiently large $\fwrtotallength$,
\begin{align}
    \approxfactor\,
    \fwrtotallength\log\fwrtotallength
    <
    \frac{\fwrtotallength^3}{4}.
\end{align}

Hence, on a \yes instance of \fwr, the algorithm returns a solution of value
strictly below $\fwrtotallength^3/4$, while on a \no instance every feasible
solution has value strictly above $\fwrtotallength^3/4$.
We could thus decide \fwr as follows: build $\fwrdataset$, run the approximation algorithm, compute the objective of the vocabulary it returns, and answer ``$\wordrank(\wordset)\leq\vocabsizefwr$'' if and only if it is below $\fwrtotallength^{3}/4$.
Each of these steps takes polynomial time, which is impossible unless $\pequalnp$, as \fwr is \np-hard.

Since $\approxfactor$ was arbitrary, no polynomial-time constant-factor approximation exists.

\paragraph{Step \circled{4}. No $N^{1-\varepsilon}$-approximation.}
As for \cref{thm:binary_compression_not_apx}, the same argument rules out much larger ratios.
Fix $\varepsilon>0$ and set $\fwrrep=\fwrtotallength^{c}$ for a constant $c\geq \max\{3, 4/\varepsilon\}$. 

Suppose for a contradiction that a polynomial-time $N^{1-\varepsilon}$-approximation algorithm exists, and let $\vocab$ be the vocabulary this algorithm returns on $\fwrdataset$.
We show that the objective of this vocabulary would reveal whether $\wordrank(\wordset)\leq\vocabsizefwr$; for sufficiently large $\fwrtotallength$,
\begin{subequations}
\begin{align}
    \wordrank(\wordset)\leq\vocabsizefwr
    &\implies
    \objectivefuncnll(\directtoken[\vocab],\fwrdataset)
    <
    \frac{\fwrtotallength^{c-1}}{4},
    \label{eq:fwr_loglik_alg_yes}
    \\
    \wordrank(\wordset)>\vocabsizefwr
    &\implies
    \objectivefuncnll(\directtoken[\vocab],\fwrdataset)
    >
    \frac{\fwrtotallength^{c-1}}{4}
    \label{eq:fwr_loglik_alg_no}
\end{align}
\end{subequations}
and we could then use the value given by the approximation algorithm to solve the \fwr decision problem in polynomial time. 
We now prove both implications.

\textit{First implication (\cref{eq:fwr_loglik_alg_yes}).}
The returned vocabulary is within a factor $N^{1-\varepsilon}$ of the optimum, which step \circled{1} (\cref{eq:fwr_loglik_yes}) bounds by $\fwrtotallength\log\fwrtotallength$ on \yes instances. 
Its objective is thus at most $N^{1-\varepsilon}\fwrtotallength\log\fwrtotallength$, and it suffices to show that this is below $\fwrtotallength^{c-1}/4$.
The dataset has total length $N=\fwrtotallength\fwrrep\fwrpairlength=\fwrtotallength^{c+1}\fwrpairlength$, so for sufficiently large $\fwrtotallength$:
\begin{subequations}
\begin{align}
    N^{1-\varepsilon}\fwrtotallength\log\fwrtotallength
    &=
    \big(\fwrtotallength^{c+1}\fwrpairlength\big)^{1-\varepsilon}\,\fwrtotallength\log\fwrtotallength
    & \mathcomment{$N=\fwrtotallength^{c+1}\fwrpairlength$} \\
    &\leq
    \fwrtotallength^{(c+1)(1-\varepsilon)}\,\fwrpairlength\,\fwrtotallength\log\fwrtotallength
    & \mathcomment{$\fwrpairlength^{1-\varepsilon}\leq\fwrpairlength$} \\
    &\leq
    \fwrtotallength^{c-2}\,\fwrpairlength\log\fwrtotallength
    & \mathcomment{$\varepsilon(c+1)\geq4$} \\
    &<
    \frac{\fwrtotallength^{c-1}}{4}
    & \mathcomment{$4\fwrpairlength\log\fwrtotallength<\fwrtotallength$}
\end{align}
\end{subequations}
which proves the implication.

\textit{Second implication (\cref{eq:fwr_loglik_alg_no}).}
No vocabulary has a lower objective than the optimum, so it suffices to show that the optimum itself exceeds $\fwrtotallength^{c-1}/4$ on \no instances.
For this, we reuse the bound of step \circled{2} (\cref{eq:fwr_nll_R_lower_bound}), which holds for every choice of $\fwrrep$.
Each encoded pair still has $\fwrpairlength=O(\log\fwrtotallength)$ characters, so $\fwrrunbound=\fwrtotallength+2\fwrpairlength-1<2\fwrtotallength$ for sufficiently large $\fwrtotallength$, and we get:

\begin{subequations}
\begin{align}
    \objectivefuncnllopt(\fwrdataset)
     &>
    \frac{\fwrtotallength^{c}-\fwrrunbound M}{\fwrrunbound+1}
    & \mathcomment{\cref{eq:fwr_nll_R_lower_bound}, $\fwrrep=\fwrtotallength^{c}$} \\
    &\geq
    \frac{\fwrtotallength^{c}-\fwrrunbound\,\fwrtotallength}{\fwrrunbound+1}
    & \mathcomment{$M\leq\fwrtotallength$} \\
    &>
    \frac{\fwrtotallength^{c}-2\fwrtotallength^{2}}{2\fwrtotallength}
    & \mathcomment{$\fwrrunbound<2\fwrtotallength$} \\
    &=
    \frac{\fwrtotallength^{c-1}}{2}-\fwrtotallength \\
    &\geq
    \frac{\fwrtotallength^{c-1}}{4}
    & \mathcomment{$\fwrtotallength^{c-2}\geq4$}
\end{align}
\end{subequations}
which proves the second implication.

An $N^{1-\varepsilon}$-approximation would thus return objective below $\fwrtotallength^{c-1}/4$ on \yes instances,
while every tokeniser exceeds it on \no instances, deciding \fwr in polynomial time.
\end{proof}

\begin{lemma}
\label{lemma:boundedfreq}
    Under \cref{red:fwr_to_tok}, every \defn{run}, i.e., maximal sequence of consecutive occurrences of the same token in a tokenised string of $\fwrdataset$, has length at most $\fwrtotallength+2\fwrpairlength-1$.
\end{lemma}
\begin{proof}
    
Consider any token $\subword$ of length $d$ and any maximal run of $r$ consecutive
copies of $\subword$ in the tokenisation of some encoded word
$\fwrencoded{\word}$.

If $d<\fwrpairlength$ and $r\geq2\fwrpairlength$, then the overlap between
the run and its shift by $d$ has length at least
\begin{align}
    (r-1)d
    \geq
    2\fwrpairlength-1
    \geq
    \fwrpairlength+1.
\end{align}
This overlap therefore contains a complete occurrence of the marker
$\symbolone\symbolone$.
Since the run is periodic with period $d$, another occurrence of
$\symbolone\symbolone$ appears exactly $d$ positions later.
However, by construction, $\symbolone\symbolone$ occurs only at the
beginnings of encoded pairs, which are spaced
$\fwrpairlength$ positions apart.
Hence $\fwrpairlength$ divides $d$, contradicting
$d<\fwrpairlength$.
Thus,
\begin{align}
    d<\fwrpairlength
    \implies
    r\leq2\fwrpairlength-1.
    \label{eq:fwr_short_token_run}
\end{align}

If $d=\fwrpairlength$, three consecutive copies of $\subword$ would contain two
complete consecutive pair encodings at corresponding positions.
Periodicity would force these two pair encodings to be identical, although
their indices differ by one modulo $\fwrrep$.
Since $\fwrrep\geq2$, this is impossible, and hence $r\leq2$.

Finally, suppose $d>\fwrpairlength$ and $r\geq2$.
The first occurrence of $s$ contains a complete marker
$\symbolone\symbolone$, and the identical second occurrence contains the
same marker exactly $d$ positions later.
As above, this implies that $\fwrpairlength$ divides $d$.
Write $d=k\fwrpairlength.$
Since $d>\fwrpairlength$, we have $k\geq2$, so the first copy of $\subword$
contains at least one complete encoded pair.
The corresponding pair in the second copy lies exactly $k$ blocks later
and must be identical.
The block indices cycle through
$1,\dots,\fwrrep$, so equality of the two encoded pairs implies
\begin{align}
    k\equiv0\pmod{\fwrrep}.
\end{align}
Consequently,
$d\geq\fwrrep\fwrpairlength.$ Since    $ |\fwrencoded{\word}|
    =
|\word|\fwrrep\fwrpairlength,$ 
such a token can occur consecutively at most $|\word|\leq\fwrtotallength$
times.

Combining the three cases, every run of any token has length at most
\begin{align}
    \fwrrunbound
    =
    \fwrtotallength+2\fwrpairlength-1.
    \label{eq:fwr_any_run_bound}
\end{align}

\end{proof}

\crefname{proposition}{Proposition}{Propositions}
\tikzset{
    buvtx/.style={circle,draw,minimum size=5.5mm,inner sep=0pt,font=\small},
    buvtxB/.style={buvtx,fill=black!20},
    bukept/.style={-latex,thick},
    bumerged/.style={-latex,thick,dashed,draw=blind_magenta},
    buleaf/.style={circle,draw,minimum size=2.2mm,inner sep=0pt},
    bucentre/.style={buleaf,fill=black},
    bunode/.style={circle,draw,fill=white,minimum size=2.6mm,inner sep=0pt},
    bujoin/.style={circle,draw=blind_magenta,fill=blind_magenta,minimum size=3mm,inner sep=0pt},
    bucell/.style={draw,minimum height=6mm,inner sep=2pt,font=\small},
}

\section{Proof that Binary Bottom-up Tokenisation is not in  \boldmath\apx\unboldmath}
\label{appendix:bu}

This appendix proves \cref{thm:bu_compression_not_apx} and \cref{thm:bu_loglik_not_apx}.
Both follow from a single reduction from gap maximum cut on cubic graphs, and from the gap it creates, which we state as a forward and a backward implication (\cref{prop:bu_gap}).
Taking this gap as given, \cref{appendix:bu_compression_proof} and \cref{appendix:bu_loglik_proof} first prove the two theorems.
The remaining subsections build up to \cref{prop:bu_gap}: \cref{sec:cubic_maxcut} recalls maximum cut on cubic graphs, \cref{appendix:bu_reduction} constructs the reduction, and \cref{appendix:bu_gap} states the gap.
Finally, \cref{appendix:bu_forward} turns a maximum cut into a short merge sequence that collapses the encoded string into a single token, and \cref{appendix:bu_backward} shows that no merge sequence over the binary alphabet can do better.

\subsection{Proof of \cref{thm:bu_compression_not_apx} (Compression)}
\label{appendix:bu_compression_proof}

\bucompressionnotinapx*

\begin{proof}
Suppose that there is a polynomial-time $\approxfactor$-approximation algorithm for some constant $\approxfactor\geq1$.
We use it to decide \threesat in polynomial time.
Given a \threesat formula $\sateq$, we compute $(\srcgraph,\cuttarget)$ by \cref{lemma:gap_maxcut}, and let $\vcnvertices_0\defeq\max\{9\nclones+1,\lceil2\approxfactor/(\gapconst\nclones^2)\rceil\}$.

\textit{Small graphs.}
If $\vcnvertices<\vcnvertices_0$, we compute $\maxcutof{\srcgraph}$ by enumerating all colourings, which takes constant time, and accept if and only if $\maxcutof{\srcgraph}\geq\cuttarget$.
This is correct by \cref{lemma:gap_maxcut}, as $\cuttarget-\gapconst\vcnvertices<\cuttarget$.

\textit{Large graphs.}
Otherwise, we compute $(\buenc{\srcword},\vocabsize)=\reductionfuncbufull$, run the algorithm on the dataset $\{\buenc{\srcword}\}$ with budget $\vocabsize$, and accept if and only if its merge sequence leaves at most $\approxfactor$ tokens.
If $\sateq$ is satisfiable, then $\maxcutof{\srcgraph}\geq\cuttarget$, so the optimal compressed length is $1$ by \cref{prop:bu_gap}\ref{item:bu_forward}, and the algorithm leaves at most $\approxfactor$ tokens.
If $\sateq$ is unsatisfiable, then $\maxcutof{\srcgraph}\leq\cuttarget-\gapconst\vcnvertices$, so by \cref{prop:bu_gap}\ref{item:bu_backward}, every feasible merge sequence leaves more than $\frac{\gapconst}{2}\nclones^2\vcnvertices\geq\approxfactor$ tokens.

All steps take polynomial time, so \threesat would be in \p, which contradicts its \np-hardness unless \pequalnp.

The same argument rules out much larger ratios.
The string $\buenc{\srcword}$ has length $N=\buenclen(4\nwalks+1)=\Theta(\vcnvertices\log\vcnvertices)$, and thus $\vcnvertices=\Theta(N/\log N)$.
Hence, for any ratio $\approxfactor(N)=o(N/\log N)$, we have $\approxfactor(N)<\frac{\gapconst}{2}\nclones^2\vcnvertices$ once $\vcnvertices$ exceeds some constant, below which we again enumerate all colourings.
Accepting if and only if the algorithm leaves at most $\approxfactor(N)$ tokens then decides \threesat as before.
\end{proof}

\subsection{Proof of \cref{thm:bu_loglik_not_apx} (Log-Likelihood)}
\label{appendix:bu_loglik_proof}
For the log-likelihood objective, we use the following lower bound.

\begin{lemma}
\label{lemma:bu_nll_floor}
If a tokenised dataset consists of $\totalfreq\geq2$ tokens of at least two distinct types, its unigram negative log-likelihood is at least $\log\totalfreq$.
\end{lemma}

\begin{proof}
Let $\subword^{\star}$ be a most frequent token, and let $g\defeq\totalfreq-\tokenfreq_{\subword^{\star}}$ count all other tokens.
As at least two types occur, $1\leq g\leq\totalfreq-1$, and every token $\subword\neq\subword^{\star}$ has $\tokenfreq_{\subword}\leq g$.
Writing $\mathrm{H}_2(x)\defeq-x\log x-(1-x)\log(1-x)$ for the binary entropy,
\begin{subequations}
\begin{align}
    \objectivefuncnll
    &=
    \tokenfreq_{\subword^{\star}}\log\frac{\totalfreq}{\tokenfreq_{\subword^{\star}}}
    +\sum_{\subword\neq\subword^{\star}}\tokenfreq_{\subword}\log\frac{\totalfreq}{\tokenfreq_{\subword}}
    \\
    &\geq
    (\totalfreq-g)\log\frac{\totalfreq}{\totalfreq-g}
    +g\log\frac{\totalfreq}{g}
    && \mathcomment{$\tokenfreq_{\subword}\leq g$ for $\subword\neq\subword^{\star}$}
    \\
    &=
    \totalfreq\cdot\mathrm{H}_2\Big(\frac{g}{\totalfreq}\Big)
    \geq
    \totalfreq\cdot\mathrm{H}_2\Big(\frac{1}{\totalfreq}\Big)
    && \mathcomment{$\mathrm{H}_2$ concave and symmetric}
    \\
    &=
    \log\totalfreq+(\totalfreq-1)\log\frac{\totalfreq}{\totalfreq-1}
    \geq
    \log\totalfreq.
\end{align}
\end{subequations}
The inequality in the third line holds since a concave function attains its minimum over the interval $[\frac{1}{\totalfreq},1-\frac{1}{\totalfreq}]$ at an endpoint, where $\mathrm{H}_2$ takes the same value.
\end{proof}
\bulogliknotinapx*

\begin{proof}
For an output $(\buenc{\srcword},\vocabsize)=\reductionfuncbufull$, consider the dataset $\dataset\defeq\{\buenc{\srcword},\symbolzero\}$ with budget $\vocabsize$.
No merge acts on the string $\symbolzero$, which is always the single token $\symbolzero$; and every tokenisation of $\buenc{\srcword}$ contains a token with a $\symbolone$, which differs from $\symbolzero$.
So every merge sequence produces $\totalfreq\geq2$ tokens of at least two types on $\dataset$, and by \cref{lemma:bu_nll_floor},
\begin{align}
    \objectivefuncnll(\bottomuptoken[\merges],\dataset)
    \geq
    \log\totalfreq
    \geq
    \log2.
    \label{eq:bu_nll_positive}
\end{align}
In particular, the optimal objective is strictly positive.
We now establish the gap.

\textit{\yes instances.} 
If $\maxcutof{\srcgraph}\geq\cuttarget$, the merge sequence of \cref{prop:bu_gap}\ref{item:bu_forward} tokenises $\dataset$ into the two distinct tokens $\buenc{\srcword}$ and $\symbolzero$, each with frequency $1$, so $\objectivefuncnllopt(\dataset)\leq2\log2$.

\textit{\no instances.}
If $\maxcutof{\srcgraph}\leq\cuttarget-\gapconst\vcnvertices$, every feasible merge sequence leaves more than $\frac{\gapconst}{2}\nclones^2\vcnvertices$ tokens on $\buenc{\srcword}$ by \cref{prop:bu_gap}\ref{item:bu_backward}, so $\totalfreq>\frac{\gapconst}{2}\nclones^2\vcnvertices$, and $\objectivefuncnllopt(\dataset)>\log(\frac{\gapconst}{2}\nclones^2\vcnvertices)$ by \cref{eq:bu_nll_positive}.

\textit{Deciding \threesat.}
Suppose that there is a polynomial-time $\approxfactor$-approximation algorithm for some constant $\approxfactor\geq1$.
Given a \threesat formula $\sateq$, we compute $(\srcgraph,\cuttarget)$ by \cref{lemma:gap_maxcut}, and let $\vcnvertices_0\defeq\max\{9\nclones+1,\lceil2\cdot4^{\approxfactor}/(\gapconst\nclones^2)\rceil\}$.
If $\vcnvertices<\vcnvertices_0$, we decide $\sateq$ by enumerating all colourings of $\srcgraph$, as in the proof of \cref{thm:bu_compression_not_apx}.
Otherwise, we run the algorithm on $\dataset$ with budget $\vocabsize$, count the number $\totalfreq$ of tokens its merge sequence produces, and accept if and only if $\totalfreq\leq4^{\approxfactor}$.
If $\sateq$ is satisfiable, the returned objective is at most $2\approxfactor\log2$, so $\log\totalfreq\leq2\approxfactor\log2$ by \cref{eq:bu_nll_positive}, i.e., $\totalfreq\leq4^{\approxfactor}$.
If $\sateq$ is unsatisfiable, every feasible merge sequence produces $\totalfreq>\frac{\gapconst}{2}\nclones^2\vcnvertices\geq4^{\approxfactor}$ tokens.
This test only compares integers, and all steps take polynomial time; so \threesat would be in \p, which contradicts its \np-hardness unless \pequalnp.

The same argument rules out ratios $\approxfactor(N)=o(\log N)$, where $N=|\buenc{\srcword}|+1$.
We accept if and only if $\totalfreq\leq4^{\approxfactor(N)}$; on \yes instances this holds as before, while on \no instances
\begin{align}
    \totalfreq
    >
    \frac{\gapconst}{2}\nclones^2\vcnvertices
    =
    \Theta\Big(\frac{N}{\log N}\Big)
    >
    N^{o(1)}
    =
    4^{\approxfactor(N)}
\end{align}
once $\vcnvertices$ exceeds some constant.
\end{proof}

\subsection{Maximum Cut on Cubic Graphs}
\label{sec:cubic_maxcut}
 
Let $\srcgraph=(\vertices,\edges)$ be a finite, simple, undirected graph with $\vcnvertices=|\vertices|$ vertices and $\vcmedges=|\edges|$ edges; we call it \defn{cubic} if every vertex has degree exactly $3$, in which case $\vcmedges=\frac{3}{2}\vcnvertices$ and $\vcnvertices\geq4$.
A \defn{colouring} $\colouring\colon\vertices\to\{\markB,\markW\}$ places each vertex on one of two sides; unlike in graph colouring, adjacent vertices may share a side.\footnote{We name the two sides after the two marker strings of \cref{red:unary_apx}; the reduction identifies a side with its marker.}
Its \defn{cut size} is the number of edges whose endpoints lie on different sides:
\begin{align}
    \cutsize{\srcgraph}{\colouring}
    \defeq
    \sum_{(\vertex,\vertex')\in\edges}
    \one_{\colouring}\{\colouring(\vertex)\neq\colouring(\vertex')\}.
\end{align}
 
\begin{defin}
\label{def:cubic_maxcut}
Let $\srcgraph=(\vertices,\edges)$ be a simple cubic graph.
Given a target $\cuttarget\in\N$, the \defn{\cubicmaxcut decision problem} requires deciding whether there exists a colouring $\colouring$ with $\cutsize{\srcgraph}{\colouring}\geq\cuttarget$.
The \defn{\cubicmaxcut optimisation problem} requires finding the maximum cut size.
Formally:
\begin{align}
    \underbrace{
        \cuttarget
        \stackrel{?}{\leq}
        \max_{\colouring\in\{\markB,\markW\}^{\vertices}}
        \cutsize{\srcgraph}{\colouring}
    }_{\text{\cubicmaxcut decision problem}}
    \qquad
    \underbrace{
        \maxcutof{\srcgraph}
        =
        \max_{\colouring\in\{\markB,\markW\}^{\vertices}}
        \cutsize{\srcgraph}{\colouring}
    }_{\text{\cubicmaxcut optimisation problem}}
\end{align}
\end{defin}

\citet{alimonti2000some} proved that \cubicmaxcut is \apx-complete.
We will also use the following standard facts.
 
\begin{myfact}
\label{fact:local_max_cut}
Given a simple cubic graph $\srcgraph$, a colouring $\colouring$ in which every vertex has at least two cut incident edges (a \defn{locally maximal} colouring) can be found in polynomial time.
Consequently,
\begin{align}
    \vcnvertices \leq \maxcutof{\srcgraph} \leq \vcmedges.
    \label{eq:cubic_cut_bounds}
\end{align}
\end{myfact}

\begin{proof}
Start from any colouring and repeatedly move a vertex to the other side while this increases the cut size.
Each move increases an integer bounded by $\vcmedges$, so at most $\vcmedges$ moves occur.
At termination, no vertex has more uncut than cut incident edges; as each vertex has degree $3$, it has at least two cut incident edges.
Counting cut edges from both endpoints gives $\cutsize{\srcgraph}{\colouring}\geq 2\vcnvertices/2=\vcnvertices$.
The upper bound is trivial.
\end{proof}
 
\begin{lemma}[Gap maximum cut]
\label{lemma:gap_maxcut}
There is a constant $\gapconst>0$ and a polynomial-time algorithm which maps every \threesat formula $\sateq$ to a pair $(\srcgraph,\cuttarget)$, where $\srcgraph$ is a simple cubic graph with $\vcnvertices$ vertices and $\cuttarget\leq\vcmedges$, such that
\begin{align}
    \sateq\text{ satisfiable}
    &\implies
    \maxcutof{\srcgraph}\geq\cuttarget,
    \\
    \sateq\text{ unsatisfiable}
    &\implies
    \maxcutof{\srcgraph}\leq\cuttarget-\gapconst\vcnvertices.
\end{align}
\end{lemma}
\begin{proof}
By the \pcp theorem and \citet{hastad2001optimal}, there are constants $0<s<c\leq1$ such that it is \np-hard to decide whether a \texttt{MAX-E2-LIN2} instance has a $c$-fraction or at most an $s$-fraction of its equations simultaneously satisfiable, and this hardness is witnessed by a polynomial-time reduction from \threesat.
\citet{berman2001efficient} give an explicit, deterministic gap-preserving reduction from \texttt{MAX-E2-LIN2} to maximum cut on simple $3$-regular graphs whose number of edges is linear in the number of equations.
Composing the two yields, for each formula $\sateq$, a simple cubic graph $\srcgraph$ and thresholds $\cuttarget_{\mathrm{yes}}>\cuttarget_{\mathrm{no}}$ with $\cuttarget_{\mathrm{yes}}-\cuttarget_{\mathrm{no}}=\Theta(\vcmedges)=\Theta(\vcnvertices)$, such that $\maxcutof{\srcgraph}\geq\cuttarget_{\mathrm{yes}}$ if $\sateq$ is satisfiable and $\maxcutof{\srcgraph}\leq\cuttarget_{\mathrm{no}}$ otherwise.
Setting $\cuttarget\defeq\cuttarget_{\mathrm{yes}}\leq\vcmedges$ and choosing $\gapconst$ such that $\cuttarget_{\mathrm{yes}}-\cuttarget_{\mathrm{no}}\geq\gapconst\vcnvertices$ proves the lemma.
\end{proof}

\begin{defin}
\label{def:blowup}
Let $\srcgraph=(\vertices,\edges)$ be a graph and $t\geq1$ an integer.
Its \defn{$t$-blow-up} $\srcgraph_t=(\vertices_t,\edges_t)$ has vertices $\vertices_t\defeq\vertices\times\{1,\ldots,t\}$ and edges
\begin{align}
    \edges_t
    \defeq
    \big\{
        \{(\vertex,k),(\vertex',k')\}
        \mid
        \{\vertex,\vertex'\}\in\edges,\quad 1\leq k,k'\leq t
    \big\}.
\end{align}
\end{defin}

Each vertex thus becomes a \defn{cluster} of $t$ clones, and each edge a complete bipartite graph between two clusters.
If $\srcgraph$ is cubic, $\srcgraph_t$ is a simple $3t$-regular graph with $t\vcnvertices$ vertices and $t^2\vcmedges$ edges (\cref{fig:blowup}).
\begin{figure}[t]
\centering
\begin{tikzpicture}[
    blowclone/.style={circle,draw,fill=white,minimum size=3mm,inner sep=0pt},
    blowcluster/.style={draw=black!35,fill=black!4,rounded corners=2.6mm,line width=0.4pt},
  ]
\colorlet{blowab}{blind_blue}
  \colorlet{blowbc}{blind_magenta}
  \colorlet{blowca}{blind_orange}

\node[buvtx] (a) at (90:0.95)  {$a$};
  \node[buvtx] (b) at (210:0.95) {$b$};
  \node[buvtx] (c) at (330:0.95) {$c$};
  \draw[blowab,thick] (a) -- (b);
  \draw[blowbc,thick] (b) -- (c);
  \draw[blowca,thick] (c) -- (a);
  \node[font=\small] at (0,-1.7) {$\srcgraph$};

  \draw[-latex,thick] (1.55,0.1) -- node[above,font=\small] {$t=3$} (2.75,0.1);

\begin{scope}[xshift=5.15cm]
    \def\R{1.7}  \def\d{0.42} \foreach \v/\ang in {a/90, b/210, c/330} {
      \begin{scope}[shift={(\ang:\R)}, rotate=\ang+90]
        \draw[blowcluster] (-0.72,-0.27) rectangle (0.72,0.27);
        \coordinate (\v1) at (-\d,0);
        \coordinate (\v2) at (0,0);
        \coordinate (\v3) at (\d,0);
      \end{scope}
      \node[font=\small] at ({\ang}:{\R+0.6}) {$\v$};
    }
    \foreach \u/\w/\col in {a/b/blowab, b/c/blowbc, c/a/blowca} {
      \foreach \i in {1,2,3} {
        \foreach \j in {1,2,3} {
          \draw[\col, line width=0.45pt] (\u\i) -- (\w\j);
        }
      }
    }
    \foreach \v in {a,b,c} {
      \foreach \k in {1,2,3} {
        \node[blowclone] at (\v\k) {};
      }
    }
    \node[font=\small] at (0,-1.7) {$\srcgraph_3$};
  \end{scope}
\end{tikzpicture}
\caption{
The $3$-blow-up of a triangle (\cref{def:blowup}).
Each edge becomes a complete bipartite graph between two clusters, drawn in the edge's colour.
}
\label{fig:blowup}
\end{figure}
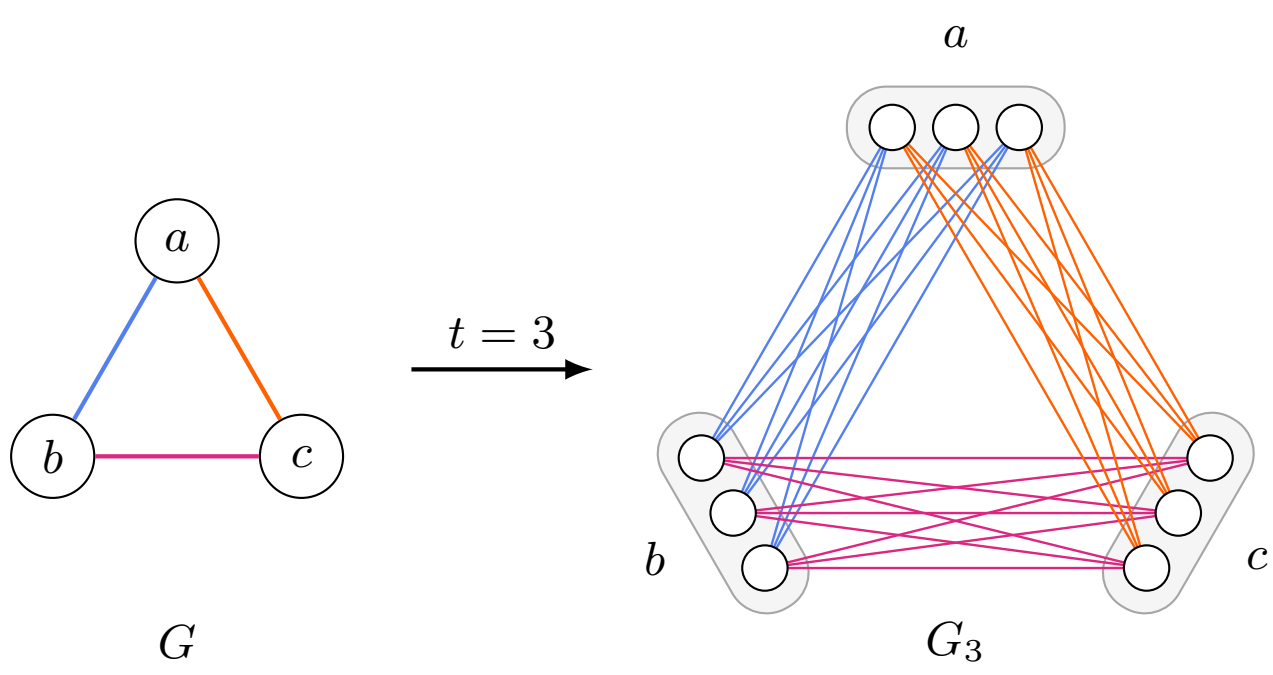 
\begin{lemma}
\label{lemma:blowup}
Let $\srcgraph$ be a simple graph and $t\geq1$.
\begin{enumerate}[label=(\alph*)]
    \item \label{item:blowup_induced}
    A colouring $\srccolouring$ of $\srcgraph$ induces the colouring $\colouring$ of $\srcgraph_t$ which gives every clone $(\vertex,k)$ the colour $\srccolouring(\vertex)$, and $\cutsize{\srcgraph_t}{\colouring}=t^2\cutsize{\srcgraph}{\srccolouring}$.
    \item \label{item:blowup_max}
    $\maxcutof{\srcgraph_t}=t^2\maxcutof{\srcgraph}$.
\end{enumerate}
\end{lemma}

\begin{proof}
Part \ref{item:blowup_induced}: every cut edge of $\srcgraph$ yields $t^2$ cut edges of $\srcgraph_t$, and no other edge is cut.
Part \ref{item:blowup_max}: part \ref{item:blowup_induced} gives $\maxcutof{\srcgraph_t}\geq t^2\maxcutof{\srcgraph}$.
Conversely, let $\colouring$ be any colouring of $\srcgraph_t$, and let $x_{\vertex}\in\{0,\ldots,t\}$ be the number of clones of $\vertex$ coloured $\markB$.
Then
\begin{align}
    \cutsize{\srcgraph_t}{\colouring}
    =
    \sum_{\{\vertex,\vertex'\}\in\edges}
    \big(x_{\vertex}(t-x_{\vertex'})+(t-x_{\vertex})x_{\vertex'}\big).
\end{align}
As $\srcgraph$ has no loops, this sum is affine in each $x_{\vertex}$ when all other counts are fixed, and is thus maximised at $x_{\vertex}=0$ or $x_{\vertex}=t$.
Recolouring all clones of each vertex in turn with the better of the two colours therefore never decreases the cut, and ends in a colouring induced by some colouring $\srccolouring$ of $\srcgraph$.
Hence $\cutsize{\srcgraph_t}{\colouring}\leq t^2\cutsize{\srcgraph}{\srccolouring}\leq t^2\maxcutof{\srcgraph}$.
\end{proof}

\subsection{The Reduction}
\label{appendix:bu_reduction}

We reduce from the gap version of \cubicmaxcut in \cref{lemma:gap_maxcut}, using the notation of \cref{sec:cubic_maxcut}: $\srcgraph=(\vertices,\edges)$ is a simple cubic graph with $\vcnvertices$ vertices and $\vcmedges=\frac{3}{2}\vcnvertices$ edges, and $\maxcutof{\srcgraph}$ is the size of its maximum cut.
Given an instance $(\srcgraph,\cuttarget)$, our reduction proceeds in three stages: it blows up $\srcgraph$, writes down a source string over a large alphabet, and encodes this string in binary.

\paragraph{Stage 1: blowing up the graph.}
   Fix the constant $\nclones\defeq\lceil40/\gapconst\rceil$, and let $\blowgraphl=(\blowverticesl,\blowedgesl)$ be the $\nclones$-blow-up of $\srcgraph$ (\cref{def:blowup}).
   It is a simple $3\nclones$-regular graph with $\nclones\vcnvertices$ vertices, and $\maxcutof{\blowgraphl}=\nclones^2\maxcutof{\srcgraph}$ by \cref{lemma:blowup}\ref{item:blowup_max}. This is the graph $\blowgraphl$ of \cref{red:bu}, whose copy $\vertex_{j,n}$ is the clone $(\vertex_j,n)$.

\begin{figure}[t]
\centering
\begin{subfigure}[b]{0.24\textwidth}
\centering
\begin{tikzpicture}
    \node[buvtx] (a) at (0,0) {$a$};
    \node[buvtx] (b) at (1.6,0) {$b$};
    \node[buvtx] (c) at (3.2,0) {$c$};
    \draw[bukept] (a) to[bend left=30] (b);
    \draw[bukept] (b) to[bend left=30] (a);
    \draw[bukept] (b) to[bend left=30] (c);
    \draw[bukept] (c) to[bend left=30] (b);
    \node[font=\scriptsize] at (1.6,-0.85) {\phantom{$\markB$}};
\end{tikzpicture}
\caption{Arcs of the graph.}
\label{fig:bu_example_arcs}
\end{subfigure}
\hfill
\begin{subfigure}[b]{0.24\textwidth}
\centering
\begin{tikzpicture}
    \node[buvtx] (a) at (0,0) {$a$};
    \node[buvtxB] (b) at (1.6,0) {$b$};
    \node[buvtx] (c) at (3.2,0) {$c$};
    \draw[bumerged] (a) to[bend left=30] (b);
    \draw[bukept] (b) to[bend left=30] (a);
    \draw[bukept] (b) to[bend left=30] (c);
    \draw[bumerged] (c) to[bend left=30] (b);
    \node[font=\scriptsize] at (0,-0.85) {$\markW$};
    \node[font=\scriptsize] at (1.6,-0.85) {$\markB$};
    \node[font=\scriptsize] at (3.2,-0.85) {$\markW$};
\end{tikzpicture}
\caption{A maximum cut.}
\label{fig:bu_example_cut}
\end{subfigure}
\hfill
\begin{subfigure}[b]{0.24\textwidth}
\centering
\footnotesize
\begin{tabular}{@{}c@{\quad}c@{}}
    walk & after phase 1 \\
    \midrule
    $\charstring{aba}$ & $\subwordstring{\langle ab,a\rangle}$ \\
    $\charstring{abc}$ & $\subwordstring{\langle ab,c\rangle}$ \\
    $\charstring{cba}$ & $\subwordstring{\langle cb,a\rangle}$ \\
    $\charstring{cbc}$ & $\subwordstring{\langle cb,c\rangle}$ \\
    $\charstring{bab}$ & $\subwordstring{\langle b,ab\rangle}$ \\
    $\charstring{bcb}$ & $\subwordstring{\langle b,cb\rangle}$ \\
\end{tabular}
\caption{Merging $\mergestring{a}{b}$, $\mergestring{c}{b}$.}
\label{fig:bu_example_walks}
\end{subfigure}
\hfill
\begin{subfigure}[b]{0.24\textwidth}
\centering
\begin{tikzpicture}
    \node[buvtx] (a) at (0,0) {$a$};
    \node[buvtx,draw=blind_magenta,very thick] (b) at (1.6,0) {$b$};
    \node[buvtx] (c) at (3.2,0) {$c$};
    \draw[bukept] (a) to[bend left=30] (b);
    \draw[bumerged] (b) to[bend left=30] (a);
    \draw[bukept] (b) to[bend left=30] (c);
    \draw[bumerged] (c) to[bend left=30] (b);
    \node[font=\scriptsize] at (1.6,-0.85) {\phantom{$\markB$}};
\end{tikzpicture}
\caption{Not a cut.}
\label{fig:bu_example_bad}
\end{subfigure}
\caption{
The source string of \cref{eq:bu_source_string}, for the path $a$--$b$--$c$ and $\nclones=1$ (for illustration only, as this graph is not cubic).
\textbf{(a)}~The graph has four arcs and six walks, $\charstring{aba}$, $\charstring{abc}$, $\charstring{cba}$, $\charstring{cbc}$, $\charstring{bab}$, and $\charstring{bcb}$; with distinct separators, its source string is $\srcword=\charstring{aba}\sepsym{1}\charstring{abc}\sepsym{2}\charstring{cba}\sepsym{3}\charstring{cbc}\sepsym{4}\charstring{bab}\sepsym{5}\charstring{bcb}\sepsym{6}\buendsym$, of length $4\nwalks+1=25$.
\textbf{(b)}~A maximum cut, with $b$ coloured $\markB$ and $a,c$ coloured $\markW$. The two arcs from $\markB$ to $\markW$ are kept (solid); the other two arcs, $(a,b)$ and $(c,b)$, are merged (dashed) and form the covering set $\bigramset^{\star}$ of \cref{lemma:bu_cover}.
\textbf{(c)}~Merging these two bigrams shortens every walk by one token (phase 1 of \cref{lemma:bu_collapse}). The remaining $3\nwalks+1=19$ tokens are then joined by $18$ merges (phase 2), so collapsing $\srcword$ takes $\collapsecost=2+18=20$ merges, as given by \cref{eq:bu_collapse_cost}.
\textbf{(d)}~Keeping arcs which enter and leave the same vertex (here, $(a,b)$ and $(b,c)$ at $b$) leaves a walk with no merged bigram (here, $\charstring{abc}$): the kept arcs of a covering set must form a cut.
}
\label{fig:bu_example}
\end{figure}
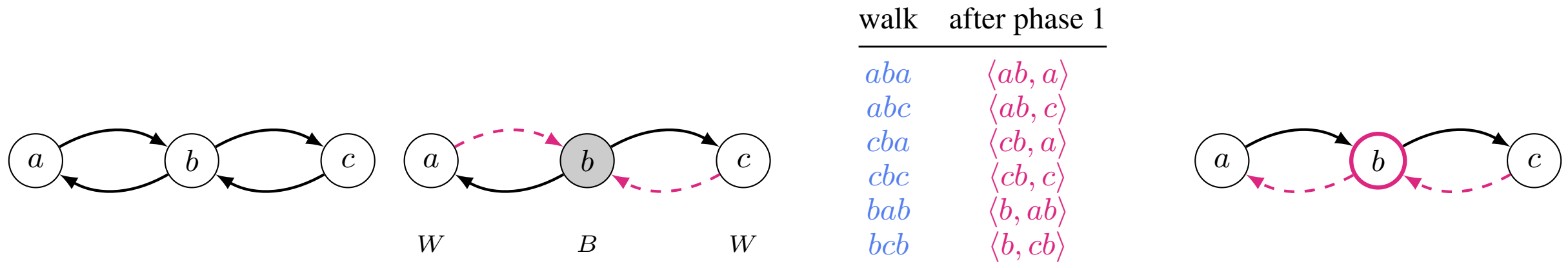

\paragraph{Stage 2: a source string.}
An \defn{arc} is an edge of $\blowgraphl$ with a direction; the set of arcs is
\begin{align}
    \arcs
    \defeq
    \{(\vertex,\vertex')\mid\{\vertex,\vertex'\}\in\blowedgesl\},
    \qquad
    |\arcs|=3\nclones^2\vcnvertices.
\end{align}
A \defn{walk} is a string $\walk=\vertex\vertex'\vertex''$ of three vertices such that $(\vertex,\vertex')$ and $(\vertex',\vertex'')$ are arcs, where $\vertex=\vertex''$ is allowed; we call these two arcs the \defn{bigrams} of $\walk$.
Let $\walks\defeq\walkfunc(\blowgraphl)$ be the set of walks.
A walk is determined by its middle vertex and that vertex's two neighbours in the walk, so
\begin{align}
    \nwalks\defeq|\walks|=\nclones\vcnvertices\cdot(3\nclones)^2=9\nclones^3\vcnvertices.
\end{align}
We list the walks in an arbitrary order $\walk_1,\ldots,\walk_{\nwalks}$ and define the \defn{source string} $\srcword\defeq\graphtostringfunc(\blowgraphl)$ as
\begin{align}
    \graphtostringfunc(\blowgraphl)
    \defeq
    \walk_1\sepsym{\chi_1}\walk_2\sepsym{\chi_2}\cdots\walk_{\nwalks}\sepsym{\chi_{\nwalks}}\buendsym,
    \label{eq:bu_source_string}
\end{align}
in which every walk is followed by a \defn{separator} $\sepsym{\chi_i}$, with $\chi_i\in\{1,\ldots,9\nclones^2\}$, and which ends with an end symbol $\buendsym$.
We choose the separators such that walks ending in the same vertex are followed by different separators, and walks starting with the same vertex are preceded by different separators.\footnote{Such separators exist and can be found in polynomial time. Consider the bipartite multigraph with one edge for each $i\in\{1,\ldots,\nwalks\}$, joining the last vertex of $\walk_i$, as a left vertex, to the first vertex of $\walk_{i+1}$, as a right vertex (or to a dummy right vertex, for $i=\nwalks$). Every vertex ends exactly $9\nclones^2$ walks and starts exactly $9\nclones^2$ walks, so this multigraph has maximum degree $9\nclones^2$. By K\H{o}nig's edge-colouring theorem, its edges can be properly coloured with $9\nclones^2$ colours, and such a colouring can be computed in polynomial time; we let $\chi_i$ be the colour of edge $i$.}
We write $\srcalphabet$ for the set of symbols occurring in $\srcword$: the $\nclones\vcnvertices$ vertices, at most $9\nclones^2$ separators, and $\buendsym$.
The source string has length $|\srcword|=4\nwalks+1$, and \cref{fig:bu_example} shows an example.
The separators ensure that only walk bigrams repeat in $\srcword$.

\begin{lemma}[Uniqueness]
\label{lemma:bu_unique}
The source string $\srcword$ satisfies the following:
\begin{enumerate}[label=(\alph*)]
    \item \label{item:bu_unique_sep}
    every bigram of $\srcword$ containing a separator or $\buendsym$ occurs only once in $\srcword$;
    \item \label{item:bu_unique_long}
    every substring of $\srcword$ of length at least $3$ occurs only once in $\srcword$;
    \item \label{item:bu_unique_adj}
    any two adjacent vertices in $\srcword$ lie in the same walk, and thus form one of its bigrams.
\end{enumerate}
\end{lemma}

\begin{proof}
Part \ref{item:bu_unique_adj} holds since consecutive walks are separated by a separator.
For part \ref{item:bu_unique_sep}, note that separators are never adjacent to each other, and that $\buendsym$ occurs only once.
A bigram consisting of a vertex $\vertex$ followed by a separator $\sepsym{\chi}$ marks the end of a walk which ends in $\vertex$; by our choice of separators, no other walk ending in $\vertex$ is followed by $\sepsym{\chi}$, so this bigram occurs only once.
The same holds, symmetrically, for a separator followed by a vertex.
For part \ref{item:bu_unique_long}, it suffices to consider substrings of length exactly $3$, as longer substrings contain one.
A substring of length $3$ containing a separator or $\buendsym$ contains a bigram from part \ref{item:bu_unique_sep}, and thus occurs only once.
Otherwise, its three symbols are vertices, which by part \ref{item:bu_unique_adj} lie in the same walk; as walks have length $3$, the substring is a walk, and walks are pairwise distinct.
\end{proof}

\begin{figure}[t]
\centering
\begin{subfigure}[b]{0.56\textwidth}
\centering
\begin{tikzpicture}[bucell/.append style={minimum height=6.5mm,inner sep=2.5pt}]
    \node[bucell] (a1) {$\symbolone$};
    \node[bucell,right=0pt of a1] (a2) {$\symbolone$};
    \node[bucell,right=0pt of a2] (a3) {$\symbolzero\fwrbit_1$};
    \node[bucell,right=0pt of a3] (a4) {$\cdots$};
    \node[bucell,right=0pt of a4] (a5) {$\symbolzero\fwrbit_{\bucodelength}$};
    \node[bucell,right=0pt of a5] (a6) {$\symbolzero$};
    \node[bucell,right=2pt of a6,fill=black!15] (b1) {$\symbolone$};
    \node[bucell,right=0pt of b1] (b2) {$\symbolone$};
    \node[bucell,right=0pt of b2] (b3) {$\symbolzero\fwrbit_1$};
    \node[bucell,right=0pt of b3] (b4) {$\cdots$};
    \node[bucell,right=0pt of b4] (b5) {$\symbolzero\fwrbit_{\bucodelength}$};
    \node[bucell,right=0pt of b5] (b6) {$\symbolzero$};
    \node[bucell,right=2pt of b6] (c1) {$\symbolzero$};
    \draw[decorate,decoration={brace,amplitude=4pt,mirror}]
        ([yshift=-2pt]a1.south west) -- ([yshift=-2pt]a6.south east)
        node[midway,below=5pt,font=\small] {$\buhalf{a}$, length $\buhalflen$};
    \draw[decorate,decoration={brace,amplitude=4pt,mirror}]
        ([yshift=-2pt]b1.south west) -- ([yshift=-2pt]b6.south east)
        node[midway,below=5pt,font=\small] {$\buhalf{a}$, length $\buhalflen$};
    \draw[decorate,decoration={brace,amplitude=4pt,mirror}]
        ([yshift=-24pt]a1.south west) -- ([yshift=-24pt]c1.south east)
        node[midway,below=5pt,font=\small] {$\buenc{a}$, length $\buenclen=2\buhalflen+1$};
    \draw[latex-,thick] ([yshift=2pt]b1.north) -- ++(0,0.55) node[above,font=\small] {centre};
\end{tikzpicture}
\caption{A codeword.}
\label{fig:bu_codeword_single}
\end{subfigure}
\hfill
\begin{subfigure}[b]{0.42\textwidth}
\centering
\begin{tikzpicture}[bucell/.append style={minimum height=6.5mm,inner sep=2.5pt}]
    \node[bucell,minimum width=12mm] (a1) {$\buhalf{a}$};
    \node[bucell,minimum width=12mm,right=0pt of a1] (a2) {$\buhalf{a}$};
    \node[bucell,right=0pt of a2] (a3) {$\symbolzero$};
    \node[bucell,minimum width=12mm,right=2pt of a3] (b1) {$\buhalf{b}$};
    \node[bucell,minimum width=12mm,right=0pt of b1] (b2) {$\buhalf{b}$};
    \node[bucell,right=0pt of b2] (b3) {$\symbolzero$};
    \fill[black] ([xshift=3pt,yshift=-3pt]a2.north west) circle (1.6pt);
    \fill[black] ([xshift=3pt,yshift=-3pt]b2.north west) circle (1.6pt);
    \draw[decorate,decoration={brace,amplitude=4pt}]
        ([yshift=2pt]a1.north west) -- ([yshift=2pt]a3.north east)
        node[midway,above=5pt,font=\small] {$\buenc{a}$};
    \draw[decorate,decoration={brace,amplitude=4pt}]
        ([yshift=2pt]b1.north west) -- ([yshift=2pt]b3.north east)
        node[midway,above=5pt,font=\small] {$\buenc{b}$};
    \draw[decorate,decoration={brace,amplitude=4pt,mirror}]
        ([yshift=-2pt]a2.south west) -- ([yshift=-2pt]b2.south west)
        node[midway,below=5pt,font=\small] {$\buhalf{a}\circ\symbolzero\circ\buhalf{b}$, length $\buenclen$};
\end{tikzpicture}
\caption{Two consecutive codewords.}
\label{fig:bu_codeword_pair}
\end{subfigure}
\caption{
The binary encoding of \cref{eq:bu_codeword}.
\textbf{(a)}~The pattern $\symbolone\symbolone$ starts each half-word, and the centre (shaded) starts the second one.
\textbf{(b)}~Between two consecutive centres (dots) lies $\buhalf{a}\circ\symbolzero\circ\buhalf{b}$, so a token containing both reveals both symbols (\cref{lemma:bu_recognition}).
}
\label{fig:bu_codeword}
\end{figure}
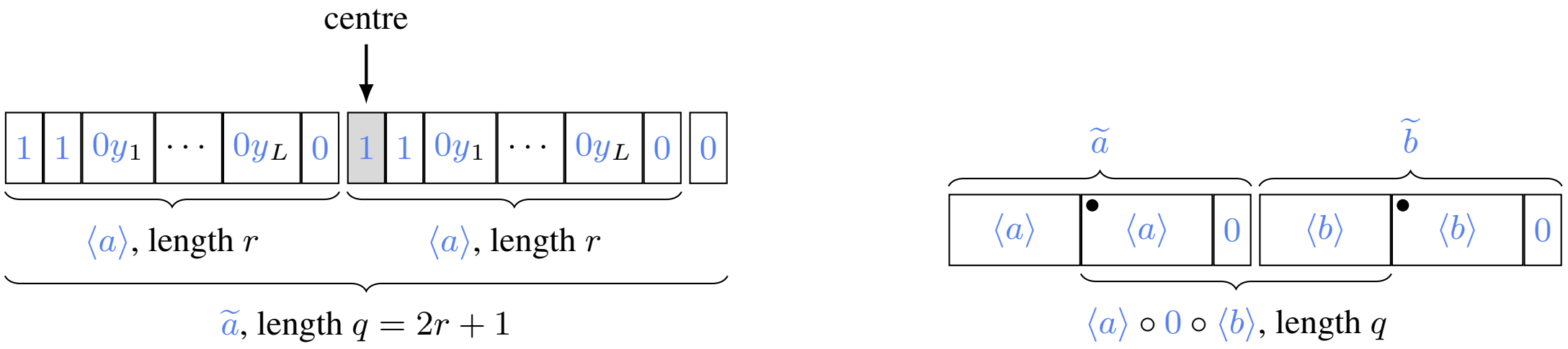

\paragraph{Stage 3: a binary encoding.}
Let $\bucodelength\defeq\lceil\log_2|\srcalphabet|\rceil$, and give every symbol $a\in\srcalphabet$ a distinct identifier $\idfunc(a)=\fwrbit_1\cdots\fwrbit_{\bucodelength}\in\{\symbolzero,\symbolone\}^{\bucodelength}$.
We encode $a$ by its \defn{codeword} $\binaryfunc(a)$, which we write $\buenc{a}$ for short; it consists of two copies of a \defn{half-word} $\buhalf{a}$, followed by a $\symbolzero$:
\begin{align}
    \buhalf{a}
    \defeq
    \symbolone\symbolone
    \symbolzero\fwrbit_1
    \symbolzero\fwrbit_2
    \cdots
    \symbolzero\fwrbit_{\bucodelength}
    \symbolzero,
    \qquad
    \buenc{a}
    \defeq
    \buhalf{a}\circ\buhalf{a}\circ\symbolzero.
    \label{eq:bu_codeword}
\end{align}
We encode a string $\characters=a_1\cdots a_m\in\srcalphabet^+$ codeword by codeword, as $\buenc{\characters}\defeq\BigCirc_{i=1}^{m}\binaryfunc(a_i)=\buenc{a_1}\circ\cdots\circ\buenc{a_m}$.
Half-words have length $\buhalflen\defeq2\bucodelength+3$, and codewords length $\buenclen\defeq2\buhalflen+1$.
As in \cref{sec:fwr_binary_encoding}, the pattern $\symbolone\symbolone$ occurs in an encoded string exactly at the starts of half-words.
We call the first character of the second half-word of a codeword its \defn{centre} (\cref{fig:bu_codeword}).
Centres play no role in the reduction itself, but are our main tool to analyse it in \cref{appendix:bu_backward}.

\begin{reduction}[Full version of \cref{red:bu}]
\label{red:bu_not_apx}
Given an instance $(\srcgraph,\cuttarget)$ of \cref{lemma:gap_maxcut}, construct the source string $\srcword$ as above, and output the dataset $\dataset\defeq\{\buenc{\srcword}\}$ over the alphabet $\alphabet\defeq\{\symbolzero,\symbolone\}$, together with the budget
\begin{align}
    \vocabsize
    \defeq
    \nprep+3\nwalks+|\arcs|-\nclones^2\cuttarget,
    \label{eq:bu_budget}
\end{align}
where $\nprep\defeq|\mergesprep|$ is the number of merges in the sequence $\mergesprep$ of \cref{lemma:bu_prep}, which builds the codewords.
\end{reduction}

We write $\reductionfuncbufull$ for the output $(\buenc{\srcword},\vocabsize)$.
Since $\cuttarget\leq\vcmedges$ (\cref{lemma:gap_maxcut}), we have $\nclones^2\cuttarget\leq|\arcs|$, and thus $\vocabsize>0$.
As $\nclones$ is a constant, $|\srcalphabet|=O(\vcnvertices)$, $\bucodelength=O(\log\vcnvertices)$, and $|\buenc{\srcword}|=\buenclen(4\nwalks+1)=O(\vcnvertices\log\vcnvertices)$; all three stages run in polynomial time.

\subsection{The Gap}
\label{appendix:bu_gap}

Both implications revolve around the \defn{collapse cost}
\begin{align}
    \collapsecost
    \defeq
    3\nwalks+|\arcs|-\nclones^2\maxcutof{\srcgraph},
    \label{eq:bu_collapse_cost}
\end{align}
the number of merges needed to tokenise the source string $\srcword$ as a single token.
Collapsing $\srcword$ naively takes $|\srcword|-1=4\nwalks$ merges, each making a single replacement.
Merging a walk bigram is cheaper, as one such merge shortens many walks at once; a maximum cut tells us which bigrams to merge, saving $\nwalks-|\arcs|+\nclones^2\maxcutof{\srcgraph}$ merges in total (\cref{fig:bu_example}).
Conversely, no merge sequence does better, even over the binary alphabet: a merge sequence which is $x$ merges short of $\collapsecost$ leaves at least $x+1$ tokens.
The budget $\vocabsize$ pays for building the codewords, plus the collapse cost of a graph with maximum cut $\cuttarget$.
In a \yes instance, the collapse cost thus fits in the budget; in a \no instance, it exceeds the budget by at least $\gapconst\nclones^2\vcnvertices-\nprep$, which is more than $\frac{\gapconst}{2}\nclones^2\vcnvertices$ by our choice of $\nclones$.

\begin{proposition}
\label{prop:bu_gap}
Let $(\buenc{\srcword},\vocabsize)=\reductionfuncbufull$, where $\srcgraph$ has $\vcnvertices\geq9\nclones+1$ vertices.
\begin{enumerate}[label=(\roman*)]
    \item \label{item:bu_forward}
    \textbf{Forward implication.}
    If $\maxcutof{\srcgraph}\geq\cuttarget$, then some merge sequence $\merges$ with $|\merges|\leq\vocabsize$ tokenises $\buenc{\srcword}$ as a single token.
    \item \label{item:bu_backward}
    \textbf{Backward implication.}
    If some merge sequence $\merges$ with $|\merges|\leq\vocabsize$ leaves at most $\frac{\gapconst}{2}\nclones^2\vcnvertices$ tokens on $\buenc{\srcword}$, then $\maxcutof{\srcgraph}>\cuttarget-\gapconst\vcnvertices$.
\end{enumerate}
\end{proposition}

In particular, if $\maxcutof{\srcgraph}\geq\cuttarget$, the optimal compressed length of $\{\buenc{\srcword}\}$ is $1$; and if $\maxcutof{\srcgraph}\leq\cuttarget-\gapconst\vcnvertices$, every merge sequence with $|\merges|\leq\vocabsize$ leaves more than $\frac{\gapconst}{2}\nclones^2\vcnvertices$ tokens on $\buenc{\srcword}$.

\subsection{Proof of the Forward Implication}
\label{appendix:bu_forward}

We now build a merge sequence which collapses $\buenc{\srcword}$ into a single token.
We proceed in three steps: we relate maximum cuts to sets of bigrams which shorten every walk (\cref{lemma:bu_cover}); we use such a set to collapse $\srcword$ with $\collapsecost$ merges (\cref{lemma:bu_collapse}); and we transfer this collapse to $\buenc{\srcword}$ (\cref{lemma:bu_prep,lemma:bu_simulate}).

We say that a set of arcs $\bigramset\subseteq\arcs$ \defn{covers} a walk if it contains one of the walk's bigrams, and we write $\covfunc(\bigramset)$ for the number of walks covered by $\bigramset$.
Merging the bigrams in $\bigramset$ shortens each covered walk by one token (\cref{lemma:bu_collapse}).
Part (b) of the following lemma is only needed for the backward implication.

\begin{lemma}[Covering walks]
\label{lemma:bu_cover}
\begin{enumerate}[label=(\alph*)]
    \item \label{item:bu_cover_all}
    Every set of arcs $\bigramset$ which covers all walks has $|\bigramset|\geq|\arcs|-\nclones^2\maxcutof{\srcgraph}$, and some such set $\bigramset^{\star}$ attains equality.
    \item \label{item:bu_cover_some}
    Every set of arcs $\bigramset\subseteq\arcs$ satisfies $\covfunc(\bigramset)-|\bigramset|\leq\nwalks-|\arcs|+\nclones^2\maxcutof{\srcgraph}$.
\end{enumerate}
\end{lemma}

\begin{proof}
We prove the two parts separately.

\textit{Part \ref{item:bu_cover_all}, lower bound.}
Let $\bigramset$ cover all walks, and call the arcs not in $\bigramset$ \defn{kept}.
No vertex $\vertex'$ has both a kept arc $(\vertex,\vertex')$ entering it and a kept arc $(\vertex',\vertex'')$ leaving it, as the walk $\vertex\vertex'\vertex''$ would then not be covered (\cref{fig:bu_example_bad}).
Now colour every vertex with a kept arc leaving it $\markB$, and all other vertices $\markW$.
Every kept arc $(\vertex,\vertex')$ then crosses this cut from $\markB$ to $\markW$: $\vertex$ is coloured $\markB$ since $(\vertex,\vertex')$ leaves it, and $\vertex'$ is coloured $\markW$ since $(\vertex,\vertex')$ enters it, so no kept arc leaves it.
Moreover, two kept arcs never lie on the same edge, as $(\vertex,\vertex')$ and $(\vertex',\vertex)$ would both enter and leave $\vertex'$.
The kept arcs thus lie on distinct cut edges, and by \cref{lemma:blowup},
\begin{align}
    |\arcs|-|\bigramset|
    \leq
    \maxcutof{\blowgraphl}
    =
    \nclones^2\maxcutof{\srcgraph}.
\end{align}

\textit{Part \ref{item:bu_cover_all}, equality.}
Take a colouring of $\blowgraphl$ with maximum cut, and let $\bigramset^{\star}$ contain all arcs except those from a $\markB$ vertex to a $\markW$ vertex (\cref{fig:bu_example_cut}).
Then $|\arcs|-|\bigramset^{\star}|=\maxcutof{\blowgraphl}=\nclones^2\maxcutof{\srcgraph}$.
Moreover, $\bigramset^{\star}$ covers every walk $\vertex\vertex'\vertex''$: otherwise, both $(\vertex,\vertex')$ and $(\vertex',\vertex'')$ would go from a $\markB$ vertex to a $\markW$ vertex, and $\vertex'$ would be coloured both $\markW$ and $\markB$.

\textit{Part \ref{item:bu_cover_some}.}
Given $\bigramset$, add to it one bigram of each walk which it does not cover.
The resulting set covers all walks and has at most $|\bigramset|+\nwalks-\covfunc(\bigramset)$ arcs, so by part \ref{item:bu_cover_all},
\begin{align}
    |\arcs|-\nclones^2\maxcutof{\srcgraph}
    \leq
    |\bigramset|+\nwalks-\covfunc(\bigramset).
\end{align}
Rearranging gives part \ref{item:bu_cover_some}.
\end{proof}

\begin{lemma}[Collapsing the source string]
\label{lemma:bu_collapse}
There is a merge sequence $\mergessrc$ with $|\mergessrc|=\collapsecost$ which tokenises $\srcword$ as a single token.
\end{lemma}

\begin{proof}
Let $\bigramset^{\star}$ be the covering set of \cref{lemma:bu_cover}\ref{item:bu_cover_all}, with $|\bigramset^{\star}|=|\arcs|-\nclones^2\maxcutof{\srcgraph}$.
We build $\mergessrc$ in two phases.

\textit{Phase 1: merging bigrams (\cref{fig:bu_example_walks}).}
For every arc $(\vertex,\vertex')\in\bigramset^{\star}$, in arbitrary order, we merge $\mergestring{\vertex}{\vertex'}$.
By \cref{lemma:bu_unique}\ref{item:bu_unique_adj}, two adjacent vertices only occur inside walks, and since $\vertex\neq\vertex'$, the occurrences of the pair never overlap.
So a merge $\mergestring{\vertex}{\vertex'}$ only changes the walks which have $(\vertex,\vertex')$ as a bigram.
Once one bigram of a walk is merged, the walk's middle vertex lies inside a token of two symbols, and no later merge of this phase changes the walk.
As $\bigramset^{\star}$ covers all walks, every walk is shortened by exactly one token.
This phase uses $|\bigramset^{\star}|$ merges and leaves $|\srcword|-\nwalks=3\nwalks+1$ tokens, the last of which is $\buendsym$.

\textit{Phase 2: merging from the right.}
Let $\subword_1,\ldots,\subword_{3\nwalks+1}$ be the current tokens.
For $i=3\nwalks,\ldots,1$, in this order, we merge $\mergestring{\subword_i}{\subword_{i+1}\circ\cdots\circ\subword_{3\nwalks+1}}$.
The right token of this merge contains $\buendsym$ and thus occurs only once, so the merge combines $\subword_i$ with the token to its right.
After these $3\nwalks$ merges, a single token remains.

In total, $|\mergessrc|=|\bigramset^{\star}|+3\nwalks=\collapsecost$.
\end{proof}

It remains to transfer this collapse to the binary string.
We first build every codeword as a token, using few merges.

\begin{lemma}[Building the codewords]
\label{lemma:bu_prep}
There is a merge sequence $\mergesprep$ with $\nprep=|\mergesprep|<10|\srcalphabet|$ which, for every string $\characters=a_1\cdots a_m\in\srcalphabet^+$, gives
\begin{align}
    \bottomuptoken[\mergesprep](\buenc{\characters})
    =
    \langle\buenc{a_1},\ldots,\buenc{a_m}\rangle.
    \label{eq:bu_prep_output}
\end{align}
\end{lemma}

\begin{proof}
The sequence $\mergesprep$ consists of three stages:
\begin{enumerate}[label=(\roman*)]
    \item \emph{markers}: merge $\mergestring{1}{1}$;
    \item \emph{half-words}: for $k=3,\ldots,\buhalflen$, in this order: for every string $\subword$ of length $k-1$ and character $\character\in\{\symbolzero,\symbolone\}$ such that $\subword\circ\character$ is a prefix of some half-word, merge $\mergestring{\subword}{\character}$;
    \item \emph{codewords}: for every $a\in\srcalphabet$, merge $\mergestring{\buhalf{a}}{\buhalf{a}}$, and then $\mergestring{\buhalf{a}\circ\buhalf{a}}{0}$.
\end{enumerate}

\textit{Correctness.}
We follow an encoded string through the three stages.

\textit{Stage (i): markers.}
In an encoded string, every pair of adjacent $\symbolone$s starts a half-word, and is preceded by a $\symbolzero$ (or the string's start) and followed by a $\symbolzero$.
Stage (i) thus merges exactly the first two characters of every half-word.

\textit{Stage (ii): half-words.}
We show, by induction on $k$, that after the iteration for $k$, the first $k$ characters of every half-word form a single token, and all other characters are single-character tokens.
For $k=2$, this is the state after stage (i).
Now let $3\leq k\leq\buhalflen$, and assume the claim for $k-1$.
Then every token $\subword$ with at least two characters is the prefix of length $k-1$ of a half-word, at its start, and is followed by the $k$-th character $\character$ of that half-word as a single-character token.
The merge $\mergestring{\subword}{\character}$ thus applies exactly at the half-words starting with $\subword\circ\character$.
Each half-word is therefore extended by exactly one merge of this iteration; the resulting token has length $k$, so no later merge of the iteration, whose left token has length $k-1$, matches it.
After iteration $\buhalflen$, the encoded string is thus tokenised as $\langle\buhalf{a_1},\buhalf{a_1},\symbolzero,\buhalf{a_2},\buhalf{a_2},\symbolzero,\ldots\rangle$.

\textit{Stage (iii): codewords.}
As distinct symbols have distinct half-words, each token $\buhalf{a}$ occurs in runs of exactly two.
The merge $\mergestring{\buhalf{a}}{\buhalf{a}}$ thus joins the two halves of every codeword of $a$, and the next merge appends the trailing $\symbolzero$.

\textit{Size.}
Stages (i) and (ii) use one merge per distinct half-word prefix of length between $2$ and $\buhalflen$.
For each $j\in\{1,\ldots,\bucodelength\}$, the prefixes ending with $\symbolzero$ just before bit $\fwrbit_j$ number at most $2^{j-1}$, and those ending with $\fwrbit_j$ at most $2^{j}$; there is one prefix of length $2$, and at most $|\srcalphabet|$ of length $\buhalflen$.
Stage (iii) uses $2|\srcalphabet|$ merges.
As $2^{\bucodelength}<2|\srcalphabet|$, we get
\begin{align}
    \nprep
    \leq
    1+\sum_{j=1}^{\bucodelength}\big(2^{j-1}+2^{j}\big)+3|\srcalphabet|
    =
    3\cdot2^{\bucodelength}-2+3|\srcalphabet|
    <
    10|\srcalphabet|.
\end{align}
\end{proof}

Once every codeword is a token, the binary string behaves like the source string.
For a merge sequence $\merges$ over $\srcalphabet$ with merges $\merge_k=\mergestring{\subword_k}{\subword_k'}$, let
$\buenc{\merges}\defeq\langle\mergestring{\buenc{\subword_1}}{\buenc{\subword_1'}},\ldots,\mergestring{\buenc{\subword_{|\merges|}}}{\buenc{\subword_{|\merges|}'}}\rangle$.

\begin{lemma}[Simulating merges]
\label{lemma:bu_simulate}
Let $\merges$ be a merge sequence over $\srcalphabet$, and let $\characters\in\srcalphabet^+$.
If $\bottomuptoken[\merges](\characters)=\langle\subword_1,\ldots,\subword_n\rangle$, then applying $\mergesprep$ and then $\buenc{\merges}$ to $\buenc{\characters}$ gives $\langle\buenc{\subword_1},\ldots,\buenc{\subword_n}\rangle$.
\end{lemma}

\begin{proof}
Codewords have equal lengths, and distinct symbols have distinct codewords; so for strings $\subword,\subword'\in\srcalphabet^+$, we have $\buenc{\subword}=\buenc{\subword'}$ if and only if $\subword=\subword'$.
By \cref{lemma:bu_prep}, after $\mergesprep$ the binary tokenisation is the encoding, token by token, of the source string's characters.
Suppose that, before some merge $\mergestring{\subword}{\subword'}$ of $\merges$, the binary tokenisation $\langle\buenc{\subword_1},\ldots,\buenc{\subword_n}\rangle$ encodes the source tokenisation $\langle\subword_1,\ldots,\subword_n\rangle$ token by token.
Then the pair $\langle\buenc{\subword},\buenc{\subword'}\rangle$ occurs at index $j$ of the binary tokenisation if and only if $\langle\subword,\subword'\rangle$ occurs at index $j$ of the source tokenisation.
Both left-to-right scans thus replace pairs at the same indices, and as $\buenc{\subword_j}\circ\buenc{\subword_{j+1}}=\buenc{\subword_j\circ\subword_{j+1}}$, the two tokenisations still correspond afterwards.
Induction over the merges of $\merges$ proves the lemma.
\end{proof}

\begin{proof}[Proof of \cref{prop:bu_gap}\ref{item:bu_forward}]
Suppose $\maxcutof{\srcgraph}\geq\cuttarget$.
Then, by \cref{eq:bu_collapse_cost,eq:bu_budget},
\begin{align}
    \collapsecost
    =
    3\nwalks+|\arcs|-\nclones^2\maxcutof{\srcgraph}
    \leq
    3\nwalks+|\arcs|-\nclones^2\cuttarget
    =
    \vocabsize-\nprep.
\end{align}
By \cref{lemma:bu_collapse}, the merge sequence $\mergessrc$ tokenises $\srcword$ as $\langle\srcword\rangle$.
By \cref{lemma:bu_simulate}, applying $\mergesprep$ and then $\buenc{\mergessrc}$ tokenises $\buenc{\srcword}$ as $\langle\buenc{\srcword}\rangle$, a single token, using $\nprep+\collapsecost\leq\vocabsize$ merges.
\end{proof}

\subsection{Proof of the Backward Implication}
\label{appendix:bu_backward}

The forward implication used a merge sequence which respects codewords.
The backward implication must handle \emph{any} merge sequence over $\{\symbolzero,\symbolone\}$, which may split codewords, combine parts of different codewords, or never build a codeword at all.
We show that any merge sequence $\merges$ leaves at least $\collapsecost-|\merges|+1$ tokens on $\buenc{\srcword}$ (\cref{lemma:bu_centre_bound}).
To this end, we track the centres.
Every codeword contains exactly one centre, so initially $4\nwalks+1$ tokens contain a centre, and reducing this number requires replacements which combine two tokens that \emph{both} contain a centre.
Such tokens are long enough to reveal which source symbols they cover (\cref{lemma:bu_recognition}), so the merges creating them behave like merges on the source string.

\begin{lemma}[Recognition]
\label{lemma:bu_recognition}
Let $\subword\in\{\symbolzero,\symbolone\}^+$ be a string which, at some occurrence in $\buenc{\srcword}$, contains at least two centres.
Then:
\begin{enumerate}[label=(\alph*)]
    \item \label{item:bu_recog_phase}
    at every occurrence of $\subword$ in $\buenc{\srcword}$, its centres lie at the same positions within $\subword$;
    \item \label{item:bu_recog_symbols}
    at every occurrence, these centres belong to the codewords of the same substring of $\srcword$, which we denote by $\srcdecode(\subword)$.
\end{enumerate}
\end{lemma}

\begin{figure}[t]
\centering
\begin{tikzpicture}[
    recogcell/.style={draw,line width=0.4pt,minimum width=3.2mm,minimum height=4.4mm,inner sep=0pt,anchor=south west,font=\footnotesize},
    recogbrace/.style={decorate,decoration={brace,amplitude=3pt,raise=1pt},line width=0.5pt},
    recogmbrace/.style={decorate,decoration={brace,amplitude=3pt,mirror,raise=1pt},line width=0.5pt},
  ]
  \colorlet{recogmarker}{blind_magenta!12}
\node[anchor=east,font=\small] at (-0.1,0.21) {$\buenc{\srcword}$:};
  \node[anchor=west,font=\small] at (12.48,0.21) {$\cdots$};
  \node[recogcell,fill=recogmarker] at (0.00,0) {$\symbolone$};
  \node[recogcell,fill=recogmarker] at (0.32,0) {$\symbolone$};
  \node[recogcell,fill=white] at (0.64,0) {$\symbolzero$};
  \node[recogcell,fill=white] at (0.96,0) {$\symbolzero$};
  \node[recogcell,fill=white] at (1.28,0) {$\symbolzero$};
  \node[recogcell,fill=white] at (1.60,0) {$\symbolone$};
  \node[recogcell,fill=white] at (1.92,0) {$\symbolzero$};
  \node[recogcell,fill=black!15] at (2.24,0) {$\symbolone$};
  \node[recogcell,fill=recogmarker] at (2.56,0) {$\symbolone$};
  \node[recogcell,fill=white] at (2.88,0) {$\symbolzero$};
  \node[recogcell,fill=white] at (3.20,0) {$\symbolzero$};
  \node[recogcell,fill=white] at (3.52,0) {$\symbolzero$};
  \node[recogcell,fill=white] at (3.84,0) {$\symbolone$};
  \node[recogcell,fill=white] at (4.16,0) {$\symbolzero$};
  \node[recogcell,fill=white] at (4.48,0) {$\symbolzero$};
  \node[recogcell,fill=recogmarker] at (4.80,0) {$\symbolone$};
  \node[recogcell,fill=recogmarker] at (5.12,0) {$\symbolone$};
  \node[recogcell,fill=white] at (5.44,0) {$\symbolzero$};
  \node[recogcell,fill=white] at (5.76,0) {$\symbolone$};
  \node[recogcell,fill=white] at (6.08,0) {$\symbolzero$};
  \node[recogcell,fill=white] at (6.40,0) {$\symbolzero$};
  \node[recogcell,fill=white] at (6.72,0) {$\symbolzero$};
  \node[recogcell,fill=black!15] at (7.04,0) {$\symbolone$};
  \node[recogcell,fill=recogmarker] at (7.36,0) {$\symbolone$};
  \node[recogcell,fill=white] at (7.68,0) {$\symbolzero$};
  \node[recogcell,fill=white] at (8.00,0) {$\symbolone$};
  \node[recogcell,fill=white] at (8.32,0) {$\symbolzero$};
  \node[recogcell,fill=white] at (8.64,0) {$\symbolzero$};
  \node[recogcell,fill=white] at (8.96,0) {$\symbolzero$};
  \node[recogcell,fill=white] at (9.28,0) {$\symbolzero$};
  \node[recogcell,fill=recogmarker] at (9.60,0) {$\symbolone$};
  \node[recogcell,fill=recogmarker] at (9.92,0) {$\symbolone$};
  \node[recogcell,fill=white] at (10.24,0) {$\symbolzero$};
  \node[recogcell,fill=white] at (10.56,0) {$\symbolone$};
  \node[recogcell,fill=white] at (10.88,0) {$\symbolzero$};
  \node[recogcell,fill=white] at (11.20,0) {$\symbolone$};
  \node[recogcell,fill=white] at (11.52,0) {$\symbolzero$};
  \node[recogcell,fill=black!15] at (11.84,0) {$\symbolone$};
  \node[recogcell,fill=recogmarker] at (12.16,0) {$\symbolone$};
\fill (2.40,0.52) circle (1.2pt);
  \fill (7.20,0.52) circle (1.2pt);
  \fill (12.00,0.52) circle (1.2pt);
\draw[recogbrace] (0.00,0.62) -- node[above=3pt,font=\small] {$\buenc{a}$} (4.80,0.62);
  \draw[recogbrace] (4.80,0.62) -- node[above=3pt,font=\small] {$\buenc{b}$} (9.60,0.62);
  \draw[recogbrace] (9.60,0.62) -- node[above=3pt,font=\small] {$\buenc{c}$\,$\cdots$} (12.48,0.62);
\draw[blind_magenta,line width=1.6pt] (0.96,1.47) -- (11.20,1.47);
  \draw[blind_magenta,line width=0.6pt] (0.96,1.37) -- ++(0,0.2) (11.20,1.37) -- ++(0,0.2);
  \node[anchor=south,font=\small] at (6.08,1.51) {an occurrence of $\subword$};
  \draw[blind_magenta!45,densely dotted,line width=0.5pt] (0.96,1.47) -- (0.96,0) (11.20,1.47) -- (11.20,0);
\draw[recogmbrace] (0.00,-0.06) -- node[below=3pt,font=\small] {$\buhalflen$} (2.24,-0.06);
  \draw[recogmbrace,blind_magenta,line width=0.8pt] (2.24,-0.06) -- node[below=3pt,font=\small,text=blind_magenta] {$\buhalflen{+}1$} (4.80,-0.06);
  \draw[recogmbrace] (4.80,-0.06) -- node[below=3pt,font=\small] {$\buhalflen$} (7.04,-0.06);
  \draw[recogmbrace] (7.04,-0.06) -- node[below=3pt,font=\small] {$\buhalflen{+}1$} (9.60,-0.06);
  \draw[recogmbrace] (9.60,-0.06) -- node[below=3pt,font=\small] {$\buhalflen$} (11.84,-0.06);
\node[anchor=east,font=\small,black!60] at (-0.1,-0.95) {phase:};
  \node[font=\small,black!60] at (0.16,-0.95) {$0$};
  \node[font=\small,black!60] at (2.40,-0.95) {$\buhalflen$};
  \node[font=\small,black!60] at (4.96,-0.95) {$0$};
  \node[font=\small,black!60] at (7.20,-0.95) {$\buhalflen$};
  \node[font=\small,black!60] at (9.76,-0.95) {$0$};
  \node[font=\small,black!60] at (12.00,-0.95) {$\buhalflen$};
\end{tikzpicture}
\caption{
Markers $\symbolone\symbolone$ (tinted) start every half-word, at phases $0$ and $\buhalflen$, and centres (shaded, with dots) lie at phase $\buhalflen$; here $\buhalflen=7$ and $\buenclen=15$.
The gap to the next marker is $\buhalflen$ after a phase-$0$ marker and $\buhalflen+1$ after a phase-$\buhalflen$ one.
The first two markers inside $\subword$ are $\buhalflen+1$ apart, so the first of them is a centre (\cref{lemma:bu_recognition}).
}
\label{fig:bu_recognition}
\end{figure}
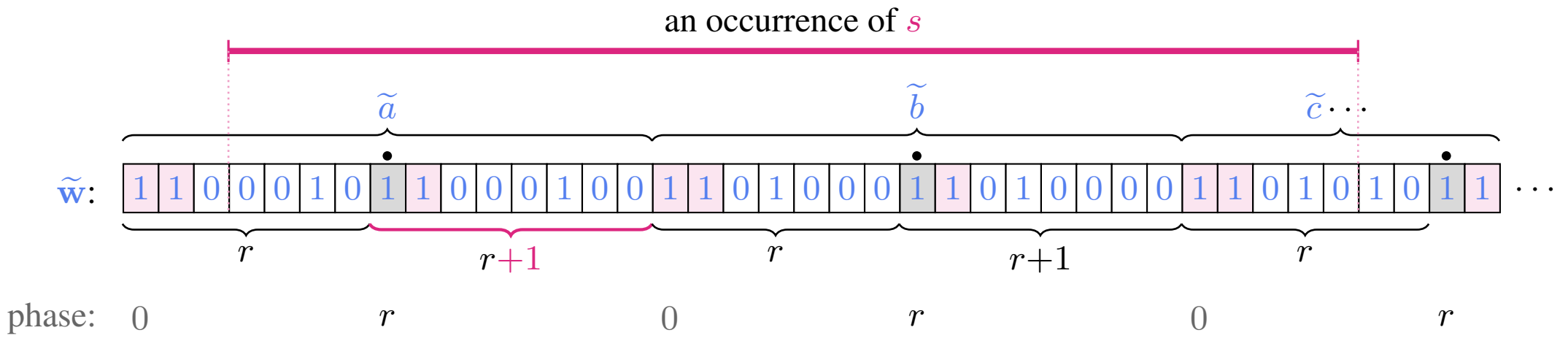

\begin{proof}
A short binary string, such as $\symbolzero\symbolone\symbolone\symbolzero$, occurs all over $\buenc{\srcword}$ and carries no information about where it sits.
The lemma says that this changes once a string contains two centres: its content alone then tells us where it sits relative to the codewords, and which source symbols it covers.

To make ``where it sits'' precise, number the positions of $\buenc{\srcword}$ from $0$, and call $p\bmod\buenclen$ the \defn{phase} of position $p$.
Every codeword starts at phase $0$, and its centre lies at phase $\buhalflen$.
We call each occurrence of $\symbolone\symbolone$ a \defn{marker}; as noted in Stage~3, markers are exactly the starts of half-words, so they start at phases $0$ and $\buhalflen$.

The difficulty is that these two kinds of markers look alike: a codeword and its second half both begin with the same half-word.
What tells them apart is the distance to the next marker.
After a marker of phase $0$, the next one starts a half-word later, that is, $\buhalflen$ positions later.
After a marker of phase $\buhalflen$, the next one starts only after the half-word \emph{and} the trailing $\symbolzero$, that is, $\buhalflen+1$ positions later ({\cref{fig:bu_recognition}}).
Two consecutive markers thus reveal the phase of the first: it is $0$ if they are $\buhalflen$ apart, and $\buhalflen$ if they are $\buhalflen+1$ apart.

\textit{Part \ref{item:bu_recog_phase}.}
It therefore suffices to show that every occurrence of $\subword$ contains two complete consecutive markers.
As $\subword$ contains two centres, which are $\buenclen$ positions apart, we have $|\subword|\geq\buenclen+1$.
Let an occurrence of $\subword$ start at position $p$.
If a marker starts at $p$, the next one starts at most $\buhalflen+1$ positions later, and thus ends by position $p+\buenclen$.
Otherwise, let $m<p$ be the start of the last marker before $p$; the two markers following it both start after $p$, and since consecutive gaps sum to $\buenclen$, the second of them starts at $m+\buenclen$ and ends by position $p+\buenclen$.
Either way, the occurrence contains two complete consecutive markers.
Since every $\symbolone\symbolone$ in $\buenc{\srcword}$ is a marker, these are simply the first two occurrences of $\symbolone\symbolone$ in $\subword$; their distance depends on $\subword$ alone, and fixes the phase of the first of them, and hence of every position in $\subword$.
All occurrences of $\subword$ thus start at the same phase, and their centres, the positions of phase $\buhalflen$, lie at the same positions within $\subword$.

\textit{Part \ref{item:bu_recog_symbols}.}
Knowing where the centres lie, we can now read off the symbols.
Take two consecutive centres in an occurrence of $\subword$, belonging to the codewords $\buenc{a}$ and $\buenc{b}$ of two adjacent symbols $ab$ of $\srcword$.
Between them, $\subword$ contains $\buhalf{a}\circ\symbolzero\circ\buhalf{b}$, the second half-word of $\buenc{a}$, its trailing $\symbolzero$, and the first half-word of $\buenc{b}$ (\cref{fig:bu_codeword_pair}), at positions fixed by part \ref{item:bu_recog_phase}.
As identifiers are distinct, these half-words determine $a$ and $b$.
This is why codewords repeat their half-word: with a single half-word, the string between two centres would reveal only the first of the two symbols.
Applying this to every pair of consecutive centres, the content of $\subword$ determines every symbol whose centre it contains.
As consecutive centres belong to consecutive codewords, these symbols form a substring of $\srcword$, the same at every occurrence.
\end{proof}

To count how often tokens containing centres are combined, we record the entire run of a merge sequence as a forest.

\begin{defin}[Merge forest]
\label{def:merge_forest}
Let $\merges$ be a merge sequence, whose merges may be arbitrary pairs of strings, and apply it to $\buenc{\srcword}$.
Its \defn{merge forest} records every replacement as a node whose two children are the nodes (or characters) of the two tokens it combines.
A node is \defn{created by step $k$} if the replacement is made by the merge $\merge_k$, and it is a \defn{join} if \emph{both} of its children contain a centre.
\end{defin}

\begin{figure}[t]
\centering
\begin{tikzpicture}[x=5.5mm,y=9mm]
\foreach \i in {1,...,20} {
        \pgfmathtruncatemacro{\r}{mod(\i,5)}
        \ifnum\r=3
            \node[bucentre] (l\i) at (\i,0) {};
        \else
            \node[buleaf] (l\i) at (\i,0) {};
        \fi
    }
\foreach \k/\a/\b in {1/1/5,2/6/10,3/11/15,4/16/20} {
        \draw[decorate,decoration={brace,amplitude=4pt,mirror}]
            ([yshift=-6pt]l\a.south west) -- ([yshift=-6pt]l\b.south east)
            node[midway,below=5pt,font=\small] {$\buenc{a_{\k}}$};
    }
\foreach \k/\a/\b/\m in {1/1/5/3,2/6/10/8,3/11/15/13} {
        \draw[black!50,fill=black!6] ([yshift=2pt]l\a.north west) -- ([yshift=2pt]l\b.north east) -- (\m,1.25) -- cycle;
        \node[bunode] (x\k) at (\m,1.25) {};
    }
\node[bujoin] (j) at (5.5,2.3) {};
    \draw[very thick,blind_magenta] (j) -- (x1);
    \draw[very thick,blind_magenta] (j) -- (x2);
\node[bunode] (t) at (16.5,0.85) {};
    \draw (t) -- (l16);
    \draw (t) -- (l17);
    \node[bunode] (r2) at (14.5,2.3) {};
    \draw (r2) -- (x3);
    \draw (r2) -- (t);
\node[bunode] (u) at (18.5,0.85) {};
    \draw (u) -- (l18);
    \draw (u) -- (l19);
    \node[bunode] (r3) at (19.25,1.6) {};
    \draw (r3) -- (u);
    \draw (r3) -- (l20);
\node[font=\small,above=2pt] at (j.north) {join};
    \node[font=\small,above=2pt] at (r2.north) {not a join};
\end{tikzpicture}
\caption{
A merge forest on the encoding of a source string $a_1a_2a_3a_4$, with codewords drawn schematically as five characters each and centres in black.
Shaded triangles abbreviate subtrees which build a whole codeword; they contain no join, as every codeword contains a single centre.
The forest has one join (highlighted), which combines two tokens that each contain a centre.
The other two roots have only one child containing a centre each; in particular, the middle root spans codeword $\buenc{a_3}$ together with the first two characters of $\buenc{a_4}$, ignoring codeword boundaries.
The forest has $\ncentres=3$ roots containing a centre and $4-3=1$ join, as in the first step of the proof of \cref{lemma:bu_centre_bound}.
}
\label{fig:bu_forest}
\end{figure}
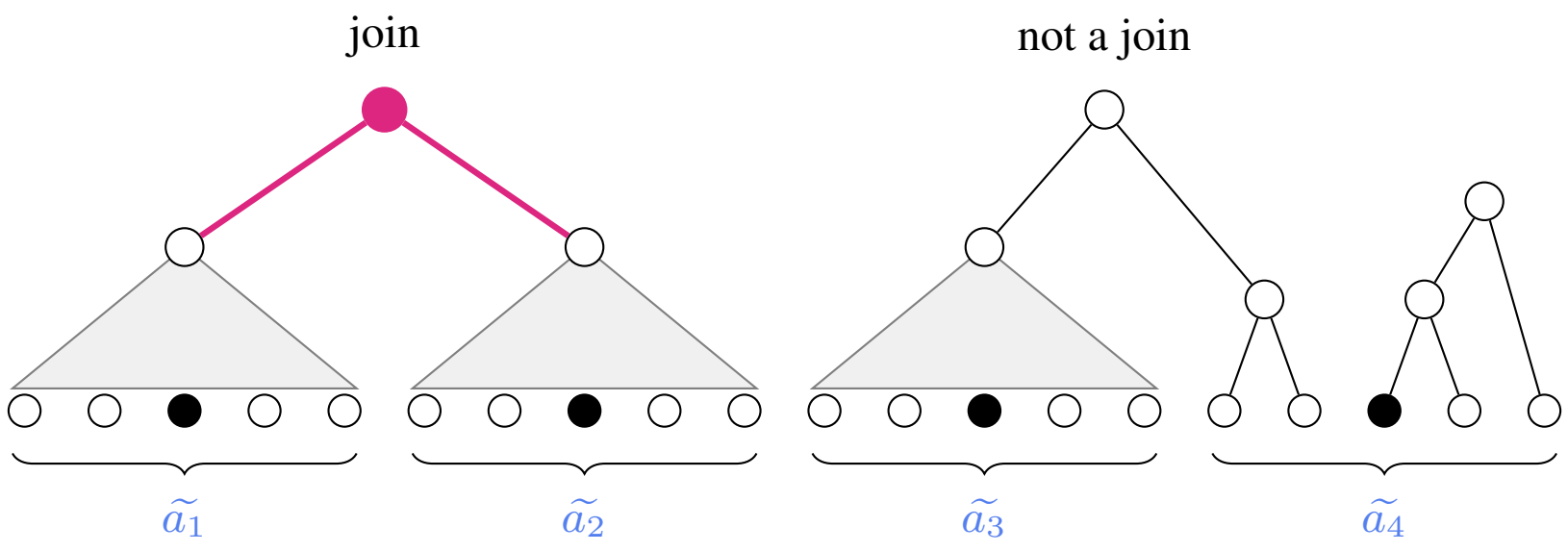

The leaves of the merge forest are the characters of $\buenc{\srcword}$, and its roots are the tokens of $\bottomuptoken[\merges](\buenc{\srcword})$ (\cref{fig:bu_forest}).
Every node spans an interval of positions, which spells out its token; the intervals of two nodes are either nested or disjoint, and distinct nodes span distinct intervals.

\begin{lemma}[Centre bound]
\label{lemma:bu_centre_bound}
Let $\merges$ be any merge sequence, and let $\ncentres$ be the number of tokens of $\bottomuptoken[\merges](\buenc{\srcword})$ which contain a centre.
Then
\begin{align}
    \ncentres\geq\collapsecost-|\merges|+1.
\end{align}
\end{lemma}

\begin{proof}
We proceed in five steps:
\circled{1} we count the joins in the merge forest;
\circled{2} we show that each merge step creates either only joins or none;
\circled{3} we show that most such joining steps create a single node;
\circled{4} we bound the joins created by the remaining steps, which merge walk bigrams;
\circled{5} we combine these bounds.

\begin{mylemmastep}
The merge forest contains exactly $4\nwalks+1-\ncentres$ joins.
\end{mylemmastep}

\begin{proof}
Track the number of current tokens which contain a centre.
Initially, it is $4\nwalks+1$, as each centre is a single character.
A join decreases this number by one, and any other replacement leaves it unchanged.
At the end, it equals $\ncentres$.
\end{proof}

\begin{mylemmastep}
If step $k$ creates a join, then every node created by step $k$ is a join.
\end{mylemmastep}

\begin{proof}
Let $\merge_k=\mergestring{\subword}{\subword'}$.
Every node created by step $k$ spans an occurrence of $\subword\circ\subword'$, with its left child spanning the first $|\subword|$ characters.
If one of these nodes is a join, then $\subword\circ\subword'$ contains two centres at that occurrence; by \cref{lemma:bu_recognition}\ref{item:bu_recog_phase}, it then contains its centres at the same positions at every occurrence.
So, at every node created by step $k$, both children contain a centre.
\end{proof}

We call a step \defn{joining} if it creates a join.
For a joining step with $\merge_k=\mergestring{\subword}{\subword'}$, we write $\srcdecode(k)\defeq\srcdecode(\subword\circ\subword')$ for the substring of $\srcword$ it covers (\cref{lemma:bu_recognition}\ref{item:bu_recog_symbols}).
We call it a \defn{bigram step} if $\srcdecode(k)$ consists of exactly two vertices; by \cref{lemma:bu_unique}\ref{item:bu_unique_adj}, $\srcdecode(k)$ is then an arc.

\begin{mylemmastep}
Every joining step which is not a bigram step creates exactly one node.
\end{mylemmastep}

\begin{proof}
For such a step, $\srcdecode(k)$ either has length at least $3$, or is a bigram containing a separator or $\buendsym$.
By \cref{lemma:bu_unique}\ref{item:bu_unique_sep}--\ref{item:bu_unique_long}, it thus occurs at a single position of $\srcword$.
By \cref{lemma:bu_recognition}, every occurrence of $\subword\circ\subword'$ covers the centres of exactly these symbols, at fixed positions within $\subword\circ\subword'$; all its occurrences therefore start at the same position.
As distinct nodes span distinct intervals, step $k$ creates only one node.
\end{proof}

\begin{mylemmastep}
Let $\bigramset\subseteq\arcs$ be the set of arcs $\srcdecode(k)$ of all bigram steps $k$.
Then bigram steps create at most $\covfunc(\bigramset)$ nodes in total.
\end{mylemmastep}

\begin{proof}
Identify every centre with the position of its symbol in $\srcword$.
A node created by a bigram step contains exactly two centres, at adjacent positions $i$ and $i+1$, whose symbols form an arc in $\bigramset$.
By \cref{lemma:bu_unique}\ref{item:bu_unique_adj}, both positions lie in the same walk, which $\bigramset$ thus covers; we assign the node to this walk.
It suffices to show that every walk is assigned at most one node.
If a walk occupies positions $j$, $j+1$ and $j+2$, the nodes assigned to it have centres $\{j,j+1\}$ or $\{j+1,j+2\}$.
Two joins with the same two centres coincide, as each separates these centres between its two children, and is thus their lowest common ancestor.
And a node with centres $\{j,j+1\}$ and one with centres $\{j+1,j+2\}$ cannot both exist: both contain centre $j+1$, so their intervals would be nested, and so would their sets of centres.
\end{proof}

\begin{mylemmastep}
$\ncentres\geq\collapsecost-|\merges|+1$.
\end{mylemmastep}

\begin{proof}
There are at least $|\bigramset|$ bigram steps, so at most $|\merges|-|\bigramset|$ other joining steps.
By Steps \circled{2} to \circled{4}, the merge forest thus contains at most $\covfunc(\bigramset)+|\merges|-|\bigramset|$ joins.
With Step \circled{1} and \cref{lemma:bu_cover}\ref{item:bu_cover_some}, we get
\begin{align}
    4\nwalks+1-\ncentres
    \leq
    \covfunc(\bigramset)-|\bigramset|+|\merges|
    \leq
    \nwalks-|\arcs|+\nclones^2\maxcutof{\srcgraph}+|\merges|.
\end{align}
Rearranging gives $\ncentres\geq3\nwalks+|\arcs|-\nclones^2\maxcutof{\srcgraph}-|\merges|+1=\collapsecost-|\merges|+1$.
\end{proof}

Step \circled{5} concludes the proof.
\end{proof}

\begin{proof}[Proof of \cref{prop:bu_gap}\ref{item:bu_backward}]
We first bound the cost of building the codewords.
The alphabet $\srcalphabet$ has at most $\nclones\vcnvertices+9\nclones^2+1\leq2\nclones\vcnvertices$ symbols, as $\vcnvertices\geq9\nclones+1$.
By \cref{lemma:bu_prep} and $\nclones\geq40/\gapconst$, we thus have
\begin{align}
    \nprep
    <
    10|\srcalphabet|
    \leq
    20\nclones\vcnvertices
    \leq
    \frac{\gapconst}{2}\nclones^2\vcnvertices.
    \label{eq:bu_prep_bound}
\end{align}
Now let $\merges$ be a merge sequence with $|\merges|\leq\vocabsize$ which leaves at most $\frac{\gapconst}{2}\nclones^2\vcnvertices$ tokens.
As $\ncentres$ counts only some of the remaining tokens, \cref{lemma:bu_centre_bound} gives
\begin{align}
    \collapsecost-\vocabsize+1
    \leq
    \ncentres
    \leq
    \frac{\gapconst}{2}\nclones^2\vcnvertices.
\end{align}
Substituting $\collapsecost$ and $\vocabsize$ from \cref{eq:bu_collapse_cost,eq:bu_budget}, the terms $3\nwalks+|\arcs|$ cancel, and with \cref{eq:bu_prep_bound}:
\begin{align}
    \nclones^2\big(\cuttarget-\maxcutof{\srcgraph}\big)
    \leq
    \frac{\gapconst}{2}\nclones^2\vcnvertices+\nprep-1
    <
    \gapconst\nclones^2\vcnvertices.
\end{align}
Dividing by $\nclones^2$ gives $\maxcutof{\srcgraph}>\cuttarget-\gapconst\vcnvertices$.
\end{proof}

\section{Unary Direct Tokenisation is  \boldmath\apx\unboldmath-hard}
\label{sec:unary_apx}
 
\citet{kastreva2026bounded} showed that unary direct tokenisation is strongly \np-complete; this appendix shows that it is also \apx-hard (\cref{thm:unary_apx}), by an AP-reduction from \cubicmaxcut. 
As in their work, we identify a unary string $\charstring{a}^{\ell}$ with its length $\ell\in\N$, a dataset with a multiset of lengths, and a vocabulary with the set of its token lengths; the alphabet symbol is then the length-$1$ token, which every vocabulary contains. We write \dutok for unary direct tokenisation under the compression objective $\objectivefunclength$, viewed as an optimisation problem: given a dataset and a budget $\vocabsize$, find a vocabulary that minimises the compressed length.
\Cref{appendix:ap_reductions} recalls AP-reductions, and \cref{sec:unary_apx_reduction} constructs the reduction.
\Cref{sec:unary_apx_theorem} states its two key properties: every cut-compliant vocabulary achieves compressed length $\basecost$ minus the cut it encodes (\cref{lemma:unary_apx_forward}), and every vocabulary decodes into a cut that does no worse (\cref{lemma:unary_apx_backward}).
Taking these as given, \cref{appendix:unary_apx_ap} proves \cref{thm:unary_apx}.
The remaining subsections prove the two lemmas, using the label arithmetic of \cref{appendix:unary_apx_arithmetic} and the rounding argument of \cref{appendix:unary_apx_rounding}.
 
\subsection{AP-Reductions}
\label{appendix:ap_reductions}
 
For an instance $\instance$ of an optimisation problem and a feasible solution $\solutionlred$ with objective value $\objectivefunc(\solutionlred)$, let the \defn{performance ratio} be
\begin{align}
    \perfratio(\instance,\solutionlred)
    \defeq
    \max\left(
        \frac{\objectivefunc(\solutionlred)}{\OPTop(\instance)},
        \frac{\OPTop(\instance)}{\objectivefunc(\solutionlred)}
    \right)
    \geq 1,
\end{align}
which covers both minimisation and maximisation problems.
 
\begin{defin}
\label{def:ap_reduction}
Let $\mathsf{A}$ and $\mathsf{B}$ be optimisation problems.
We say that $\mathsf{A}$ AP-reduces to $\mathsf{B}$ (denoted $\mathsf{A}\apred\mathsf{B}$) if there exist functions $\ffunctionlreddef$ and $\gfunctionlreddef$ and a constant $\apconst\geq1$ such that, for every instance $\instance_{\mathsf{A}}$ of $\mathsf{A}$ and every rational $\targetratio>1$:
\begin{enumerate}[label=(\roman*)]
    \item 
$\instance_{\mathsf{B}}=\ffunctionlreddef(\instance_{\mathsf{A}},\targetratio)$ is an instance of $\mathsf{B}$, computable in time polynomial in $|\instance_{\mathsf{A}}|$ for every fixed $\targetratio$;
    \item for every feasible solution $\solutionlred_{\mathsf{B}}$ of $\instance_{\mathsf{B}}$, $\gfunctionlreddef(\instance_{\mathsf{A}},\solutionlred_{\mathsf{B}},\targetratio)$ is a feasible solution of $\instance_{\mathsf{A}}$, computable in time polynomial in $|\instance_{\mathsf{A}}|$ and $|\solutionlred_{\mathsf{B}}|$ for every fixed $\targetratio$;
    \item the approximation quality is preserved up to $\apconst$:
    \begin{align}
        \perfratio_{\mathsf{B}}(\instance_{\mathsf{B}},\solutionlred_{\mathsf{B}})\leq\targetratio
        \implies
        \perfratio_{\mathsf{A}}\big(\instance_{\mathsf{A}},\gfunctionlreddef(\instance_{\mathsf{A}},\solutionlred_{\mathsf{B}},\targetratio)\big)\leq1+\apconst(\targetratio-1).
    \end{align}
\end{enumerate}
\end{defin}
 
AP-reductions compose, map polynomial-time approximation schemes for $\mathsf{B}$ to ones for $\mathsf{A}$, and thus preserve \apx-hardness: if $\mathsf{A}$ is \apx-hard and $\mathsf{A}\apred\mathsf{B}$, then $\mathsf{B}$ is \apx-hard \citep{crescenzi1997short,ausiello2012complexity}.
The reduction we give below does not depend on $\targetratio$.

\subsection{The Reduction}
\label{sec:unary_apx_reduction}
 
Our reduction first blows up $\srcgraph$ with $\vcnvertices$ clones per vertex: we write $\blowgraph=(\blowvertices,\blowedges)$ for its $\vcnvertices$-blow-up (\cref{def:blowup}), so each vertex becomes a cluster of $\vcnvertices$ clones and each edge a complete bipartite graph between two clusters.
If $\srcgraph$ is cubic, $\blowgraph$ is simple, has $\vcnvertices^2$ vertices and $\vcmedges\vcnvertices^2$ edges, each of its vertices has degree $3\vcnvertices$, and colourings induced from $\srcgraph$ cut $\vcnvertices^2$ times as many edges (\cref{lemma:blowup}\ref{item:blowup_induced}).
 
\begin{reduction}
\label{red:unary_apx}
Consider an instance $\srcgraph=(\vertices,\edges)$ of the \cubicmaxcut optimisation problem.
Index the vertices of $\blowgraph$ as  $\blowvertices=\{1,\dots,\vcnvertices^2\}$.
Now, define the label base, the two markers, and a label for each vertex:
\begin{align}
    \labelbase \defeq 8\vcnvertices^8+1,
    \qquad
    \markB \defeq \labelbase^{10},
    \qquad
    \markW \defeq \labelbase^{15},
    \qquad
    \enc{i} \defeq \sum_{k=1}^{4} i^k\labelbase^k.
\end{align}
From these, we define the string-lengths, for $i\in\blowvertices$, $\theta\in\{\markB,\markW\}$, and $\{i,j\}\in\blowedges$:
\begin{align}
    \vtxlen{i} &\defeq \enc{i},
    &
    \collen{i}{\theta} &\defeq \enc{i}+\theta,
    \\
    \bridgelen{i} &\defeq \enc{i}+\markB+\markW,
    &
    \edgelenij{i}{j} &\defeq \enc{i}+\enc{j}+\markB+\markW.
\end{align}
We build five subdatasets from these values:
\begin{subequations}
\label{eq:unary_apx_datasets}
\begin{align}
    \dataset_1 &= \{\markB, \markW\}\times\nrepeatvar\vcnvertices^2
        && \mathcomment{marker-strings}\\
    \dataset_2 &= \{\vtxlen{i}\mid i\in\blowvertices\}\times\nrepeatvar'
        && \mathcomment{vertex-strings}\\
    \dataset_3 &= \{\collen{i}{\markB}, \collen{i}{\markW}\mid i\in\blowvertices\}\times\nrepeatvar''
        && \mathcomment{colour-strings}\\
    \dataset_4 &= \{\bridgelen{i}\mid i\in\blowvertices\}\times1
        && \mathcomment{bridge-strings}\\
    \dataset_5 &= \{\edgelenij{i}{j}\mid\{i,j\}\in\blowedges\}\times1
        && \mathcomment{edge-strings}
\end{align}
\end{subequations}
with multiplicities $\nrepeatvar\defeq20\vcnvertices$, $\nrepeatvar'\defeq17\vcnvertices$, and $\nrepeatvar''\defeq15\vcnvertices$.
A full dataset is then formed by joining these subdatasets: $\dataset=\dataset_1\cup\dots\cup\dataset_5$.
Finally, we set the number of allowed tokens to $\vocabsize=2\vcnvertices^2+2$, and define the base cost
\begin{align}
    \basecost
    \defeq
    (2\nrepeatvar+\nrepeatvar'+3\nrepeatvar''+2)\vcnvertices^2+3\vcmedges\vcnvertices^2
    =
    \frac{213}{2}\vcnvertices^3+2\vcnvertices^2.
    \label{eq:unary_apx_basecost}
\end{align}
\end{reduction}
 
We write $\reductionfuncfivefull$ for the \dutok instance $(\dataset,\vocabsize)$ output by this reduction.
It contains $\frac{177}{2}\vcnvertices^3+\vcnvertices^2$ strings, each shorter than $2\markW=O(\vcnvertices^{120})$, so it can be written down explicitly in unary in polynomial time.
Following \citet{kastreva2026bounded}, we split the intended tokens into structural tokens $\vocabboundary$, which every good vocabulary should contain, and choice tokens $\vocabchoice$, from which it should pick one per vertex:
\begin{align}
    \vocabboundary
    \defeq
    \{\markB,\markW\}
    \cup
    \{\vtxlen{i}\mid i\in\blowvertices\},
    \qquad
    \vocabchoice
    \defeq
    \{\collen{i}{\markB}, \collen{i}{\markW}\mid i\in\blowvertices\}.
    \label{eq:unary_apx_compliant}
\end{align}
We call the members of $\vocabcompl\defeq\alphabet\cup\vocabboundary\cup\vocabchoice$ \defn{compliant} tokens, and all other tokens \defn{non-compliant}.
We further call a vocabulary \defn{cut-compliant} if it contains $\alphabet\cup\vocabboundary$ and exactly one of $\collen{i}{\markB}$ or $\collen{i}{\markW}$ for each $i\in\blowvertices$.
Cut-compliant vocabularies are in one-to-one correspondence with colourings of $\blowgraph$: colouring $\colouring$ corresponds to
\begin{align}
    \vocabcanon
    \defeq
    \alphabet
    \cup
    \vocabboundary
    \cup
    \{\collen{i}{\colouring(i)}\mid i\in\blowvertices\},
    \label{eq:unary_apx_canonical}
\end{align}
which has exactly $|\alphabet|+\vocabsize=2\vcnvertices^2+3$ tokens.
Under $\vocabcanon$, an edge-string $\edgelenij{i}{j}$ is compressed to two symbols, $\langle\collen{i}{\markB},\collen{j}{\markW}\rangle$ or $\langle\collen{i}{\markW},\collen{j}{\markB}\rangle$, if $\colouring(i)\neq\colouring(j)$, and to three symbols otherwise.
 
\paragraph{Why the blow-up?}
We aim to show that no vocabulary beats all cut-compliant ones; the danger is non-compliant tokens, which may each serve many edge-strings at once.
Our labels bound the savings of a set $\vocabnc$ of non-compliant tokens by a degree term, $3\vcnvertices|\vocabnc|$ in $\blowgraph$, and a term $\sqrt{|\vocabnc|}|\vocabnc|$ (\cref{sublemma:pair_sum}).
Without the blow-up, the degree of each vertex is a constant $3$ while $|\vocabnc|$ can grow with the graph, so no constant multiplicities pay for non-compliant tokens; yet non-constant multiplicities destroy the constant ratio between the tokenisation and the cut objectives. 
With $\vcnvertices$ clones per vertex, both the degree $3\vcnvertices$ and $\sqrt{|\vocabnc|}\leq\sqrt{2}\vcnvertices$ are $O(\vcnvertices)$, so multiplicities linear in $\vcnvertices$ suffice.
Both the full objective $\basecost$ and the amplified cut $\vcnvertices^2\maxcutof{\srcgraph}$ are then $\Theta(\vcnvertices^3)$, which preserves a constant approximation gap.

\subsection{Unary Direct Tokenisation is Hard to Approximate}
\label{sec:unary_apx_theorem}
 
\Cref{red:unary_apx} is not an L-reduction: the optimum of $\reductionfuncfivefull$ is $\Theta(\vcnvertices^2)$ times larger than $\maxcutof{\srcgraph}$, so it cannot be an L-reduction with constant parameters.
Its two L-reduction constants are, however, $O(\vcnvertices^2)$ and $\vcnvertices^{-2}$, with constant product.
This is exactly the situation handled by AP-reductions (\citealp{crescenzi1997short}; see \cref{def:ap_reduction} in \cref{appendix:ap_reductions}), which also preserve \apx-hardness.
 
\begin{theorem}
\label{thm:unary_apx}
Unary direct tokenisation under the compression objective is \apx-hard.
In particular, unless \pequalnp, it admits no polynomial-time approximation scheme.
\end{theorem}
\begin{proof}[Proof sketch]
This proof is done in three steps.
 
\textbf{Forward step} (cut-compliant vocabularies encode cuts).
See a formal proof in \cref{lemma:unary_apx_forward} in \cref{appendix:unary_apx_forward}.
Under the cut-compliant vocabulary $\vocabcanon$, every marker- and vertex-string is a single token, the two colour-strings of vertex $i$ cost $1+2$ symbols, and its bridge-string costs $2$.
An edge-string costs $2$ symbols if its endpoints have different colours and $3$ otherwise.
Hence $\objectivefunclength(\directtoken[\vocabcanon],\dataset)=\basecost-\cutsize{\blowgraph}{\colouring}$.
Colourings induced from $\srcgraph$ cut $\vcnvertices^2$ times as many edges, giving $\objectivefunclengthopt(\dataset)\leq\basecost-\vcnvertices^2\maxcutof{\srcgraph}$.
 
\textbf{Backward step} (every vocabulary decodes to a cut).
See a formal proof in \cref{lemma:unary_apx_backward} in \cref{appendix:unary_apx_backward}.
Given any vocabulary $\vocab$, we compute in polynomial time a colouring of $\srcgraph$ with cut size $\cutvalue$ such that $\objectivefunclength(\directtoken[\vocab],\dataset)\geq\basecost-\vcnvertices^2\cutvalue$.
If a marker is missing, $\dataset_1$ alone pushes the cost above $\basecost$.
Otherwise, we count whole-string savings on each subdataset, bound the edge-strings compressed by non-compliant tokens with our label arithmetic, and round the colour tokens in $\vocab$ to a single colouring of $\srcgraph$.
Each non-compliant token costs a vocabulary slot, which is charged $\nrepeatvar''+\vcnvertices=16\vcnvertices$ symbols, while all non-compliant tokens together save at most $12\vcnvertices+3$ symbols per token.
 
\textbf{Approximation accounting.}
See a formal proof in \cref{lemma:unary_apx_ap} in \cref{appendix:unary_apx_ap}.
The two steps together give $\objectivefunclengthopt(\dataset)=\basecost-\vcnvertices^2\maxcutof{\srcgraph}\leq106\vcnvertices^2\maxcutof{\srcgraph}$ and $\vcnvertices^2(\maxcutof{\srcgraph}-\cutvalue)\leq\objectivefunclength(\directtoken[\vocab],\dataset)-\objectivefunclengthopt(\dataset)$.
A tokeniser within ratio $1+\varepsilon$ of optimal thus decodes to a cut within ratio roughly $1+106\varepsilon$, which yields an AP-reduction with constant $\apconst=212$.
Since \cubicmaxcut is \apx-hard \citep{alimonti2000some}, so is \dutok.
\end{proof}

\subsection{Proof of Forward Step of \texorpdfstring{\cref{thm:unary_apx}}{Theorem \getrefnumber{thm:unary_apx}}}
\label{appendix:unary_apx_forward}
 
\begin{lemma}
\label{lemma:unary_apx_forward}
Let $(\dataset,\vocabsize)=\reductionfuncfivefull$.
For every colouring $\colouring$ of $\blowgraph$, the cut-compliant vocabulary $\vocabcanon$ satisfies
\begin{align}
    \objectivefunclength(\directtoken[\vocabcanon],\dataset)=\basecost-\cutsize{\blowgraph}{\colouring}.
    \label{eq:unary_apx_forward}
\end{align}
Consequently, $\objectivefunclengthopt(\dataset)\leq\basecost-\vcnvertices^2\maxcutof{\srcgraph}$.
\end{lemma}
 
\begin{proof}
Note that $|\vocabcanon|=|\alphabet|+\vocabsize$, as required.
We write $\opp{\theta}$ for the marker other than $\theta$.
By \cref{sublemma:compliant_structure}\ref{item:compl_distinct}, a string is compressed to one symbol if and only if its length is in $\vocabcanon$.
Under this tokeniser, we have: 
\begin{subequations}
\begin{align}
    \directtoken[\vocabcanon](\dataset_1) &= \big\{\langle\markB\rangle, \langle\markW\rangle \text{ (length 1)}\big\}\times\nrepeatvar\vcnvertices^2\\
    \directtoken[\vocabcanon](\dataset_2) &= \big\{\langle\vtxlen{i}\rangle \text{ (length 1)}\mid i\in\blowvertices\big\}\times\nrepeatvar'\\
    \directtoken[\vocabcanon](\dataset_3) &= \big\{\langle\collen{i}{\colouring(i)}\rangle \text{ (length 1)}, \langle\vtxlen{i},\opp{\colouring(i)}\rangle \text{ (length 2)}\mid i\in\blowvertices\big\}\times\nrepeatvar''\\
    \directtoken[\vocabcanon](\dataset_4) &= \big\{\langle\collen{i}{\colouring(i)},\opp{\colouring(i)}\rangle \text{ (length 2)}\mid i\in\blowvertices\big\}\times1\\
    \directtoken[\vocabcanon](\edgelenij{i}{j}) &=
    \begin{cases}
        \langle\collen{i}{\colouring(i)},\collen{j}{\colouring(j)}\rangle \text{ (length 2)} & \text{if } \colouring(i)\neq\colouring(j)\\
        \langle\collen{i}{\colouring(i)},\vtxlen{j},\opp{\colouring(i)}\rangle \text{ (length 3)} & \text{else}
    \end{cases}
    \quad\text{for } \{i,j\}\in\blowedges
\end{align}
\end{subequations}
The strings of length $2$ cannot be compressed to one symbol, as their lengths are not in $\vocabcanon$.
An uncut edge-string cannot be compressed to two symbols: by \cref{sublemma:compliant_structure}\ref{item:compl_pairs}, this would require $\collen{i}{\markB},\collen{j}{\markW}\in\vocabcanon$ or $\collen{i}{\markW},\collen{j}{\markB}\in\vocabcanon$, i.e., $\colouring(i)\neq\colouring(j)$.
Summing these compressed lengths, we get
\begin{align}
    \objectivefunclength(\directtoken[\vocabcanon],\dataset)
    &= (2\nrepeatvar+\nrepeatvar'+3\nrepeatvar''+2)\vcnvertices^2+3\vcmedges\vcnvertices^2-\cutsize{\blowgraph}{\colouring}
    = \basecost-\cutsize{\blowgraph}{\colouring}.
\end{align}
Finally, let $\srccolouring$ be a maximum cut of $\srcgraph$ and $\colouring$ the colouring of $\blowgraph$ it induces.
Then $\cutsize{\blowgraph}{\colouring}=\vcnvertices^2\maxcutof{\srcgraph}$, which proves the bound on the optimum.
\end{proof}

\subsection{Proof of Backward Step of \texorpdfstring{\cref{thm:unary_apx}}{Theorem \getrefnumber{thm:unary_apx}}}
\label{appendix:unary_apx_backward}
 
 Before starting our lemma's proof, we define a few notions for an arbitrary feasible vocabulary $\vocab$, with $\alphabet\subseteq\vocab$ and $|\vocab|\leq|\alphabet|+\vocabsize=2\vcnvertices^2+3$.
As in the proofs of \citet{kastreva2026bounded}, we write $\vocabnc\defeq\vocab\setminus\vocabcompl$ for its non-compliant tokens and $\vocabmissing\defeq\vocabboundary\setminus\vocab$ for its missing structural tokens.
Further, each vertex $i\in\blowvertices$ gets the palette $\palette{i}\defeq\{\theta\in\{\markB,\markW\}\mid\collen{i}{\theta}\in\vocab\}$ of colours whose colour-string is a token; note that $\sum_i|\palette{i}|=|\vocab\cap\vocabchoice|$.
Let $\nbichrom$ count the edges of $\blowgraph$ whose palettes permit unequal colours, as in \cref{sublemma:rounding}.
We define the decoder $\decodefunc(\vocab)$ as follows: if $\markB\notin\vocab$ or $\markW\notin\vocab$, it returns a locally maximal colouring of $\srcgraph$ (\cref{fact:local_max_cut}); otherwise, it returns the colouring of \cref{sublemma:rounding} applied to the palettes $\palette{i}$. Both branches run in deterministic polynomial time.
 
\begin{lemma}
\label{lemma:unary_apx_backward}
Let $(\dataset,\vocabsize)=\reductionfuncfivefull$, let $\vocab$ be any feasible vocabulary, and let $\cutvalue=\cutsize{\srcgraph}{\decodefunc(\vocab)}$.
Then
\begin{align}
    \objectivefunclength(\directtoken[\vocab],\dataset)\geq\basecost-\vcnvertices^2\cutvalue.
    \label{eq:unary_apx_backward}
\end{align}
\end{lemma}
 
\begin{proof}
We proceed in five steps:
\circled{1} we show that a vocabulary missing a marker costs more than $\basecost$;
\circled{2} we relate non-compliant tokens to missing compliant tokens via the vocabulary budget;
\circled{3} we lower-bound the compressed length of $\dataset_1$ to $\dataset_4$ by counting whole-string tokens;
\circled{4} we lower-bound the compressed length of $\dataset_5$ by bounding the edge-strings compressed to one or two symbols;
\circled{5} we combine these bounds with \cref{sublemma:rounding}.
 
\begin{mylemmastep}
If $\markB\notin\vocab$ or $\markW\notin\vocab$, then $\objectivefunclength(\directtoken[\vocab],\dataset)>\basecost$.
\end{mylemmastep}
 
\begin{proof}
Each of the $\nrepeatvar\vcnvertices^2$ copies of the missing marker costs at least two symbols, and every other string costs at least one.
With $\nrepeatvar-\nrepeatvar''-1=5\vcnvertices-1$ and $2\vcmedges=3\vcnvertices$:
\begin{subequations}
\begin{align}
    \objectivefunclength(\directtoken[\vocab],\dataset)
    &\geq (3\nrepeatvar+\nrepeatvar'+2\nrepeatvar''+1)\vcnvertices^2+\vcmedges\vcnvertices^2\\
    &= \basecost+(\nrepeatvar-\nrepeatvar''-1)\vcnvertices^2-2\vcmedges\vcnvertices^2\\
    &= \basecost+(2\vcnvertices-1)\vcnvertices^2>\basecost.
\end{align}
\end{subequations}
As $\cutvalue\geq0$, \cref{eq:unary_apx_backward} holds in this case.
\end{proof}

From now on, we assume $\markB,\markW\in\vocab$, so $\vocabmissing$ contains only vertex-strings.
 
\begin{mylemmastep}
The vocabulary budget implies
\begin{align}
    \vcnvertices^2-|\vocab\cap\vocabchoice|\geq|\vocabnc|-|\vocabmissing|
    \qquad\text{and}\qquad
    |\vocabnc|\leq2\vcnvertices^2.
    \label{eq:unary_apx_budget}
\end{align}
\end{mylemmastep}
 
\begin{proof}
The vocabulary partitions into $\alphabet$, the $2+\vcnvertices^2-|\vocabmissing|$ structural tokens it contains, $\vocab\cap\vocabchoice$, and $\vocabnc$.
Its size is at most $2\vcnvertices^2+3$, so
\begin{align}
    3+\vcnvertices^2-|\vocabmissing|+|\vocab\cap\vocabchoice|+|\vocabnc|\leq2\vcnvertices^2+3.
\end{align}
Rearranging gives the first claim; the second follows from $|\vocabmissing|\leq\vcnvertices^2$.
\end{proof}
 
\begin{mylemmastep}
The compressed lengths of the first four subdatasets satisfy:
\begin{subequations}
\begin{align}
    \objectivefunclength(\directtoken[\vocab],\dataset_1) &= 2\nrepeatvar\vcnvertices^2,\\
    \objectivefunclength(\directtoken[\vocab],\dataset_2) &\geq \nrepeatvar'\big(\vcnvertices^2+|\vocabmissing|\big),\\
    \objectivefunclength(\directtoken[\vocab],\dataset_3) &\geq \nrepeatvar''\big(4\vcnvertices^2-|\vocab\cap\vocabchoice|\big),\\
    \objectivefunclength(\directtoken[\vocab],\dataset_4) &\geq 2\vcnvertices^2-|\vocabnc|.
\end{align}
\end{subequations}
\end{mylemmastep}
 
\begin{proof}
A string costs one symbol if its length is in $\vocab$, and at least two otherwise.
Both markers are tokens.
Exactly $|\vocabmissing|$ vertex-strings and $2\vcnvertices^2-|\vocab\cap\vocabchoice|$ colour-strings are not tokens, each repeated $\nrepeatvar'$ and $\nrepeatvar''$ times, respectively; this gives the bounds on $\dataset_2$ and $\dataset_3$.
Finally, bridge-strings are not compliant (\cref{sublemma:compliant_structure}\ref{item:compl_distinct}), so at most $|\vocabnc|$ of them are tokens.
\end{proof}
 
\begin{mylemmastep}
The compressed length of the edge-strings satisfies
\begin{align}
    \objectivefunclength(\directtoken[\vocab],\dataset_5)\geq3\vcmedges\vcnvertices^2-\nbichrom-(12\vcnvertices+2)|\vocabnc|.
    \label{eq:unary_apx_edges}
\end{align}
\end{mylemmastep}
 
\begin{proof}
Let $\blowedges^{1}$ and $\blowedges^{2}$ be the sets of edges whose strings are compressed to one and two symbols, respectively; all other edge-strings cost at least three.
Hence $\objectivefunclength(\directtoken[\vocab],\dataset_5)\geq3\vcmedges\vcnvertices^2-2|\blowedges^{1}|-|\blowedges^{2}|$.
An edge in $\blowedges^{1}$ has $\edgelen{e}\in\vocab$, and since $\edgelen{e}$ is not compliant, $\edgelen{e}\in\vocabnc$; so $|\blowedges^{1}|\leq|\vocabnc|$.
An edge in $\blowedges^{2}$ has $\edgelen{e}=p+p'$ for two tokens $p,p'\in\vocab$, and we distinguish three cases:
\begin{itemize}
    \item \textit{Both compliant.} By \cref{sublemma:compliant_structure}\ref{item:compl_pairs}, $\{p,p'\}$ is $\{\collen{i}{\markB},\collen{j}{\markW}\}$ or $\{\collen{i}{\markW},\collen{j}{\markB}\}$, so the vocabulary contains colour tokens for the two endpoints with opposite colours. There are at most $\nbichrom$ such edges.
    \item \textit{One compliant, one non-compliant.} By \cref{sublemma:compliant_structure}\ref{item:compl_mixed}, each non-compliant token serves at most $3\vcnvertices$ edges this way, for at most $3\vcnvertices|\vocabnc|$ edges.
    \item \textit{Both non-compliant.} Then $\edgelen{e}\in\vocabnc+\vocabnc$. By \cref{sublemma:pair_sum} with $Q=\vocabnc$, there are at most $(6\vcnvertices+\sqrt{|\vocabnc|}+1)|\vocabnc|$ such edges. As $|\vocabnc|\leq2\vcnvertices^2$ (Step \circled{2}), we have $\sqrt{|\vocabnc|}\leq2\vcnvertices$, and this is at most $9\vcnvertices|\vocabnc|$.
\end{itemize}
Thus $|\blowedges^{2}|\leq\nbichrom+12\vcnvertices|\vocabnc|$, and \cref{eq:unary_apx_edges} follows.
\end{proof}
 
\begin{mylemmastep}
If $\markB,\markW\in\vocab$, then \cref{eq:unary_apx_backward} holds.
\end{mylemmastep}
 
\begin{proof}
Summing Steps \circled{3} and \circled{4}, and subtracting $\basecost$ from \cref{eq:unary_apx_basecost}, we get
\begin{subequations}
\begin{align}
    \objectivefunclength(\directtoken[\vocab],\dataset)-\basecost
    &\geq
    \nrepeatvar'|\vocabmissing|+\nrepeatvar''\big(\vcnvertices^2-|\vocab\cap\vocabchoice|\big)\nonumber\\
    &\quad-\nbichrom-(12\vcnvertices+3)|\vocabnc|.
\end{align}
\end{subequations}
By \cref{sublemma:rounding}, $\cutvalue=\cutsize{\srcgraph}{\decodefunc(\vocab)}$ satisfies $-\nbichrom\geq-\vcnvertices^2\cutvalue+\vcnvertices(\vcnvertices^2-|\vocab\cap\vocabchoice|)$.
Then, as $\nrepeatvar''+\vcnvertices=16\vcnvertices>0$, \cref{eq:unary_apx_budget} gives:
\begin{align}
    \objectivefunclength(\directtoken[\vocab],\dataset)-\basecost+\vcnvertices^2\cutvalue
    &\geq
    \nrepeatvar'|\vocabmissing|+(\nrepeatvar''+\vcnvertices)\big(\vcnvertices^2-|\vocab\cap\vocabchoice|\big)\nonumber\\
    &\qquad\qquad-(12\vcnvertices+3)|\vocabnc|\\
    &\geq
    17\vcnvertices|\vocabmissing|+16\vcnvertices\big(|\vocabnc|-|\vocabmissing|\big)-(12\vcnvertices+3)|\vocabnc|\\
    &=
    \vcnvertices|\vocabmissing|+(4\vcnvertices-3)|\vocabnc|
    \geq 0.
\end{align}
In words: after rounding, each non-compliant token occupies a slot charged $16\vcnvertices$ symbols, while all non-compliant tokens together save at most $12\vcnvertices+3$ symbols per token on edge- and bridge-strings.
\end{proof}
 
Steps \circled{1} and \circled{5} cover both cases, which concludes the proof.
\end{proof}
 
\begin{corollary}
\label{cor:unary_apx_opt}
For $(\dataset,\vocabsize)=\reductionfuncfivefull$, we have $\objectivefunclengthopt(\dataset)=\basecost-\vcnvertices^2\maxcutof{\srcgraph}$, and every feasible $\vocab$ satisfies
\begin{align}
    \vcnvertices^2\big(\maxcutof{\srcgraph}-\cutsize{\srcgraph}{\decodefunc(\vocab)}\big)
    \leq
    \objectivefunclength(\directtoken[\vocab],\dataset)-\objectivefunclengthopt(\dataset).
    \label{eq:unary_apx_error}
\end{align}
\end{corollary}
 
\begin{proof}
\cref{lemma:unary_apx_backward} and $\cutsize{\srcgraph}{\decodefunc(\vocab)}\leq\maxcutof{\srcgraph}$ give $\objectivefunclengthopt(\dataset)\geq\basecost-\vcnvertices^2\maxcutof{\srcgraph}$; \cref{lemma:unary_apx_forward} gives the reverse inequality.
Substituting the optimum into \cref{eq:unary_apx_backward} gives \cref{eq:unary_apx_error}.
\end{proof}

\subsection{Proof of \texorpdfstring{\cref{thm:unary_apx}}{Theorem \getrefnumber{thm:unary_apx}} (Unary Direct Tokenisation is APX-hard)}
\label{appendix:unary_apx_ap}
 
\begin{lemma}
\label{lemma:unary_apx_ap}
\cubicmaxcut AP-reduces to \dutok with constant $\apconst=212$.
\end{lemma}
 
\begin{proof}
We define $\ffunctionap(\srcgraph,\targetratio)\defeq\reductionfuncfivefull$, ignoring $\targetratio$; it outputs $O(\vcnvertices^3)$ unary strings of length $O(\vcnvertices^{120})$, so it is computable in polynomial time.
We define $\gfunctionap(\srcgraph,\vocab,\targetratio)$ as whichever of the two colourings $\decodefunc(\vocab)$ and $\colouring_{\mathrm{loc}}$ cuts more edges of $\srcgraph$ (either, in case of a tie), where $\colouring_{\mathrm{loc}}$ is the locally maximal colouring returned by the local search of \cref{fact:local_max_cut}; both are computable in polynomial time.
Let $\cutvalue$ be the cut size of this output.
It remains to show condition (iii) of \cref{def:ap_reduction}.
 
Recall that $\vcnvertices\geq4$ and, by \cref{eq:cubic_cut_bounds}, $\maxcutof{\srcgraph}\geq\vcnvertices$.
Then, by \cref{cor:unary_apx_opt} and \cref{eq:unary_apx_basecost}:
\begin{align}
    \objectivefunclengthopt(\dataset)
    \leq
    \basecost-\vcnvertices^3
    =
    \frac{211}{2}\vcnvertices^3+2\vcnvertices^2
    \leq
    106\vcnvertices^3
    \leq
    106\vcnvertices^2\maxcutof{\srcgraph},
    \label{eq:unary_apx_alpha}
\end{align}
where the second inequality uses $\vcnvertices\geq4$.
Now let $\vocab$ achieve ratio at most $\targetratio=1+\varepsilon$, i.e., $\objectivefunclength(\directtoken[\vocab],\dataset)-\objectivefunclengthopt(\dataset)\leq\varepsilon\objectivefunclengthopt(\dataset)$.
Combining \cref{eq:unary_apx_error,eq:unary_apx_alpha} and dividing by $\vcnvertices^2$:
\begin{align}
    \maxcutof{\srcgraph}-\cutvalue
    \leq
    \maxcutof{\srcgraph}-\cutsize{\srcgraph}{\decodefunc(\vocab)}
    \leq
    106\varepsilon\maxcutof{\srcgraph}.
\end{align}
If $\varepsilon\leq1/212$, then
\begin{align}
    \frac{\maxcutof{\srcgraph}}{\cutvalue}
    \leq
    \frac{1}{1-106\varepsilon}
    \leq
    1+212\varepsilon,
\end{align}
where the last inequality is equivalent to $106\varepsilon(1-212\varepsilon)\geq0$.
If $\varepsilon>1/212$, the locally maximal colouring gives $\cutvalue\geq\vcnvertices\geq\frac{2}{3}\maxcutof{\srcgraph}$, as $\maxcutof{\srcgraph}\leq\vcmedges=\frac{3}{2}\vcnvertices$; so $\maxcutof{\srcgraph}/\cutvalue\leq3/2<1+212\varepsilon$.
In both cases, $\perfratio(\srcgraph,\gfunctionap(\srcgraph,\vocab,\targetratio))\leq1+212(\targetratio-1)$.
\end{proof}
 
\begin{proof}[Proof of \cref{thm:unary_apx}]
\cubicmaxcut is \apx-hard \citep{alimonti2000some}, and \cref{lemma:unary_apx_ap} gives an AP-reduction from it to \dutok; hence \dutok is \apx-hard.
Explicitly, a polynomial-time approximation scheme for \dutok, run with $\varepsilon=\eta/106$, would yield a cut of size at least $(1-\eta)\maxcutof{\srcgraph}$ for every fixed $\eta>0$, contradicting the hardness of \cubicmaxcut unless \pequalnp.
\end{proof}

\subsection{Arithmetic of \texorpdfstring{\cref{red:unary_apx}}{Reduction \getrefnumber{red:unary_apx}}}
\label{appendix:unary_apx_arithmetic}
 
We now show the technical sublemmas used throughout our proof.

\begin{sublemma}[Label uniqueness]
\label{sublemma:label_uniqueness}
Let $\mathcal{J}$ and $\mathcal{J}'$ be multisets of at most four indices in $\blowvertices$ each.
If $\sum_{i\in\mathcal{J}}\enc{i}=\sum_{i\in\mathcal{J}'}\enc{i}$, then $\mathcal{J}=\mathcal{J}'$.
Moreover, $\enc{i}<\labelbase^5$ for every $i\in\blowvertices$.
\end{sublemma}
 
\begin{proof}
Let a number $(d_4,d_3,d_2,d_1,d_0)_{\labelbase}$ denote $\sum_{k=0}^{4}d_k\labelbase^k$.
We have $\enc{i}=(i^4,i^3,i^2,i,0)_{\labelbase}$, and in a sum of at most four labels each digit is at most $4\vcnvertices^8<\labelbase$.
No carries thus occur, and the equality of the two sums implies equality of each digit:
\begin{align}
    \sum_{i\in\mathcal{J}}i^k=\sum_{i\in\mathcal{J}'}i^k
    \qquad\text{for } k\in\{1,2,3,4\}.
\end{align}
Pad both multisets with zeros to size four; this does not change the power sums above.
By Newton's identities, the first four power sums of four numbers determine their elementary symmetric polynomials, and hence the monic polynomial whose roots they are.
The padded multisets are therefore equal, and removing the zeros gives $\mathcal{J}=\mathcal{J}'$.
Finally, $i^k\leq\vcnvertices^8<\labelbase/8$ for $1\leq k\leq4$, so $\enc{i}<\frac{\labelbase}{8}(\labelbase+\labelbase^2+\labelbase^3+\labelbase^4)<\labelbase^5$.
\end{proof}

Every length in \cref{red:unary_apx} is an integer combination of the unit $1$, the two markers, and the labels.
For such a combination $x=c_1+c_{\markB}\markB+c_{\markW}\markW+\sum_{i\in\blowvertices}c_i\enc{i}$, we call $\mathbf{c}(x)\defeq\big(c_1,c_{\markB},c_{\markW},(c_i)_{i\in\blowvertices}\big)$ its \defn{formal coefficients}.
For an integer vector $\mathbf{d}=(d_i)_{i\in\blowvertices}$, we write $d_i^{+}\defeq\max(d_i,0)$ and $d_i^{-}\defeq\max(-d_i,0)$, so that $d_i=d_i^{+}-d_i^{-}$ and $d_i^{+}d_i^{-}=0$.
\begin{sublemma}[Coefficient matching]
\label{sublemma:coefficient_matching}
Let $\alpha_1,\alpha_{\markB},\alpha_{\markW}\in\mathbb{Z}$ and $\mathbf{d}\in\mathbb{Z}^{\blowvertices}$ satisfy
\begin{align}
    |\alpha_1|,|\alpha_{\markB}|,|\alpha_{\markW}|\leq2,
    \qquad
    \sum_{i\in\blowvertices}d_i^{+}\leq4,
    \qquad
    \sum_{i\in\blowvertices}d_i^{-}\leq4.
    \label{eq:coef_bounds}
\end{align}
If
\begin{align}
    \alpha_1+\alpha_{\markB}\markB+\alpha_{\markW}\markW+\sum_{i\in\blowvertices}d_i\enc{i}=0,
    \label{eq:formal_zero}
\end{align}
then $\alpha_1=\alpha_{\markB}=\alpha_{\markW}=0$ and $\mathbf{d}=\mathbf{0}$.
\end{sublemma}

\begin{proof}
We use one elementary fact three times: for integers $a$ and $R$ and a positive integer $Y$,
\begin{align}
    aY+R=0
    \quad\text{and}\quad
    |R|<Y
    \implies
    a=R=0,
    \label{eq:cancel}
\end{align}
since $a\neq0$ would give $|aY|\geq Y>|R|=|aY|$.
Write the vertex part as $V\defeq\sum_id_i\enc{i}=P-M$, where
\begin{align}
    P\defeq\sum_{i\in\blowvertices}d_i^{+}\enc{i},
    \qquad
    M\defeq\sum_{i\in\blowvertices}d_i^{-}\enc{i}.
\end{align}
Each of $P$ and $M$ is a sum of at most four labels by \cref{eq:coef_bounds}, and every label satisfies $0<\enc{i}<\labelbase^5$ (\cref{sublemma:label_uniqueness}).
Hence $0\leq P,M<4\labelbase^5$, and
\begin{align}
    |V|<4\labelbase^5.
    \label{eq:vertex_part_bound}
\end{align}
We also use $\labelbase=8\vcnvertices^8+1\geq9$.

\begin{mysublemmastep}
 $\alpha_{\markW}=0$.
\end{mysublemmastep}

\begin{proof}
Rewrite \cref{eq:formal_zero} as $\alpha_{\markW}\markW+R_{\markW}=0$ with $R_{\markW}\defeq\alpha_1+\alpha_{\markB}\markB+V$.
By \cref{eq:coef_bounds,eq:vertex_part_bound},
\begin{align}
    |R_{\markW}|
    \leq |\alpha_1|+|\alpha_{\markB}|\markB+|V|
    < 2+2\labelbase^{10}+4\labelbase^5
    \leq 3\labelbase^{10}
    < \labelbase^{15}
    =\markW.
\end{align}
By \cref{eq:cancel} with $Y=\markW$, we get $\alpha_{\markW}=0$.
\end{proof}

\begin{mysublemmastep}
 $\alpha_{\markB}=0$.
\end{mysublemmastep}

\begin{proof}
With $\alpha_{\markW}=0$, \cref{eq:formal_zero} reads $\alpha_{\markB}\markB+R_{\markB}=0$ with $R_{\markB}\defeq\alpha_1+V$, and
\begin{align}
    |R_{\markB}|
    \leq |\alpha_1|+|V|
    < 2+4\labelbase^5
    < \labelbase^{10}
    =\markB.
\end{align}
By \cref{eq:cancel} with $Y=\markB$, we get $\alpha_{\markB}=0$.
\end{proof}

\begin{mysublemmastep}
 $\alpha_1=0$ and $V=0$.
\end{mysublemmastep}

\begin{proof}
Every label is a multiple of $\labelbase$: $\enc{i}=\labelbase\sum_{k=1}^{4}i^k\labelbase^{k-1}$.
Hence $V=\labelbase V'$ with $V'\defeq\sum_id_i\sum_{k=1}^{4}i^k\labelbase^{k-1}\in\mathbb{Z}$, and \cref{eq:formal_zero} reads $V'\labelbase+\alpha_1=0$.
As $|\alpha_1|\leq2<\labelbase$, \cref{eq:cancel} with $Y=\labelbase$ gives $V'=0$ and $\alpha_1=0$, so $V=0$.
\end{proof}

\begin{mysublemmastep}
 $\mathbf{d}=\mathbf{0}$.
\end{mysublemmastep}

\begin{proof}
From $V=0$ we get $P=M$, i.e.,
\begin{align}
    \sum_{i\in\blowvertices}d_i^{+}\enc{i}
    =
    \sum_{i\in\blowvertices}d_i^{-}\enc{i}.
\end{align}
Let $\mathcal{J}^{+}$ and $\mathcal{J}^{-}$ be the multisets containing each index $i$ with multiplicity $d_i^{+}$ and $d_i^{-}$, respectively; both have size at most four.
By \cref{sublemma:label_uniqueness}, $\mathcal{J}^{+}=\mathcal{J}^{-}$, so $d_i^{+}=d_i^{-}$ for every $i$.
Since $d_i^{+}d_i^{-}=0$, both are zero, and hence $d_i=0$ for every $i$.
\end{proof}

Steps \circled{1} to \circled{4} prove the sublemma.
\end{proof}

We now check that \cref{sublemma:coefficient_matching} applies to every comparison we need.
\Cref{tab:formal_coefficients} lists the formal coefficients of all lengths in $\dataset\cup\vocabcompl$, where $\mathbf{e}_i$ denotes the $i$-th unit vector.

\begin{table}[h]
\centering
\begin{tabular}{lcccc}
\toprule
Length $x$ & $c_1$ & $c_{\markB}$ & $c_{\markW}$ & $(c_i)_{i\in\blowvertices}$ \\
\midrule
$1$ & $1$ & $0$ & $0$ & $\mathbf{0}$ \\
$\markB$ & $0$ & $1$ & $0$ & $\mathbf{0}$ \\
$\markW$ & $0$ & $0$ & $1$ & $\mathbf{0}$ \\
$\vtxlen{i}$ & $0$ & $0$ & $0$ & $\mathbf{e}_i$ \\
$\collen{i}{\markB}$ & $0$ & $1$ & $0$ & $\mathbf{e}_i$ \\
$\collen{i}{\markW}$ & $0$ & $0$ & $1$ & $\mathbf{e}_i$ \\
$\bridgelen{i}$ & $0$ & $1$ & $1$ & $\mathbf{e}_i$ \\
$\edgelenij{i}{j}$ & $0$ & $1$ & $1$ & $\mathbf{e}_i+\mathbf{e}_j$ \\
\bottomrule
\end{tabular}
\caption{Formal coefficients of the lengths in \cref{red:unary_apx}.}
\label{tab:formal_coefficients}
\end{table}

Every entry of the table has $c_1,c_{\markB},c_{\markW}\in\{0,1\}$ and a non-negative label vector with $\sum_ic_i\leq2$.
We consider three kinds of equalities, and in each case let $(\alpha_1,\alpha_{\markB},\alpha_{\markW},\mathbf{d})$ be the difference of the formal coefficients of the two sides:
\begin{enumerate}[label=(\roman*)]
    \item $x=y$ for $x,y\in\dataset\cup\vocabcompl$.
    Each coefficient of the difference lies in $\{-1,0,1\}$, and $\sum_id_i^{+}\leq2$ and $\sum_id_i^{-}\leq2$.
    \item $\edgelen{e}=p+p'$ for $p,p'\in\vocabcompl$.
    Here $\alpha_1\in\{-2,-1,0\}$ and $\alpha_{\markB},\alpha_{\markW}\in\{-1,0,1\}$; the positive label part comes from $\edgelen{e}$ alone and the negative one from $p+p'$, so $\sum_id_i^{+}\leq2$ and $\sum_id_i^{-}\leq2$.
    \item $\edgelen{e}-p=\edgelen{f}-p'$ for $p,p'\in\vocabcompl$, i.e., $\edgelen{e}+p'=\edgelen{f}+p$.
    Each side has unit coefficient in $\{0,1\}$, marker coefficients in $\{1,2\}$, and label mass at most $3$; hence $\alpha_1,\alpha_{\markB},\alpha_{\markW}\in\{-1,0,1\}$, $\sum_id_i^{+}\leq3$, and $\sum_id_i^{-}\leq3$.
\end{enumerate}
In all three cases \cref{eq:coef_bounds} holds, so \cref{sublemma:coefficient_matching} gives $\mathbf{c}(\text{left-hand side})=\mathbf{c}(\text{right-hand side})$.
In words, a numerical equality of any of these three kinds holds only if the two sides agree as \defn{formal} expressions, and we may compare them coefficient by coefficient.
\begin{sublemma}[Compliant-token structure]
\label{sublemma:compliant_structure}
The following hold for the instance output by \cref{red:unary_apx}:
\begin{enumerate}[label=(\alph*)]
    \item \label{item:compl_distinct}
    The distinct strings in $\dataset$ have distinct lengths, the $3+3\vcnvertices^2$ compliant tokens are distinct, and $\bridgelen{i}\notin\vocabcompl$ and $\edgelen{e}\notin\vocabcompl$ for all $i$ and $e$.
    \item \label{item:compl_pairs}
    For $e=\{i,j\}\in\blowedges$, the only ways to write $\edgelen{e}=p+p'$ with $p,p'\in\vocabcompl$ are, up to order, $\collen{i}{\markB}+\collen{j}{\markW}$ and $\collen{i}{\markW}+\collen{j}{\markB}$.
    \item \label{item:compl_mixed}
    For every $q\in\N\setminus\vocabcompl$, at most $3\vcnvertices$ edges $e\in\blowedges$ satisfy $\edgelen{e}-q\in\vocabcompl$.
\end{enumerate}
\end{sublemma}
 
\begin{proof}
By \cref{sublemma:coefficient_matching}, all comparisons below can be made formally.
 
\textit{Part \ref{item:compl_distinct}.}
Each string-length has a distinct formal pattern: its marker coefficients identify the family, and its label coefficients identify the index or the edge.
Bridge- and edge-strings contain both markers, which no compliant token does.
 
\textit{Part \ref{item:compl_pairs}.}
The formal expression $\edgelen{e}$ has marker coefficients $(1,1)$, unit coefficient $0$, and label coefficients $1$ at exactly $i$ and $j$.
A sum of two compliant tokens matches the marker coefficients only if it takes one $\markB$ and one $\markW$, either from $\{\markB,\markW\}$ themselves or from colour-strings; it matches the unit coefficient only if it avoids $1$; and it matches the label coefficients only if it contributes exactly one label at $i$ and one at $j$.
This leaves only the two decompositions in the statement.

\textit{Part \ref{item:compl_mixed}.}
Fix a non-compliant token $q\notin\vocabcompl$. We bound the number of edges $e\in\blowedges$ for which
\begin{align}
    \edgelen{e}-q\in\vocabcompl.
\end{align}
Equivalently, suppose that for some edge $e=\{i,j\}$ there exists a compliant token $p\in\vocabcompl$ such that
\begin{align}
    q=\edgelen{e}-p.
\end{align}
We classify the possibilities according to the form of $p$.

\begin{itemize}[leftmargin=2.2em]
    \item[($\alpha$)]
    Suppose $p\in\{1,\markB,\markW\}$.
    None of these tokens contains a vertex label, whereas $\edgelen{e}$ contains exactly the two labels $\enc{i}$ and $\enc{j}$.
    Hence the label coefficients of $q$ are $+1$ exactly at $i$ and $j$.
    In particular, the positive label coefficients of $q$ uniquely determine the edge $e=\{i,j\}$.

    \item[($\beta$)]
    Suppose
    \begin{align}
        p\in\{\vtxlen{k},\collen{k}{\markB},\collen{k}{\markW}\}
        \qquad\text{for some }k\notin\{i,j\}.
    \end{align}
    The token $p$ contributes one copy of the label $\enc{k}$, so the label coefficients of
    $q=\edgelen{e}-p$ are $+1$ at $i$ and $j$, $-1$ at $k$, and zero elsewhere.
    Again, the two positive label coefficients uniquely identify the endpoints of $e$.

    \item[($\gamma$)]
    It remains to consider the case where $p$ is associated with an endpoint of $e$.
    Without loss of generality, let
    \begin{align}
        p\in\{\vtxlen{i},\collen{i}{\markB},\collen{i}{\markW}\}.
    \end{align}
    Since
    \begin{align}
        \edgelen{e}
        =
        \enc{i}+\enc{j}+\markB+\markW,
    \end{align}
    the three possible choices of $p$ give
    \begin{align}
        \edgelen{e}-\vtxlen{i}
        &= \enc{j}+\markB+\markW
         = \bridgelen{j},
        \\
        \edgelen{e}-\collen{i}{\markB}
        &= \enc{j}+\markW
         = \collen{j}{\markW},
        \\
        \edgelen{e}-\collen{i}{\markW}
        &= \enc{j}+\markB
         = \collen{j}{\markB}.
    \end{align}
    The latter two values belong to $\vocabcompl$, whereas $q\notin\vocabcompl$ by assumption.
    They are therefore impossible, and necessarily
    \begin{align}
        q=\bridgelen{j}.
    \end{align}
    Thus the fixed token $q$ determines one endpoint $j$ of $e$, but it does not determine the other endpoint.
    Consequently, every edge that yields this same $q$ must be incident to $j$.
    Since every vertex of $\blowgraph$ has degree $3\vcnvertices$, at most $3\vcnvertices$ edges can arise in this case.
\end{itemize}

It remains to verify that the same numerical value of $q$ cannot arise from different cases in a way that evades these bounds.
Suppose
\begin{align}
    q
    =
    \edgelen{e}-p
    =
    \edgelen{f}-p'
\end{align}
for edges $e,f\in\blowedges$ and compliant tokens $p,p'\in\vocabcompl$.
By comparison~(iii) above and \cref{sublemma:coefficient_matching}, this numerical equality is also an equality of formal coefficient vectors.
The three cases have distinct label-coefficient patterns: case~($\alpha$) has two positive coefficients, case~($\beta$) has two positive and one negative coefficient, and case~($\gamma$) has a single positive coefficient.
Hence the two representations of $q$ must fall in the same case.

In cases~($\alpha$) and~($\beta$), the positive label coefficients determine both endpoints of the edge, so necessarily $e=f$.
In case~($\gamma$), the fixed token $q=\bridgelen{j}$ determines one endpoint $j$, so all possible edges are incident to $j$; there are exactly $3\vcnvertices$ such edges in $\blowgraph$.
Therefore, for every $q\notin\vocabcompl$, at most $3\vcnvertices$ edges $e\in\blowedges$ satisfy $\edgelen{e}-q\in\vocabcompl$, as claimed.
\end{proof}

In the backward step, an edge-string may be compressed to two symbols by splitting it into two non-compliant tokens.
The next sublemma bounds how many edge-strings a set $Q$ of tokens can split in this way.
A priori, this number could be as large as the number of pairs, roughly $|Q|^2/2$.
We show that it is instead at most linear in $|Q|$, up to a factor of order $\vcnvertices+\sqrt{|Q|}$.
The reason is that our labels leave the edge-strings with almost no additive structure: the only coincidences between their sums and differences come from edges sharing a vertex.

\begin{sublemma}[Pair-sum bound]
\label{sublemma:pair_sum}
Let $Q\subset\mathbb{Z}$ be a finite set of integers.
Then the number of edge-strings that are the sum of two (not necessarily distinct) elements of $Q$ satisfies
\begin{align}
    \big|\{e\in\blowedges\mid \edgelen{e}=q+q' \text{ for some } q,q'\in Q\}\big|
    \leq
    \big(6\vcnvertices+\sqrt{|Q|}+1\big)|Q|.
    \label{eq:pair_sum}
\end{align}
\end{sublemma}

\begin{proof}
Let $\blowedges^{Q}$ denote the set of edges counted on the left-hand side of \cref{eq:pair_sum}.
We turn the counting problem into a question about a graph.

\paragraph{The split graph.}
For every edge $e\in\blowedges^{Q}$, fix one pair $\{q_e,q'_e\}\subseteq Q$ with $q_e+q'_e=\edgelen{e}$; if several such pairs exist, choose any one of them.
The \defn{split graph} $\Gamma$ has vertex set $Q$ and, for every $e\in\blowedges^{Q}$, one edge between $q_e$ and $q'_e$, which we \defn{label} with $e$.
(To avoid confusion, we call the elements of $Q$ \emph{tokens} and the edges of $\Gamma$ \emph{splits}; ``vertex'' and ``edge'' always refer to $\blowgraph$.)
Two different edges $e\neq f$ cannot produce the same split: that would give $\edgelen{e}=q_e+q'_e=q_f+q'_f=\edgelen{f}$, whereas edge-strings of different edges have different lengths (\cref{sublemma:compliant_structure}\ref{item:compl_distinct}).
Hence $\Gamma$ has no parallel splits, each split carries a different label, and
\begin{align}
    |\blowedges^{Q}| = \text{number of splits of }\Gamma.
\end{align}
A split with $q_e=q'_e$ is a loop.
Each token $q$ has at most one loop, since a loop at $q$ has length $2q$ and hence determines its label; so there are at most $|Q|$ loops.
We set them aside and bound the number of remaining splits, which form a simple graph $\Gamma_0$.

Consider two consecutive splits $q-r$ and $r-z$ in $\Gamma_0$, labelled with edges $e=\{i,j\}$ and $f=\{k,l\}$, respectively.
By definition,
\begin{align}
    q+r=\edgelenij{i}{j}
    \qquad\text{and}\qquad
    r+z=\edgelenij{k}{l}.
\end{align}
Subtracting the first equation from the second, both the middle token $r$ and the markers $\markB+\markW$ cancel:
\begin{align}
    z-q
    =
    \edgelenij{k}{l}-\edgelenij{i}{j}
    =
    \enc{k}+\enc{l}-\enc{i}-\enc{j}.
    \label{eq:walk_difference}
\end{align}
The difference of the outer tokens is thus a signed sum of four labels, and by \cref{sublemma:label_uniqueness} such a sum reveals its indices.
We call the path $q-r-z$ a \defn{disjoint two-step path} if its labels $e$ and $f$ share no vertex.
The first step of the proof shows that such a path is determined by its two ends.

\begin{mysublemmastep}
For any two tokens $q,z\in Q$, there is at most one disjoint two-step path from $q$ to $z$ in $\Gamma_0$.
\end{mysublemmastep}

\begin{proof}
Let $q-r-z$ and $q-r'-z$ be disjoint two-step paths with labels $(e,f)$ and $(e',f')$, respectively.
Applying \cref{eq:walk_difference} to both paths gives
\begin{align}
    \edgelen{f}-\edgelen{e}
    =
    z-q
    =
    \edgelen{f'}-\edgelen{e'}.
\end{align}
Cancelling the markers and moving the negative terms to the other side, we obtain an equality between two sums of four labels:
\begin{align}
    \sum_{i\in f}\enc{i}+\sum_{i\in e'}\enc{i}
    =
    \sum_{i\in f'}\enc{i}+\sum_{i\in e}\enc{i}.
\end{align}
By \cref{sublemma:label_uniqueness}, the two index multisets agree: $f\uplus e'=f'\uplus e$.
Equivalently, the signed indicator vectors agree, $\one_f-\one_e=\one_{f'}-\one_{e'}$.
Since $e\cap f=\emptyset$, the vector $\one_f-\one_e$ equals $+1$ exactly on the two endpoints of $f$ and $-1$ exactly on the two endpoints of $e$; the same holds for $\one_{f'}-\one_{e'}$, since $e'\cap f'=\emptyset$.
Comparing where the two vectors are positive gives $f=f'$, and comparing where they are negative gives $e=e'$.
Finally, the middle token is determined by $q$ and $e$, since $r=\edgelen{e}-q=\edgelen{e'}-q=r'$.
The two paths are therefore identical.
\end{proof}

Step \circled{1} limits how many disjoint two-step paths can leave a single token: they all end at different tokens, and there are only $|Q|$ tokens.
If every token had many splits, there would be too many such paths.
The next step makes this precise.

\begin{mysublemmastep}
Every non-empty subgraph of $\Gamma_0$ contains a token with at most $6\vcnvertices+\sqrt{|Q|}$ splits in that subgraph.
\end{mysublemmastep}

\begin{proof}
Suppose, for contradiction, that some non-empty subgraph $\Gamma'$ of $\Gamma_0$ has minimum degree $d>6\vcnvertices+\sqrt{|Q|}$.
Fix any token $q$ in $\Gamma'$ and count the disjoint two-step paths starting at $q$ within $\Gamma'$.
\begin{itemize}
    \item \textit{First split.} Token $q$ has at least $d$ splits $q-r$ in $\Gamma'$. Fix one, and let $e=\{i,j\}$ be its label.
    \item \textit{Second split.} Token $r$ also has at least $d$ splits in $\Gamma'$. We must discard those whose label shares a vertex with $e$. Each vertex of $\blowgraph$ has degree $3\vcnvertices$, so at most $3\vcnvertices+3\vcnvertices-1=6\vcnvertices-1$ edges of $\blowgraph$ touch $i$ or $j$. As different splits carry different labels, at most $6\vcnvertices-1$ splits at $r$ are discarded, leaving at least $d-6\vcnvertices+1$.
\end{itemize}
Different choices of the two splits give different paths, so at least $d(d-6\vcnvertices+1)$ disjoint two-step paths start at $q$.
By Step \circled{1}, they end at pairwise different tokens, so their number is at most $|Q|$.
However, $d>\sqrt{|Q|}$ and $d-6\vcnvertices+1>\sqrt{|Q|}$, so
\begin{align}
    d(d-6\vcnvertices+1)>\sqrt{|Q|}\cdot\sqrt{|Q|}=|Q|,
\end{align}
a contradiction.
\end{proof}

\begin{mysublemmastep}
$\Gamma_0$ has at most $(6\vcnvertices+\sqrt{|Q|})|Q|$ splits.
\end{mysublemmastep}

\begin{proof}
We delete the tokens of $\Gamma_0$ one by one.
By Step \circled{2}, applied to what remains of $\Gamma_0$ at each point, we can always delete a token that has at most $6\vcnvertices+\sqrt{|Q|}$ remaining splits, and we delete those splits with it.
After $|Q|$ deletions nothing remains, so every split was deleted in some round.
Each round deletes at most $6\vcnvertices+\sqrt{|Q|}$ splits, which gives the bound.
\end{proof}

Adding back the at most $|Q|$ loops, the split graph $\Gamma$ has at most $(6\vcnvertices+\sqrt{|Q|}+1)|Q|$ splits, which proves \cref{eq:pair_sum}.
\end{proof}

The degree term cannot be avoided.
Since $\labelbase$ is odd, $\markB+\markW$ is even, and the $\vcnvertices^2$ tokens $x_i\defeq\enc{i}+\frac{1}{2}(\markB+\markW)$ satisfy $x_i+x_j=\edgelenij{i}{j}$ for every edge $\{i,j\}\in\blowedges$; together they split all $\frac{3}{2}\vcnvertices^3$ edge-strings, that is, $\frac{3}{2}\vcnvertices$ per token.
Non-compliant tokens can thus save $\Theta(\vcnvertices)$ symbols each, so the multiplicities must grow linearly in $\vcnvertices$; the blow-up makes this affordable, as both $\basecost$ and $\vcnvertices^2\maxcutof{\srcgraph}$ are then $\Theta(\vcnvertices^3)$.
\subsection{Rounding Palettes to a Cut}
\label{appendix:unary_apx_rounding}
 
\begin{sublemma}[Rounding]
\label{sublemma:rounding}
Let $\srcgraph$ be a simple cubic graph and $\blowgraph$ its $\vcnvertices$-blow-up.
Give each $i\in\blowvertices$ a \defn{palette} $\palette{i}\subseteq\{\markB,\markW\}$, and let $\nbichrom$ count the edges $\{i,j\}\in\blowedges$ whose palettes permit unequal colours, i.e., for which there exist $\theta\in\palette{i}$ and $\theta'\in\palette{j}$ with $\theta\neq\theta'$.
Then a colouring $\srccolouring$ of $\srcgraph$ with
\begin{align}
    \vcnvertices^2\cutsize{\srcgraph}{\srccolouring}
    \geq
    \nbichrom+\vcnvertices\Big(\vcnvertices^2-\sum_{i\in\blowvertices}|\palette{i}|\Big)
    \label{eq:rounding}
\end{align}
can be found in deterministic polynomial time.
\end{sublemma}
 
\begin{proof}
We proceed in two steps: we first round palettes on $\srcgraph$ itself, and then reduce the blow-up to that case.
 
\begin{mysublemmastep}
Given palettes $\palette{\vertex}$ on the vertices $\vertex\in\vertices$, with $\nbichrom_{\srcgraph}$ edges of $\srcgraph$ permitting unequal colours, we can find in polynomial time a colouring $\srccolouring$ with $\cutsize{\srcgraph}{\srccolouring}\geq\nbichrom_{\srcgraph}+\vcnvertices-\sum_{\vertex\in\vertices}|\palette{\vertex}|$.
\end{mysublemmastep}
 
\begin{proof}
Let $d$ vertices have both colours and $z$ have neither, so $\sum_{\vertex}|\palette{\vertex}|=\vcnvertices+d-z$.
Colour each singly coloured vertex with its colour.
Colour the doubly coloured vertices arbitrarily, and then perform improving moves, only on doubly coloured vertices, in the subgraph induced by vertices with non-empty palettes.
At termination, each doubly coloured vertex has at most one uncut incident edge in this subgraph, since its degree is at most three.
Every edge counted by $\nbichrom_{\srcgraph}$ that is uncut has a doubly coloured endpoint, so at least $\nbichrom_{\srcgraph}-d$ of them are cut.
Next, colour the vertices with empty palettes, and perform improving moves only on them, now in the full graph.
Each such vertex finishes with at least two cut incident edges; counting each edge at most twice, at least $z$ cut edges are incident to empty-palette vertices.
The edges counted before are unaffected, so the final cut has size at least $\nbichrom_{\srcgraph}-d+z=\nbichrom_{\srcgraph}+\vcnvertices-\sum_{\vertex}|\palette{\vertex}|$.
Each phase terminates after at most $\vcmedges$ improving moves.
\end{proof}
 
\begin{mysublemmastep}
We can select in deterministic polynomial time one representative clone per cluster whose palettes, assigned to the corresponding source vertices, satisfy
\begin{align}
    \nbichrom_{\srcgraph}+\vcnvertices-\sum_{\vertex\in\vertices}|\palette{\vertex}|
    \geq
    \frac{\nbichrom}{\vcnvertices^2}+\vcnvertices-\frac{1}{\vcnvertices}\sum_{i\in\blowvertices}|\palette{i}|.
\end{align}
\end{mysublemmastep}
 
\begin{proof}
Choose one representative independently and uniformly from each cluster.
Each edge of $\blowgraph$ lies between two clusters joined by a source edge, and each source edge has $\vcnvertices^2$ such edges; hence the expected value of $\nbichrom_{\srcgraph}$ is $\nbichrom/\vcnvertices^2$.
Each cluster has $\vcnvertices$ clones, so the expected value of $\sum_{\vertex}|\palette{\vertex}|$ is $\frac{1}{\vcnvertices}\sum_i|\palette{i}|$.
The left-hand side is a sum of vertex and edge contributions, so its conditional expectation given the representatives of any subset of clusters can be computed exactly in polynomial time.
Fixing clusters one at a time, always keeping a representative whose conditional expectation is at least the current one (method of conditional expectations), yields the claim.
\end{proof}
 
Applying Step \circled{1} to the representatives of Step \circled{2} and multiplying by $\vcnvertices^2$ gives \cref{eq:rounding}.
No random choices are made.
\end{proof}

\end{document}